\documentclass[fleqn,10pt]{wlscirep}
\usepackage[utf8]{inputenc}
\usepackage[T1]{fontenc}
\usepackage{cite}
\usepackage[ruled,vlined]{algorithm2e}
\usepackage{amsmath,amssymb,amsfonts}
\usepackage{placeins}
\usepackage{multirow}
\usepackage{graphicx}
\usepackage{algpseudocode} 
\usepackage{textcomp}
\usepackage{bbm}
\usepackage{booktabs}
\usepackage{hyperref}
\usepackage{morefloats}
\usepackage{comment} 
\usepackage{adjustbox}
\usepackage{makecell}

\begin{abstract}

Conventional subgroup analyses can yield unstable and difficult-to-interpret conclusions. Multiplicity, post-hoc subgroup searching, and small within-subgroup samples can make apparent treatment-effect heterogeneity difficult to distinguish from sampling variation. These challenges are even greater in fully observational biomedical datasets. Each individual is observed under only one exposure state, ground-truth individual treatment effects are unavailable, and the causal structure is uncertain. In addition, clinically relevant states such as obesity or elevated glucose are not directly assignable treatments. Furthermore, standard hard-assignment rules often ignore inherent phenotypic ambiguities, failing to account for clustering assignment uncertainties near subgroup boundaries. We therefore investigate whether subgroups constructed from pre-treatment patient characteristics, without using exposure, outcome, or estimated treatment-effect information, can serve as interpretable units for budget-constrained and uncertainty-aware policy prioritization. We propose an end-to-end framework combining ensemble causal discovery and domain knowledge, discovery--evaluation sample splitting, inductive unsupervised clustering, uncertainty-aware subgroup selection, and held-out doubly robust policy evaluation. The discovered graphs are treated only as potential causal structures for identifying plausible pre-treatment covariates and adjustment variables, rather than as confirmed causal graphs. We compare K-means, hard, membership-weighted, and stochastic Fuzzy C-means (FCM), and Bayesian Gaussian mixture (GMM) models with a supervised causal-forest-derived conditional average treatment effect (CATE)-tree comparator. Subgroups are prioritized under a 70\% intervention budget using ungated allocation, Empirical Bernstein (EB) safety gating, or hierarchical Bayesian pooling. The framework is evaluated for hypothetical obesity-to-non-obesity and elevated-to-lower-glucose state shifts for diabetes risk in the PIMA Indians Diabetes dataset (PIMA) and for a lifetime-smoking-history contrast for sleep disturbance in the National Health and Nutrition Examination Survey (NHANES). In representative held-out evaluations, the highest estimated ungated utilities were 0.799 for the body mass index (BMI) policy using Bayesian GMM, 0.735 for the glucose policy using hard or membership-weighted FCM, and 0.775 for the smoking-history policy using K-means. Paired bootstrap comparisons of held-out policy-risk differences were conducted using common resamples of the evaluation individuals. All paired 95\% confidence intervals included zero, and no comparison among the evaluated subgroup policies remained statistically significant after Holm adjustment. Thus, the methods attaining the most favorable point estimates should be interpreted as descriptive point-estimate leaders rather than as policies with established superior held-out performance. The identity of the point-estimate-leading method also varied across policy experiments, providing no evidence that a single subgrouping algorithm consistently dominated. The supervised CATE-tree comparator produced greater within-group effect homogeneity, as expected from its effect-guided construction, but did not consistently provide higher held-out policy utility. Bayesian pooling generally preserved ungated allocations, whereas EB gating was more conservative and occasionally produced a no-shift policy. Policies with similar estimated utility could nevertheless prioritize different individuals. These results suggest that causally constrained, phenotype-first subgroups can provide transparent units for observational policy prioritization. However, because the analyses are observational, the causal graphs are uncertain, the interventions are hypothetical state contrasts, and identification depends on exchangeability, positivity, consistency, and correct nuisance-model estimation. The findings should be interpreted as assumption-dependent decision-support evidence rather than proof of intervention benefit. Future work should propagate causal-graph uncertainty throughout the pipeline, study clinically specified and directly implementable interventions, and validate the resulting policies in external, longitudinal, and prospective cohorts.


\end{abstract}

\begin{document}

\flushbottom

%
%

\author[1]{Vasundhara Acharya}
\author[1]{B\"{u}lent Yener
}

\affil[1]{Rensselaer Polytechnic Institute, Troy, USA (e-mail: acharv2@rpi.edu)}
\affil[1]{Rensselaer Polytechnic Institute, Troy, USA (e-mail: yener@cs.rpi.edu)}

\thispagestyle{empty}
\title{From Unsupervised Subgroups to Hypothetical State-Intervention Policies: An Evaluation of Selected Subgrouping Methods in Observational Health Data
}
\maketitle

\section*{Introduction}

Heterogeneous treatment-effect (HTE) analysis investigates how the effect of an intervention varies across individuals or subgroups \cite{HTE_P3,causalforest}. The most granular estimand is the individual treatment effect (ITE), defined as the difference between an individual’s potential outcomes under treatment and control. In observational biomedical data, each individual is observed under only one exposure or risk-factor state, rendering the counterfactual and true ITE unobservable \cite{holland1986statistics}. As a result, evaluation metrics that require ground-truth individual effects, such as precision in estimation of heterogeneous effect (PEHE), treatment-effect mean squared error (MSE), or individual-level oracle regret, are generally restricted to simulated or semi-synthetic benchmarks \cite{hill2011bayesian,shalit2017}.

A more practical estimand is the conditional average treatment effect (CATE), which represents the average effect of a treatment or state change conditional on observed covariates \cite{kunzel2019metalearners}. In this study, CATE-oriented estimation is employed to summarize subgroup-level benefit rather than to assert recovery of true individual effects. Hypothetical state-shift policies are considered for modifiable risk-factor states, including obesity, elevated glucose, and smoking history. Under a fixed budget, these policies prioritize eligible subgroups for a shift from an adverse state to a lower-risk state. Decisions are made primarily at the subgroup level, with partial allocation applied only when the budget boundary intersects a subgroup.

Subgroups may be formed through supervised or unsupervised procedures. Supervised HTE methods, such as Virtual Twins, causal trees, and causal forests, utilize treatment and outcome information to estimate heterogeneity or define subgroups \cite{virtualtwins,HTE_P3,causalforest}. In contrast, unsupervised subgroup discovery forms groups using only pretreatment covariates, with estimated benefit subsequently used for policy ranking. This study investigates whether such phenotype-based subgroups can support effective and transparent intervention policies in observational data. This inquiry is motivated by clinical workflows in which patients are stratified prior to treatment based on baseline characteristics, disease stage, imaging findings, molecular profiles, or other pretreatment biomarkers. For example, oncology decisions are frequently informed by tumor or molecular characteristics measured before treatment assignment \cite{lindeman2018updated,sepulveda2017molecular}. The present setting differs from guideline-defined biomarker stratification because the subgroup-defining variables are data-driven rather than established clinical biomarkers. \textbf{In this context, the term phenotype refers to the pretreatment covariate profile used for subgroup discovery}.

Conventional subgroup analyses are susceptible to multiplicity, post-hoc subgroup searching, small within-subgroup samples, and unstable effect estimates \cite{HTE_P4}. The proposed framework addresses some of these concerns by defining subgroups without using outcomes or estimated treatment effects, separating policy construction from held-out evaluation, and fixing the subgroup structure and ranking prior to evaluation. Pretreatment covariates are selected using causal-structure discovery and causal-domain reasoning \cite{acharya2025understanding}, followed by the application of multiple inductive clustering algorithms to construct reusable subgroups. Model-based state-shift benefits are summarized among eligible discovery-cohort individuals and used to rank subgroups within the policy budget. The resulting fixed policies are then evaluated in the held-out cohort using an intervention-specific doubly robust estimator. Formal multiplicity control is implemented at two levels: the Empirical Bernstein gate uses a Bonferroni-style adjustment across the \(K\) subgroup-specific bounds within each clustering solution, while paired policy-risk comparisons are adjusted using Holm's procedure within each policy experiment.

To address uncertainty in subgroup-level benefit estimates, an EB gate, which requires a conservative positive lower bound, is compared with a hierarchical Bayesian pooling gate that borrows information across subgroups while accounting for estimation uncertainty. Stochastic FCM hardening is also employed to propagate ambiguity in fuzzy cluster membership into the resulting allocation. These procedures address specific sources of estimation and membership uncertainty but do not eliminate the causal-identification assumptions or residual uncertainty inherent in observational analyses.

The primary empirical question is whether the choice of subgrouping method materially alters policy performance or allocation composition. Population-level held-out policy risk and utility, selected subgroups, subgroup profiles, and targeted-person overlap are compared across K-means, multiple FCM variants, Bayesian GMMs, and a supervised CATE-tree comparator. The comparator evaluates whether treatment-effect-informed subgroup construction improves held-out performance relative to phenotype-first clustering. Given that the analyses are observational and the intervention targets are hypothetical state contrasts, all estimates are interpreted as assumption-dependent decision-support evidence rather than definitive proof of intervention benefit.

\section*{Related Work}
\label{sec:related_work}

The population average treatment effect summarizes the expected causal contrast across an entire target population. However, this metric can obscure important differences in average benefit among clinically distinct patient subgroups \cite{HTE_P1}\cite{HTE_P2}\cite{HTE_P4}. This limitation has driven extensive research into heterogeneous treatment-effect estimation, conditional average treatment effects, and the identification of treatment-responsive subgroups \cite{HTE_P3}.

Several widely used approaches are supervised, meaning that treatment and outcome information directly influence the estimation of heterogeneity or the resulting subgroup structure. Foster et al. \cite{virtualtwins} introduced the Virtual Twins method, which first predicts each individual's response under both treatment and control conditions. The predicted treatment contrast is then used as the target in a second-stage classification or regression model to identify subgroups of patients with potentially enhanced treatment effects.
Athey and Imbens \cite{HTE_P3} introduced causal trees, which recursively partition the pre-treatment covariate space to identify regions with differing estimated treatment effects. Honest estimation separates tree construction from effect estimation to reduce adaptive bias. The causal forest, proposed by Wager and Athey \cite{causalforest}, extends recursive partitioning by employing ensembles of trees and offers flexible, nonparametric estimates of conditional treatment effects. In contrast to the causal tree approach, which aims to construct a single well-trained tree, the causal forest builds multiple deep trees with small leaves.

Chernozhukov et al. \cite{GATES} introduced the Grouped Average Treatment Effects Sorted by the proxy (GATES) procedure, which sorts individuals according to a machine-learning proxy for treatment benefit and estimates average treatment effects within the resulting impact groups using sample splitting. These groups are defined by the ranking of an estimated effect score rather than by similarity in original patient characteristics. Kim et al. \cite{causalkmeans} proposed Causal K-Means clustering, an unsupervised framework for identifying subgroups with heterogeneous treatment effects by clustering on unknown counterfactual regression functions instead of observed covariates. The authors developed both a simple plug-in estimator and a bias-corrected semiparametric estimator using efficient influence functions and cross-fitting. This framework accommodated multi-treatment and multi-outcome settings and was demonstrated through simulations and a PROPEL chronic low back pain trial case study. Wang et al. \cite{wang2025causal} constructed a learned similarity kernel from an orthogonalized causal forest and applied kernelized clustering to identify individuals with similar estimated treatment responsiveness. Their procedure residualized treatment and outcomes using the Robinson decomposition and employed forest-derived weights to encode local similarity in the estimated CATE surface. Kernelized clustering was then applied to this similarity representation, and treatment-effect estimates were summarized within the resulting clusters to obtain cluster-specific average effects.

Unsupervised phenotype-based subgrouping has also been utilized in randomized clinical studies. Bellavia et al. \cite{bellavia2025unsupervised} proposed an unsupervised phenotype-based approach for assessing treatment-effect heterogeneity in randomized clinical trials. This method applies model-based clustering to baseline patient characteristics, followed by estimation of randomized treatment effects within the resulting phenotypes. In the ENGAGE AF-TIMI 48 trial, this approach identified clinically distinguishable patient groups with different estimated responses to edoxaban compared to warfarin. Comparison with generalized random forests highlighted a central tradeoff. The effect-informed stratification produced stronger treatment-effect separation, while baseline phenotype clustering yielded groups that were more easily characterized using observed patient attributes. Sinha et al. \cite{sinha2021comparison} compared multiple supervised and unsupervised subgrouping algorithms across three randomized ARDS trials and found that no single method consistently detected treatment-effect heterogeneity across all studies. They also observed that subgroup composition and HTE conclusions were often sensitive to algorithm choice, random initialization, and biomarker inclusion, motivating robustness analyses across clustering methods, random seeds, and selected-population overlap. Bhatraju et al. \cite{bhatraju2019identification} identified acute-kidney-injury subphenotypes with differing molecular profiles and responses to vasopressin. Zampieri et al. \cite{zampieri2019heterogeneous} used machine-learning-derived subgroups to examine heterogeneity in the effect of alveolar recruitment in acute respiratory distress syndrome. This subgroup-first logic resembles certain clinical decision-making workflows, although clinical stratification should not be equated with unsupervised clustering. In precision oncology, patients are commonly stratified using pretreatment disease characteristics such as histology, stage, receptor status, and molecular biomarkers \cite{lindeman2018updated,sepulveda2017molecular}. These are expert-defined and biologically validated strata rather than data-driven clusters, but they exemplify a clinically meaningful subgroup-first principle.

In addition to clinical motivations for phenotype-based subgrouping, a key methodological question concerns the algorithmic construction of such subgroups and the applicability of resulting subgroup definitions to future patients.
Clustering methods can be broadly categorized as inductive or transductive \cite{clustering1}. Inductive clustering methods learn a reusable assignment rule from observed data, enabling assignment of new observations to previously defined groups. In contrast, transductive clustering methods define partitions specific to the sample available at learning time. Classical centroid- and membership-based methods, such as K-means and FCM, are considered inductive because learned centroids or membership functions facilitate out-of-sample assignments, whereas agglomerative hierarchical clustering is typically described as transductive \cite{clustering2}. Out-of-sample approximations have been proposed for several otherwise sample-dependent methods, including spectral or manifold-based methods \cite{clustering4} and density-based hierarchical clustering \cite{clustering8}. However, these approaches require an additional approximation step to assign new observations after the original clustering has been learned. Accordingly, this work focuses on inductive clustering algorithms for constructing reusable patient subgroups.

A substantial body of literature addresses the estimation of heterogeneous treatment effects at the individual covariate-profile level. Causal forests employ ensembles of honest recursive partitions to provide flexible, nonparametric estimates of the conditional average treatment effect \cite{causalforest}. Meta-learning frameworks, including the S-, T-, and X-learners, construct CATE estimators using standard supervised-learning algorithms. The X-learner is particularly suited to settings with imbalanced treatment-group sizes \cite{kunzel2019metalearners}. Representation-learning approaches, such as counterfactual regression \cite{shalit2017}, aim to improve individual-effect estimation by learning covariate representations that reduce discrepancies between treated and untreated populations. Related methods include orthogonalized R-learners \cite{nie2021quasi} and Bayesian models such as Bayesian additive regression trees (BART) and Bayesian causal forests \cite{hill2011bayesian,hahn2020bayesian}. These approaches are primarily designed to estimate or rank the conditional treatment effect, thereby supporting policies that make different decisions for individuals with distinct covariate profiles. Souto et al. \cite{psbart} introduced K-Fold Causal BART, which demonstrated promising performance on synthetic data. Their findings indicated that ps-BART may be the most reliable general-purpose option. However, they also emphasized that model performance is highly dependent on treatment-effect heterogeneity and that uncertainty quantification in causal inference remains challenging when effects are homogeneous.

Off-policy evaluation methods can be broadly classified into three categories \cite{policy_p5}. Direct methods estimate an outcome model and use it to predict the mean outcome under the target policy, corresponding to outcome regression or the G-computation formula in causal inference. Inverse-propensity or importance-weighting methods reweight observed outcomes to account for differences between the observed treatment-assignment mechanism and the target policy. Doubly robust estimators combine these two components by using outcome-model predictions together with an inverse-propensity-weighted residual correction.
Among these approaches, doubly robust estimation is particularly relevant for observational policy evaluation because it combines model-based prediction with correction for the observed treatment-assignment mechanism \cite{athey2021policy}\cite{policy_p6}. Dudík et al. \cite{dudik2011doubly} introduced Doubly Robust Policy Evaluation and Learning, demonstrating that combining outcome modeling with propensity correction yields accurate value estimates when either nuisance component is well estimated. They reported consistent empirical improvements over pure direct-modeling or pure weighting baselines in contextual-bandit problems. Athey and Wager \cite{athey2021policy} showed that integrating doubly robust estimation with cross-fitting provides a theoretically optimal framework for empirical policy learning. This approach achieves rate-optimal regret for best-in-class treatment assignment rules and remains highly flexible, accommodating modern machine learning algorithms for nuisance estimation. The framework also addresses real-world deployment constraints by permitting sensitive attributes for confounding adjustment during training, even when such attributes must be excluded from the deployed policy. Kitagawa and Tetenov \cite{kitagawa2018treated} proposed empirical welfare maximization, which selects a treatment rule by maximizing estimated social welfare over a constrained class of candidate policies. This framework is relevant when treatment rules must remain simple or when only a limited proportion of the population can be treated due to budget or capacity constraints. Kallus et al. \cite{policy_p4} developed off-policy evaluation and learning methods robust to both distribution shift and slow estimation of nuisance functions such as behavior propensities. Their work extends doubly robust ideas from standard off-policy evaluation and learning (OPE/OPL) to distributionally robust settings under Kullback–Leibler (KL) and general f-divergence uncertainty sets.

\par Ranking subgroups solely by point-estimated causal benefit can result in unstable policy decisions. An apparently large subgroup effect may be attributable to estimation noise, limited subgroup sample size, or substantial variation in estimated benefit among subgroup members. As a result, prioritizing subgroups based only on estimated mean effects may allocate interventions to populations where the underlying benefit is uncertain. This issue is particularly significant when intervention deployment is costly or potentially harmful, highlighting the need for high-confidence lower bounds rather than relying solely on point estimates. High-confidence off-policy evaluation asserts that a candidate policy should not be evaluated solely by its point-estimated value, especially when offline estimates are noisy or highly variable. Instead, deployment decisions should be based on a lower confidence bound that penalizes estimation uncertainty and certifies that policy performance exceeds an acceptable threshold \cite{thomas2015high}. Extending this principle to policy optimization, Jin et al. \cite{jin2025policy} introduced pessimistic policy learning, which selects a policy by maximizing an augmented inverse-propensity-weighted value estimate minus a policy-dependent uncertainty penalty. Their analysis relaxes the requirement of uniform overlap across the entire policy class. This is particularly relevant for subgroup allocation, as policies recommending interventions in poorly represented patient strata may yield large but unstable value estimates due to extreme propensity weights.




Although prior research offers methods for heterogeneous-effect estimation, phenotype discovery, uncertainty-aware decision making, and off-policy evaluation, limited attention has been devoted to integrating these components when subgroups must be constructed prior to effect estimation using fully observational data. This context necessitates the selection of plausible pretreatment variables without conditioning on mediators or exposure consequences \cite{schisterman2009overadjustment,vanderweele2019principles}, the interpretation of risk-factor modifications as hypothetical state contrasts rather than directly assignable actions \cite{hernan2008does}, the consideration of subgroup-effect and membership uncertainty \cite{su2015hierarchical}, and the evaluation of both policy value and the composition of the selected population.

This work addresses this gap by proposing a causal-discovery-informed, phenotype-first framework for constructing budget-constrained and uncertainty-aware subgroup policies, evaluated using held-out doubly robust methods. The primary research question is whether pretreatment subgroups formed without treatment-effect information can serve as interpretable and actionable units for policy prioritization in the absence of counterfactual outcomes.

\newpage

\section*{Methodology}
The proposed framework comprises five stages: causal-structure-informed covariate selection, discovery–evaluation sample splitting, pretreatment subgroup construction, uncertainty-aware subgroup ranking, and held-out doubly robust policy evaluation. Subgroups are identified without reference to exposure, outcome, or estimated treatment-effect information. Benefit estimation and policy ranking are conducted only after the subgroup structure is established. Additional analyses evaluate sensitivity to the choice of clustering algorithm, safety gate, fuzzy membership uncertainty, discovery–evaluation split, and selected-population composition.

\subsection{Dataset and Pre-processing}

We use the PIMA Indians Diabetes dataset \cite{PIMA_diabetes_dataset} to evaluate BMI- and glucose-related state-intervention policies for diabetes risk. The outcome of interest is diabetes status. The BMI policy considers an obesity-related state contrast, and the glucose policy considers an elevated-glucose state contrast. The dataset is publicly available at \url{https://www.kaggle.com/datasets/uciml/PIMA-indians-diabetes-database}.

We also utilize the National Health and Nutrition Examination Survey (NHANES), available at \url{https://wwwn.cdc.gov/nchs/nhanes/}, to evaluate prioritization of smoking-related policies. The primary outcome is sleep disturbance, and the policy is operationalized as a contrast between smoking states, where individuals classified as smoking-exposed are hypothetically shifted to the reference non-smoking state. Details regarding the construction of the NHANES cohort, including survey-cycle combination and variable formation, are provided in \cite{acharya2025understanding}. In this study, the NHANES analysis serves as an analytic-sample methodological case study rather than a nationally representative U.S. population policy analysis. Consequently, NHANES survey weights, strata, and primary sampling units are not incorporated into the subgroup discovery, safety-gating, or held-out policy-value estimation steps. Therefore, the reported NHANES policy risks, utilities, and uncertainty intervals should be interpreted as analytic-sample estimates rather than as design-based nationally representative estimates.

Because the NHANES subgrouping variables are categorical, we do not apply Euclidean clustering algorithms directly to raw category codes. Instead, we transform the selected pre-treatment effect modifiers using multiple correspondence analysis (MCA), a standard dimension-reduction method for multivariate categorical data \cite{greenacre1984theory,greenacre2006multiple,abdi2007multiple}. The retained MCA coordinates provide a common Euclidean representation for clustering algorithms. We acknowledge that classical MCA treats category levels nominally and does not explicitly incorporate the ordering present in ordinal variables. We recommend that future researchers consider ordered MCA extensions \cite{MCA_variant}. The number of retained MCA dimensions is selected using adjusted inertia following the Benzécri-Greenacre correction, which is commonly used because raw MCA inertia can be distorted by the indicator coding of categorical variables \cite{nenadic2007correspondence,abdi2007multiple}. This preprocessing approach is consistent with prior uses of MCA and MCA-based clustering for categorical health data \cite{MCA_PAPER1,MCA_PAPER2,MCA_PAPER3,MCA_PAPER4,MCA_PAPER5}. Confounding adjustment and policy evaluation are performed separately using one-hot-encoded adjustment covariates.

To address missing data in both datasets, we restricted our sample to complete cases by excluding observations with missing values. Although this approach may introduce selection bias when missingness is not completely at random, standard prediction-oriented imputation models may also be unsuitable if they do not preserve the causal and temporal relationships required for the analysis. Future research should evaluate multiple-imputation strategies specifically designed for causal estimation, including graph-informed imputation models \cite{kyono2021miracle,acharya2025understanding}.


\subsection{Discovery of potential causal links}

For the PIMA dataset, we inferred potential causal links using four causal structure discovery algorithms: Fast Causal Inference (FCI) with a kernel conditional independence (KCI) test \cite{spirtes2001anytime}, Peter–Clark (PC) with KCI, the nonlinear NOTEARS score-based method \cite{zheng2020learning}, and Pillai’s test \cite{pillai_test_ankur}. We additionally used additive-noise-model (ANM) \cite{hoyer2008nonlinear} based direction tests as a supplementary orientation signal. For FCI, we adopted a conservative safety interpretation. We used only the learned adjacencies as evidence of potential links. We did not treat orientations or bidirected edges, often indicative of latent confounding rather than direct causation, as reliable directed causal relationships.

The final ensemble causal graph was constructed using the three-stage ensemble causal structure discovery (CSD) procedure described previously \cite{acharya2025understanding}. Adaptive bootstrapping was first applied to determine stable edge-confidence estimates. Algorithm-specific bootstrapped directed acyclic graphs (DAGs) were generated and filtered based on edge confidence, significance testing, direction-conflict resolution, and cycle removal. The resulting graphs were then aggregated into a consensus DAG using composite edge-confidence scores. A domain-informed causal graph was also developed by integrating prior clinical knowledge and literature evidence, retaining directed edges only when the parent-child relationship was supported by biologically or epidemiologically plausible evidence. Edges involving age, pregnancies, skinfold or skin thickness, insulin resistance, blood pressure, and diabetes outcome were supported by epidemiologic, cohort, and meta-analytic studies \cite{cheng2022age,li2016mechanisms,ruiz2020skinfold,emdin2015usual,fazeli2020aging}.

For the NHANES dataset, the same three-stage ensemble CSD procedure was applied. Candidate graphs were generated using constraint-based PC with discrete conditional-independence tests, score-based greedy hill-climbing under multiple discrete scoring criteria, and Fast Greedy Equivalence Search (FGES) with the BDeu score. These algorithm-specific graphs were aggregated using adaptive bootstrapping, edge-confidence estimation, conflict and cycle resolution, and the composite-confidence consensus procedure described above \cite{PC,hill_climb,fast_ges}.

It is important to note that causal discovery analysis used to identify candidate baseline covariates, adjustment variables, and effect modifiers was conducted as a design-stage analysis and treated as fixed for the present policy-learning pipeline. We therefore do not re-run causal discovery separately within each discovery/evaluation split.

\subsection{Clustering Algorithms}
\label{subsec:clustering_algorithms}

Several inductive clustering algorithms are compared for constructing subgroup-level policies. Each clustering algorithm is trained exclusively on pre-treatment covariates within the discovery cohort. The resulting clustering rule is then fixed and applied to the held-out evaluation cohort, ensuring that evaluation individuals are assigned to subgroups without re-estimating the clustering model. Here, $X_i$ represents the pre-treatment covariate vector for individual $i$, and $K$ denotes the number of clusters used for controlled comparison across methods.

Once subgroup assignments are determined in the discovery cohort, subgroup-level benefit estimates are calculated among eligible individuals and used to construct a ranking of subgroups on the discovery side. In ungated policy variants, this ranking is used solely for intervention budget allocation, with no uncertainty threshold imposed prior to allocation. Consequently, if the budget exceeds the number of eligible individuals in higher-ranked beneficial subgroups, the ungated policy may allocate the remaining budget to lower-ranked subgroups with weak or negative estimated gain. The resulting subgroup ranking is then fixed and applied to the held-out evaluation cohort.

\subsubsection{K-means}
\label{subsubsec:kmeans}

K-means provides a simple hard-clustering baseline \cite{macqueen1967some}. It
partitions the discovery cohort into $K$ clusters by minimizing the within-cluster
sum of squared distances as shown in the equation \ref{kmeans_equation}.

\begin{equation}
    \min_{\{C_i\}_{i=1}^n, \{\mu_c\}_{c=1}^K}
    \sum_{i=1}^n
    \left\|X_i - \mu_{C_i}\right\|_2^2,
    \label{kmeans_equation}
\end{equation}

where $C_i \in \{1,\ldots,K\}$ is the cluster assignment for individual $i$ and
$\mu_c$ is the centroid of cluster $c$.

After the centroids are learned on the discovery cohort, a new evaluation
individual is assigned to the nearest centroid as shown in the equation \ref{centroid}.
\begin{equation}
    C_i
    =
    \arg\min_{c \in \{1,\ldots,K\}}
    \left\|X_i - \mu_c\right\|_2^2.
    \label{centroid}
\end{equation}

K-means is useful in this framework because it is simple, interpretable, and
inductive.  It also
implicitly favors compact, approximately spherical clusters.

\subsubsection{FCM}
\label{subsubsec:fuzzy_cmeans}

FCM extends hard clustering by assigning each individual a membership
weight for every cluster \cite{dunn1973fuzzy, bezdek1981pattern,bezdek1984fcm}. Let
$u_{ic} \in [0,1]$ denote the membership weight of individual $i$ in cluster
$c$ as shown in the equation \ref{membership_fcm1}.

\begin{equation}
    \sum_{c=1}^K u_{ic} = 1.
    \label{membership_fcm1}
\end{equation}

FCM estimates memberships and cluster centers by minimizing objective function shown in Equation~\ref{objective_function}
\begin{equation}
    J_m(U, V)
    =
    \sum_{i=1}^n
    \sum_{c=1}^K
    u_{ic}^{m}
    \left\|X_i - v_c\right\|_2^2
    \label{objective_function}
\end{equation}

where $V=\{v_1,\ldots,v_K\}$ are the fuzzy cluster centers and $m>1$ is the
fuzziness parameter. Larger values of $m$ produce fuzzier memberships, while
values closer to 1 approach hard clustering. In our experiments, we set
$m=1.7$, which lies within the commonly suggested range
$m \in [1.5,2.5]$ for FCM applications \cite{wu2012analysis}. We did
not tune $m$ to maximize policy utility. It was fixed a priori based on the
fuzzy clustering literature \cite{fuzzycmeans2}.

For the policy experiments, we evaluate three FCM variants.

\paragraph{Hard maximum-membership assignment : FCM Hard.}
The first variant converts fuzzy memberships into hard labels using the maximum
membership rule as shown in the equation \ref{membership}.
\begin{equation}
    C_i^{\mathrm{hard}}
    =
    \arg\max_{c \in \{1,\ldots,K\}} u_{ic}.
    \label{membership}
\end{equation}

Let $\widehat{b}_i$ denote the estimated individual benefit for eligible
individual $i$, where positive values indicate that the adverse state has higher
risk than the healthier state. The hard-assignment cluster benefit is computed using the equation \ref{hard}.
\begin{equation}
    \widehat{\tau}_c^{\mathrm{hard}}
    =
    \frac{
    \sum_{i \in \mathcal{D}_{\mathrm{disc}}}
    \mathbbm{1}\{E_i=1\}
    \mathbbm{1}\{C_i^{\mathrm{hard}}=c\}
    \widehat{b}_i
    }{
    \sum_{i \in \mathcal{D}_{\mathrm{disc}}}
    \mathbbm{1}\{E_i=1\}
    \mathbbm{1}\{C_i^{\mathrm{hard}}=c\}
    },
    \label{hard}
\end{equation}
where $E_i=1$ denotes eligibility for the state intervention, and
$\mathcal{D}_{\mathrm{disc}}$ is the discovery cohort.

The train-side gain used for subgroup ranking is computed using the equation \ref{train_gain} and equation \ref{n_hard}.
\begin{equation}
    \mathrm{Gain}_c^{\mathrm{hard}}
    =
    \widehat{\tau}_c^{\mathrm{hard}}
    \times n_c^{\mathrm{hard}},
    \label{train_gain}
\end{equation}

\begin{equation}
    n_c^{\mathrm{hard}}
    =
    \sum_{i \in \mathcal{D}_{\mathrm{disc}}}
    \mathbbm{1}\{E_i=1\}
    \mathbbm{1}\{C_i^{\mathrm{hard}}=c\}.
    \label{n_hard}
\end{equation}

\paragraph{Membership-weighted cluster benefit summaries: FCM Weighted}
The second variant uses the fuzzy memberships to smooth the cluster-level benefit
estimates.

Instead of assigning each eligible individual entirely to one cluster,
each individual contributes to cluster $c$ in proportion to a membership weight.
Specifically, the membership weight is defined in Equation~\ref{eq:fcm_membership_weight} as

\begin{equation}
    w_{ic}=u_{ic}^{m}
    \label{eq:fcm_membership_weight}
\end{equation}

matching the membership weighting used in the FCM objective. The
membership-weighted cluster benefit is then computed using the equation \ref{membership_weighted}. 
\begin{equation}
    \widehat{\tau}_c^{\mathrm{soft}}
    =
    \frac{
    \sum_{i \in \mathcal{D}_{\mathrm{disc}}}
    \mathbbm{1}\{E_i=1\}
    w_{ic}
    \widehat{b}_i
    }{
    \sum_{i \in \mathcal{D}_{\mathrm{disc}}}
    \mathbbm{1}\{E_i=1\}
    w_{ic}}
    \label{membership_weighted}
\end{equation}
For membership-weighted uncertainty estimation, we compute the weighted variance (as shown in the equation \ref{weighted_variance}) and effective sample size.

\begin{equation}
    \widehat{s}_{c,\mathrm{soft}}^2
    =
    \frac{
    \sum_{i}
    \mathbbm{1}\{E_i=1\}
    w_{ic}
    \left(\widehat{b}_i - \widehat{\tau}_c^{\mathrm{soft}}\right)^2
    }{
    \sum_i \mathbbm{1}\{E_i=1\}w_{ic}
    -
    \frac{
    \sum_i \mathbbm{1}\{E_i=1\}w_{ic}^2
    }{
    \sum_i \mathbbm{1}\{E_i=1\}w_{ic}
    }
    }.
    \label{weighted_variance}
\end{equation}

The effective membership-weighted sample size is computed using \ref{effective_size}.
\begin{equation}
    n_c^{\mathrm{eff}}
    =
    \frac{
    \left(
    \sum_i \mathbbm{1}\{E_i=1\}w_{ic}
    \right)^2
    }{
    \sum_i \mathbbm{1}\{E_i=1\}w_{ic}^2
    }.
    \label{effective_size}
\end{equation}

However, the final policy still requires an operational decision about which
individuals are targeted. Therefore, we use hard maximum-membership assignments
for deployment as shown in the equation \ref{hard_deployment}
\begin{equation}
    C_i^{\mathrm{hard}}
    =
    \arg\max_c u_{ic}.
    \label{hard_deployment}
\end{equation}

Thus, in the membership-weighted variant, fuzzy memberships are used for
estimating and smoothing the cluster-level benefit, while hard labels are used
for operational policy deployment. The ranking score is computed using the equation \ref{ranking_score}.
\begin{equation}
    \mathrm{Gain}_c^{\mathrm{soft/hard}}
    =
    \widehat{\tau}_c^{\mathrm{soft}}
    \times n_c^{\mathrm{hard}}.
    \label{ranking_score}
\end{equation}

This creates an estimation--deployment distinction. Individuals with partial
membership in cluster $c$ can inform the estimated cluster benefit
$\widehat{\tau}_c^{\mathrm{soft}}$, even if they are not ultimately assigned to
cluster $c$ under the hard maximum-membership rule. For example, a borderline
individual may contribute to the estimated benefit of cluster 2 through a
nonzero membership weight, but cluster 2's policy decision is applied only to
individuals hard-assigned to cluster 2. This is not a contradiction. The soft
stage is used for estimation smoothing, while the hard stage is used for
operational decision-making. But, it is an important limitation because
the individuals who inform the cluster effect are not exactly the same as the
individuals who receive the cluster-level policy decision.

\paragraph{Stochastic hardening of fuzzy memberships: FCM Stochastic.}

The third variant propagates fuzzy membership uncertainty into policy evaluation. This is motivated by recent work emphasizing uncertainty quantification for fuzzy clustering assignments \cite{wu2026statistical}. Another work on obesity phenotyping suggests that cardiometabolic risk may vary along continuous and overlapping phenotypic gradients rather than through sharply separated subtypes \cite{zhao2026mapping}. This motivates us to avoid treating cluster membership as fully certain.

For each Monte Carlo (MC) draw (r=1,\ldots,R), we sample a hard cluster assignment for each held-out evaluation individual as shown in the equation \ref{eq1} and equation \ref{eq2}. 

\begin{equation}
C_i^{(r)}
\sim
\mathrm{Categorical}
\left(
p_{i1},\ldots,p_{iK}
\right),
\label{eq1}
\end{equation}

\begin{equation}
p_{ic}
=
\frac{
u_{ic}^{\gamma}
}{
\sum_{\ell=1}^{K}u_{i\ell}^{\gamma}
}.
\label{eq2}
\end{equation}

In the main stochastic hardening experiment, we use $(\gamma=1)$, so assignments are sampled directly from the fuzzy membership probabilities.

The discovery-side subgroup ranking is kept fixed. Specifically, subgroup effects are estimated in the discovery cohort using membership-weighted fuzzy cluster summaries, and this ranking is frozen before evaluation. Stochasticity is introduced only at deployment by repeatedly sampling held-out hard cluster assignments from the learned fuzzy memberships and applying the frozen discovery ranking.

For each MC draw, the sampled labels $C_i^{(r)}$ are used to construct a hard subgroup policy under the same budget constraint. Let $d_i^{(r)} \in \{0,1\}$ denote whether eligible individual $i$ is selected in draw $r$. To represent boundary-cluster randomization as a subgroup-level policy rather than as an individual-specific random choice, we convert the realized targeting vector into a cluster-level shift probability $s_i^{(r)}\in[0,1]$. Fully selected clusters have $s_i^{(r)}=1$, unselected clusters have $s_i^{(r)}=0$, and a boundary cluster receives a fractional shift probability equal to the proportion of eligible individuals selected within that cluster.

The policy risk for draw (r) is evaluated using the selective one-way state-shift doubly robust score as explained under the section \ref{policy}. We summarize the MC distribution by its mean, standard deviation, and central interval. A small MC standard deviation indicates that policy utility is not highly sensitive to fuzzy membership uncertainty. A large MC standard deviation indicates that ambiguous subgroup boundaries materially affect who is selected and, therefore, the resulting policy utility.

\subsubsection{Bayesian Gaussian Mixture Model}
\label{subsubsec:bgmm}

The Bayesian Gaussian mixture model provides a probabilistic clustering baseline. It assumes that each individual belongs to a latent mixture component according to the assignment model in Equation~\ref{eq:bgmm_component_assignment}, while the observed covariates follow the component-specific Gaussian distribution in Equation~\ref{eq:bgmm_component_distribution}.

\begin{align}
z_i \mid \boldsymbol{\omega}
\sim
\mathrm{Categorical}
\left(\omega_1,\ldots,\omega_K\right)
\label{eq:bgmm_component_assignment}
\end{align}
\begin{align}
X_i \mid z_i=c
&\sim
\mathcal{N}
\left(\mu_c,\Sigma_c\right).
\label{eq:bgmm_component_distribution}
\end{align}

In a Bayesian mixture model, priors are placed on the mixture weights and component-specific parameters. A Dirichlet or Dirichlet-process prior can shrink unnecessary components toward negligible weight, allowing the effective number of active components to be inferred from the data \cite{rasmussen2000infinite,blei2006variational}. The posterior responsibility of component (c) for individual (i) is defined in Equation~\ref{eq:bgmm_responsibility}.

\begin{equation}
r_{ic}
=
\Pr\left(z_i=c \mid X_i\right).
\label{eq:bgmm_responsibility}
\end{equation}

 For
policy deployment, we use the maximum a posteriori component assignment as shown in the equation \ref{policy_deploy}.
\begin{equation}
    C_i^{\mathrm{MAP}}
    =
    \arg\max_{c \in \{1,\ldots,K\}} r_{ic}.
    \label{policy_deploy}
\end{equation}

Although the posterior component responsibilities could also be used to sample evaluation-set component assignments, the primary Bayesian GMM policy used maximum-a-posteriori assignments so that it served as a fixed probabilistic-clustering comparator. This design choice does not imply that assignment uncertainty is absent from the Bayesian GMM.

For controlled comparison with K-means and FCM, we set the upper
component bound of the Bayesian GMM equal to the same number of clusters used by
the other algorithms. This makes the downstream policy comparison more
controlled because all methods are evaluated under the same nominal cluster
budget. However, this is also mentioned as a limitation.

\subsection{Uncertainty-Aware Safety Gating}

This general principle from \cite{thomas2015high} is adapted to subgroup-level intervention allocation. Instead of admitting clusters solely based on a positive estimated causal benefit, two uncertainty-aware subgroup-admission rules are compared. The Empirical Bernstein gate applies a simultaneous lower confidence bound and admits only subgroups whose estimated benefit remains positive after accounting for a finite-sample uncertainty penalty. In contrast, the hierarchical Bayesian gate partially pools subgroup effects and admits a subgroup when its posterior probability of positive benefit exceeds a prespecified threshold. These two gates represent distinct approaches to decision-making under uncertainty rather than interchangeable versions of the same criterion.


\subsubsection{Empirical Bernstein Safety Gate}
\label{subsubsec:empirical_bernstein_gate}


For each cluster $c$, $\widehat{\tau}_c$ represents the estimated mean benefit, $s_c^2$ denotes the within-cluster variance of the estimated individual benefits, and $n_c$ indicates the number of eligible individuals in the cluster. The variable $K$ refers to the total number of clusters, and $\delta$ specifies the total allowable probability that at least one of the cluster-level bounds fails. Consequently, the corresponding nominal simultaneous confidence level is $1-\delta$.

Multiplicity presents a significant concern in subgroup analysis, as evaluating multiple subgroup-specific effects increases the likelihood that a favorable estimate may result from sampling variation \cite{HTE_P4}. To address the $K$ cluster-level comparisons, the total error probability $\delta$ is distributed across clusters using a Bonferroni-style union-bound adjustment. Assigning an error probability of approximately $\delta/K$ to each cluster yields the term presented in equation \ref{cluster_K}.

\begin{equation}
\log\left(\frac{2K}{\delta}\right),
\label{cluster_K}
\end{equation}
so that the collection of cluster-level bounds is intended to hold simultaneously with probability at least $1-\delta$, under the assumptions of the empirical Bernstein inequality \cite{bernstein}.

We construct an empirical Bernstein lower confidence bound for each cluster \cite{bernstein}. Because the estimated benefit scores are differences between two predicted probabilities, they are bounded in $[-1,1]$. We first apply an affine transformation to rescale them to $[0,1]$. Let $\widetilde{\tau}_c$ and $\widetilde{s}_c^2$ denote the within-cluster mean and variance of the rescaled scores. The lower confidence bound on the rescaled scale is computed using the equation \ref{eq:eb_lcb_rescaled}.
\begin{equation}
\widetilde{\mathrm{LCB}}_{c} = \widetilde{\tau}_c - \left[ \sqrt{ \frac{ 2\widetilde{s}_c^2\log\left(\frac{2K}{\delta}\right) }{ n_c } } + \frac{ 7\log\left(\frac{2K}{\delta}\right) }{ 3(n_c-1) } \right].
\label{eq:eb_lcb_rescaled}
\end{equation}

The bound is then transformed back to the original benefit scale using the equation \ref{eq:eb_lcb_original}.

\begin{equation}
\mathrm{LCB}_{c} = -1 + 2\widetilde{\mathrm{LCB}}_{c}.
\label{eq:eb_lcb_original}
\end{equation}

In hard-assignment subgrouping methods, $n_c$ represents the number of eligible discovery-set individuals assigned to cluster $c$. Consequently, both the empirical Bernstein radius and the robust gain are calculated using the hard eligible cluster size.

In the membership-weighted FCM variant, cluster-level summaries are computed using fuzzy membership weights. Since fuzzy memberships result in fractional cluster participation, the empirical Bernstein uncertainty term is calculated based on the effective sample size.

A cluster satisfies the empirical Bernstein safety criterion if its lower confidence bound on the original benefit scale is positive, as shown in Equation \ref{eq:eb_gate_here}.

\begin{equation}
\mathrm{LCB}_{c} > 0.
\label{eq:eb_gate_here}
\end{equation}

In membership-weighted FCM, the empirical Bernstein radius is calculated using the effective sample size $n_c^{\mathrm{eff}}$, with a finite bound requiring $n_c^{\mathrm{eff}}>1$. Cluster admission is determined by the condition $\mathrm{LCB}_c^{\mathrm{w}}>0$, while the general minimum cluster-size rule is applied to the hard eligible count, reflecting the use of hard maximum membership assignments during deployment. Substituting the ordinary sample size and variance with their weighted counterparts represents an effective-sample-size adaptation of the classical empirical Bernstein bound, rather than an exact application of the original theorem. Consequently, the resulting lower bound is interpreted as a conservative subgroup-admission score, rather than an exact finite-sample confidence bound for the true subgroup effect.

To rank the hard-assignment clusters that meet the admission criteria, the robust gain is defined as presented in equation \ref{eq:eb_gain}.

\begin{equation}
\mathrm{Gain}^{\mathrm{EB}}_c=\mathrm{LCB}_{c} \times n_c^{\mathrm{train}},
\label{eq:eb_gain}
\end{equation}
where $(n_c^{\mathrm{train}})$ is the number of eligible individuals in cluster (c) in the discovery cohort.

For weighted FCM, because the final policy is deployed using hard maximum-membership cluster labels, the estimated deployment gain is scaled by the hard eligible discovery-set count as shown in the equation \ref{eq:weighted_eb_gain}.

\begin{equation}
\mathrm{Gain}^{\mathrm{EB},w}_c=\mathrm{LCB}^{w}_{c}\times n_c^{\mathrm{hard,train}}
\label{eq:weighted_eb_gain}
\end{equation}

In hard-assignment methods, clusters that pass the empirical Bernstein gate are ranked according to $(\mathrm{Gain}^{\mathrm{EB}}_c)$. For weighted fuzzy c-means (FCM), admitted clusters are ranked according to $(\mathrm{Gain}^{\mathrm{EB},w}_c)$. The gate is considered conservative because the bound penalizes clusters exhibiting high within-cluster variability, small eligible sample sizes, or a greater number of simultaneous cluster comparisons. This approach addresses cluster-level multiplicity and instability concerns identified in the subgroup-analysis literature. However, it does not address broader analysis-level multiplicity that may result from comparing multiple clustering algorithms, initialization seeds, outcomes, or policy specifications.



\subsubsection{Hierarchical Bayesian Pooling Gate}
\label{subsubsec:full_bayes_gate}

Estimating effects across multiple subgroups using separate unpooled estimates can exaggerate apparent differences, especially in small or noisy groups. Classical multiplicity corrections typically address this issue by adjusting significance thresholds or widening simultaneous uncertainty intervals, but they do not alter subgroup point estimates. Gelman, Hill, and Yajima advocate for hierarchical modeling, where related subgroup effects are estimated jointly and partially pooled toward a common population distribution \cite{gelman2012we}. Following this principle, the Bayesian safety gate estimates cluster-level benefits hierarchically and admits a subgroup for policy allocation only when its posterior probability of exceeding a prespecified benefit threshold is sufficiently high.

This Bayesian gate models cluster-level effects using a hierarchical normal 
model as shown in the equations \ref{eq:bayes_observation_model} and equation \ref{eq:bayes_hierarchical_model}.
\begin{align}
    \widehat{\tau}_c \mid \theta_c 
    &\sim \mathcal{N}\left(\theta_c, \widehat{se}_c^{\,2}\right),
    \label{eq:bayes_observation_model} \\
    \theta_c \mid \mu, \tau^2 
    &\sim \mathcal{N}\left(\mu, \tau^2\right),
    \label{eq:bayes_hierarchical_model}
\end{align}
where $\widehat{\tau}_c$ is the estimated mean benefit for cluster $c$, 
$\widehat{se}_c$ is its standard error, $\theta_c$ is the latent subgroup-level benefit for 
cluster $c$, $\mu$ is the population-level mean benefit, and $\tau^2$ is the 
between-cluster variance.

Weakly informative priors are placed on the population mean and the 
between-cluster standard deviation using the equation \ref{eq:bayes_mu_prior} and the equation \ref{eq:half_student_prior}.
\begin{align}
    \mu &\sim \mathcal{N}\left(0, \sigma_\mu^2\right), 
    \label{eq:bayes_mu_prior} \\
    \tau &\sim \mathrm{Half\text{-}Student}\text{-}t\left(\nu, \sigma_\tau\right).
    \label{eq:half_student_prior}
\end{align}

The Half-Student-$t$ prior in Equation~\ref{eq:half_student_prior} is used 
because $\tau$ is a positive scale parameter. This prior provides weak 
regularization near zero while retaining heavy tails, allowing substantial 
between-cluster heterogeneity when supported by the data 
\cite{gelman2006prior, polson2012half}. This is useful in our setting because 
the number of discovered clusters is small and some cluster-level effect 
estimates may be noisy.

The posterior distribution provides uncertainty summaries for each cluster, 
including the posterior probability that the cluster has positive benefit:
\begin{equation}
    p_c^{+}
    =
    \Pr\left(\theta_c > 0 \mid \mathcal{D}\right),
    \label{eq:posterior_positive_prob}
\end{equation}
where $\mathcal{D}$ denotes the observed discovery data.

In our implementation, the Bayesian gate selects clusters using the posterior 
probability in Equation~\ref{eq:posterior_positive_prob}. Specifically, cluster 
$c$ is considered sufficiently reliable if
\begin{equation}
    \Pr\left(\theta_c > 0 \mid \mathcal{D}\right) \geq \pi_0,
    \label{eq:bayes_prob_gate}
\end{equation}
where we set $\pi_0 = 0.90$.

Among clusters that pass the posterior probability gate in 
Equation~\ref{eq:bayes_prob_gate}, we rank clusters using their posterior mean 
benefit. Let $n_c^{\mathrm{train}}$ denote the number of eligible individuals in 
cluster $c$ in the discovery cohort. The Bayesian cluster-level gain is defined 
as
\begin{equation}
    \mathrm{Gain}^{\mathrm{Bayes}}_c
    =
    \mathbb{E}\left[\theta_c \mid \mathcal{D}\right]
    \times n_c^{\mathrm{train}}.
    \label{eq:bayes_gain}
\end{equation}

Therefore, the final set of Bayesian-gated clusters is
\begin{equation}
    \mathcal{C}_{\mathrm{Bayes}}
    =
    \left\{
    c :
    \Pr\left(\theta_c > 0 \mid \mathcal{D}\right) \geq 0.90
    \ \text{and}\
    \mathrm{Gain}^{\mathrm{Bayes}}_c > 0
    \right\}.
    \label{eq:bayes_selected_clusters}
\end{equation}

The selected clusters in Equation~\ref{eq:bayes_selected_clusters} are ordered 
by $\mathrm{Gain}^{\mathrm{Bayes}}_c$ from Equation~\ref{eq:bayes_gain}. 

For the membership-weighted FCM variant, the same hierarchical Bayesian pooling gate is applied to the membership-weighted cluster benefit estimates and their weighted standard errors. The weighted standard errors are computed using the effective sample size induced by fuzzy memberships. The Bayesian policy gain is scaled by the hard eligible discovery-set count because final deployment uses hard maximum-membership cluster labels. For stochastic FCM, the safety gate is learned once on the discovery cohort using the membership-weighted FCM cluster summaries. In each MC evaluation run, evaluation set individuals are stochastically hardened according to their fuzzy memberships, but the admitted cluster order remains frozen from discovery. Thus, stochasticity affects held-out deployment and policy-value estimation, not the discovery-side safety gate.

Bayesian pooling borrows strength across clusters by shrinking the noisy 
cluster-level estimates toward the population mean \cite{morris1983parametric}. 
Compared with empirical Bernstein gating, this gate may be less conservative 
because it uses posterior evidence of positive benefit, as shown in 
Equation~\ref{eq:bayes_prob_gate}, rather than a distribution-free lower 
confidence bound (LCB) as in Equation~\ref{eq:eb_lcb_rescaled}.

\subsection{Budgeted State-Shift Policy Evaluation}
\label{policy}

In contrast to conventional off-policy settings, where actions are logged under a historical decision rule, and static policies, which assign treatment solely based on baseline covariates, the hypothetical intervention considered here selectively shifts the naturally observed exposure state among prioritized eligible individuals. Consequently, the intervention distribution is determined by the observed exposure process. Wen et al. \cite{wen2023intervention} demonstrated that standard efficient-influence-function estimators are not inherently doubly robust for such interventions. The property of double robustness depends on the specific structure of the intervention distribution. Building on this framework, we employ an intervention-specific doubly robust score for a selective one-way shift from the adverse state to the lower-risk state.

Formally, we consider state interventions on modifiable risk-factor states. For each policy experiment, let $T_i \in \{0,1\}$ denote the relevant adverse-state indicator, where $T_i=1$ represents the eligible adverse state and $T_i=0$ represents the lower-risk state. Let $Y_i(t)$ denote the potential outcome under state $T_i=t$, and let $L_i$ denote the pre-treatment covariates used for policy evaluation.

For each dataset, the adverse-state indicator was defined using a clinically interpretable binary state rule. For the BMI policy, obesity was defined as a BMI of at least (30,$\mathrm{kg/m^2}$), as shown in Equation~\ref{eq:bmi_state_indicator}.

\begin{equation}
T_i := T_{\mathrm{BMI},i}=
\mathbbm{1}{\mathrm{BMI}_i \geq 30}.
\label{eq:bmi_state_indicator}
\end{equation}

so that ($T_i=1$) denotes obesity and ($T_i=0$) denotes non-obesity. This threshold follows the standard adult BMI classification in which obesity is defined as BMI at least (30,$\mathrm{kg/m^2}$) \cite{purnell2015definitions}.

For the glucose policy, the elevated-glucose state was defined using the threshold shown in Equation~\ref{eq:glucose_state_indicator}.

\begin{equation}
T_i := T_{\mathrm{Glucose},i}=
\mathbbm{1}\left(\mathrm{Glucose}_i \geq 140,\mathrm{mg/dL}\right).
\label{eq:glucose_state_indicator}
\end{equation}

Here ($T_i=1$) denotes elevated two-hour plasma glucose and ($T_i=0$) denotes lower glucose. This threshold corresponds to the lower boundary of impaired glucose tolerance or prediabetes in the two-hour oral glucose tolerance test, for which values below (140,$\mathrm{mg/dL}$) are considered normal and values from (140) to (199,$\mathrm{mg/dL}$) are considered prediabetic \cite{bansal2015prediabetes}.

For the smoking-history policy, the adverse-state indicator was defined using the lifetime-smoking criterion shown in Equation~\ref{smoking}.

\begin{equation}
T_i := T_{\mathrm{Smoking},i}=
\mathbbm{1}{\text{Smoker}_i},
\label{smoking}
\end{equation}

Here ($T_i=1$) denotes the Smoker state (reported smoking at least 100 cigarettes during the lifetime) and ($T_i=0$) denotes the non-smoker state. 

\textbf{Because this variable captures lifetime smoking history rather than current smoking status, the corresponding hypothetical shift should not be interpreted as an acute smoking-cessation intervention}. Instead, it represents a counterfactual shift from an established lifetime-smoking state to a non-smoker state among prioritized eligible individuals.

A learned subgroup policy $\pi_q$ induces a selective one-way shift probability $s_{\pi_q}(L_i) \in [0,1]$. For fully selected subgroups, $s_{\pi_q}(L_i)=1$. For unselected subgroups, $s_{\pi_q}(L_i)=0$. For the boundary subgroup, $s_{\pi_q}(L_i)$ equals the fraction of eligible subgroup members selected under the budget constraint. Individuals already observed in the lower-risk state are not shifted. The policy-induced post-intervention state can therefore be written as in equation \ref{intervention}.

\begin{equation}
T_i^{\pi_q}=T_i(1-S_i^{\pi_q}), \qquad S_i^{\pi_q}\sim \mathrm{Bernoulli}(s_{\pi_q}(L_i)) \quad \text{only when } T_i=1,
\label{intervention}
\end{equation}

so that the policy only shifts prioritized eligible individuals from $1 \rightarrow 0$, while all others remain in their observed state.

For a binary adverse outcome, the policy risk is computed as shown in the equation \ref{risk_equation}.

\begin{equation}
R(\pi_q) = \mathbb{E}[Y_i^{\pi_q}] = \mathbb{E}\left[(1-T_i s_{\pi_q}(L_i))Y_i + T_i s_{\pi_q}(L_i)Y_i(0)\right].
\label{risk_equation}
\end{equation}

Under consistency, no interference, conditional exchangeability of $Y_i(0)$ given $L_i$, and positivity, this risk is identified by Equation \ref{eq:identified_risk},
\begin{equation}
R(\pi_q)=\mathbb{E}\left[(1-T_i s_{\pi_q}(L_i))Y_i + T_i s_{\pi_q}(L_i)m_0(L_i)\right],
\label{eq:identified_risk}
\end{equation}
where the conditional outcome mean and propensity score are defined in Equation \ref{eq:nuisance_functions}:
\begin{equation}
m_0(L_i)=\mathbb{E}(Y_i\mid T_i=0,L_i), \qquad e(L_i)=\mathbb{P}(T_i=1\mid L_i).
\label{eq:nuisance_functions}
\end{equation}

Because extreme propensity scores can destabilize evaluation, we apply an overlap restriction. As a result, our estimator targets $R_A(\pi_q) = \mathbb{E}\left[Y_i^{\pi_q}\mid A_i=1\right]$, the policy risk among evaluation individuals satisfying the overlap criterion. We estimate this trimmed risk ($\alpha$=$0.05$)  on the held-out evaluation cohort using the cross-fitted, intervention-specific doubly robust estimator(DR) shown in Equation \ref{eq:dr_estimator}:
\begin{equation}
\widehat{R}_{DR,A}(\pi_q)=\frac{\sum_{i\in\mathcal{D}_{eval}} A_i \widehat{\phi}_i(\pi_q)}{\sum_{i\in\mathcal{D}_{eval}} A_i},
\label{eq:dr_estimator}
\end{equation}
where Equation \ref{eq:trimming_indicator} defines the propensity-overlap trimming indicator:
\begin{equation}
A_i = \mathbf{1}\{\alpha \leq \widehat{e}(L_i) \leq 1-\alpha\},
\label{eq:trimming_indicator}
\end{equation}
and Equation \ref{eq:phi_score} computes the estimated pseudo-outcome score:
\begin{equation}
\widehat{\phi}_i(\pi_q) = (1-T_i\widehat{s}_{\pi_q i})Y_i + T_i\widehat{s}_{\pi_q i}\widehat{m}_0(L_i) + (1-T_i)\widehat{s}_{\pi_q i} \frac{\widehat{e}(L_i)}{1-\widehat{e}(L_i)} \left\{ Y_i-\widehat{m}_0(L_i) \right\}.
\label{eq:phi_score}
\end{equation}

The nuisance functions $\widehat{m}_0$ and $\widehat{e}$ are estimated using cross-fitting within the held-out evaluation cohort. Finally, as defined in Equation \ref{eq:policy_utility}, the policy utility for this overlap-restricted population is reported as:
\begin{equation}
\widehat{U}(\pi_q) = 1-\widehat{R}_{DR,A}(\pi_q).
\label{eq:policy_utility}
\end{equation}


In our framework, the non-eligible individuals are not directly selected for intervention and are not included in the eligible-cluster benefit average used for subgroup ranking. However, they remain a critical component of this framework. First, they are used to estimate the lower-risk outcome model, which is needed for the model-based benefit scores during the discovery phase. Second, they contribute to the selective doubly robust policy evaluation, where individuals already observed in the lower-risk state provide information about the target-state outcome distribution and enter the propensity-corrected residual component of the value estimator. Synthesizing the design-stage causal setup, the selective shift mechanism, and the doubly robust evaluation, the pipeline for our budgeted policy framework is summarized in Algorithm \ref{alg:generic_subgroup_policy}.

\begin{algorithm*}[!h]
\caption{Unsupervised Subgroup Discovery, Policy Construction, and Evaluation}
\scriptsize
\label{alg:generic_subgroup_policy}

\KwIn{
Observed data $\mathcal{D}=\{(X_i,T_i,Y_i)\}_{i=1}^n$;
pre-treatment covariates $X$;
adjustment covariates $Z$;
treatment/state variable $T$;
outcome $Y$;
clustering algorithm $\mathcal{A}$;
number of clusters or upper truncation level $K$;
policy budget $q$;
safety gate $G \in \{\text{None}, \text{Empirical Bernstein}, \text{Bayesian pooling}\}$;
random seed $s$.
}

\KwOut{
Subgroup policy $\pi_q$;
selected clusters $\mathcal{S}_q$;
policy risk $\widehat{R}(\pi_q)$;
policy utility $\widehat{U}(\pi_q)$.
}

\textbf{Step 1: Covariate selection.} \\
Use causal discovery and domain knowledge to identify baseline covariates, adjustment variables, and candidate effect modifiers. Exclude the treatment/state variable $T$ and outcome $Y$ from the clustering variables. Let $Z$ denote the selected adjustment set used in the nuisance outcome and propensity models.

\textbf{Step 2: Discovery/evaluation split.} \\
Split $\mathcal{D}$ into discovery data $\mathcal{D}_{disc}$ and evaluation data $\mathcal{D}_{eval}$ using seed $s$. The discovery cohort is used to estimate benefit scores, learn subgroup structure, and rank subgroups. The evaluation cohort is reserved for held-out policy application and evaluation.

\textbf{Step 3: Estimate individual-level model-based benefit scores.} \\
Using only the discovery cohort, fit outcome models for the adverse state $T=1$ and the lower-risk state $T=0$ using the selected adjustment set $Z$:
\[
\widehat{m}_t(Z_i)
=
\widehat{\mathbb{E}}(Y_i \mid T_i=t,Z_i),
\quad t \in \{0,1\}.
\]
For each discovery individual, compute the estimated one-way state-shift benefit
\[
\widehat{b}_i
=
\widehat{m}_1(Z_i)-\widehat{m}_0(Z_i),
\quad i \in \mathcal{D}_{disc}.
\]
Larger values of $\widehat{b}_i$ indicate greater predicted reduction in adverse-outcome risk under the hypothetical shift $T=1 \rightarrow T=0$. These discovery benefit scores are obtained using out-of-fold predictions to reduce overfitting. These scores are model-based estimated benefits and are not doubly robust pseudo-outcomes.

\textbf{Step 4: Fit the unsupervised subgroup model.} \\
Fit preprocessing transformations, such as scaling, using only $X_{disc}$. Fit the clustering algorithm $\mathcal{A}$ on the transformed pre-treatment covariates in $\mathcal{D}_{disc}$. The treatment $T$, outcome $Y$, and benefit scores $\widehat{b}_i$ are not used to fit the clustering model.

\textbf{Step 5: Assign subgroup labels.} \\
Use the fitted preprocessing and clustering model to assign subgroup information to both discovery and evaluation individuals. 

\textbf{Step 6: Aggregate benefits to eligible-subgroup summaries.} \\
Because the hypothetical intervention is meaningful only for individuals currently in the adverse or eligible state, define the discovery eligible set as


$\mathcal{E}_{disc}=\{i \in \mathcal{D}_{disc}: T_i=1\}$.
For hard cluster assignments, define the eligible-cluster average benefit as

\[
\widehat{\tau}^{\mathrm{hard}}_c
=
\frac{1}{n_{c,elig}}
\sum_{i \in \mathcal{E}_{disc}: C_i=c}
\widehat{b}_i,
\]
where $n_{c,elig}\equiv n_c^{\mathrm{hard}} = \sum_{i \in \mathcal{D}_{disc}} \mathbf{1}(C_i=c,T_i=1)$.
Define the total eligible gain for cluster $c$ as:

\[\mathrm{Gain}^{\mathrm{hard}}_c
=
n_c^{\mathrm{hard}}
\widehat{\tau}_c^{\mathrm{hard}}\]

For membership-weighted FCM summaries, define the eligible membership weights as
\begin{equation}
    w_{ic} = u_{ic}^{m}\mathbf{1}(T_i=1),
\end{equation}
where $u_{ic}$ is the membership of individual $i$ in cluster $c$, and $m$ is the FCM fuzzifier. The membership-weighted eligible-cluster benefit is
\begin{equation}
    \widehat{\tau}_c^{\mathrm{soft}} = \frac{\sum_{i \in \mathcal{D}_{\mathrm{disc}}} u_{ic}^{m}\mathbf{1}(T_i=1)\widehat{b}_i}{\sum_{i \in \mathcal{D}_{\mathrm{disc}}} u_{ic}^{m}\mathbf{1}(T_i=1)}.
\end{equation}

For uncertainty estimation, define the effective eligible sample size as
\begin{equation}
    n^{\mathrm{eff}}_{c,\mathrm{elig}} = \frac{\left( \sum_{i \in \mathcal{D}_{\mathrm{disc}}} w_{ic} \right)^2}{\sum_{i \in \mathcal{D}_{\mathrm{disc}}} w_{ic}^{2}}.
\end{equation}
The weighted variance, standard error, and $n^{\mathrm{eff}}_{c,\mathrm{elig}}$ are used to compute the soft-membership uncertainty summaries. These subgroup summaries are used only for ranking and safety gating.

For stochastic FCM, the same membership-weighted discovery-cohort summaries are used to determine and freeze the subgroup ranking and safety-gate decisions, while stochasticity is introduced only during evaluation through repeated sampling of hard cluster assignments from the learned fuzzy memberships.

\textbf{Step 7: Apply the safety gate and rank subgroups.} \\
For the ungated policy, rank subgroups using the method-specific discovery-side gain defined for the corresponding clustering approach.  If (G=\text{Empirical Bernstein}) or (G=\text{Bayesian pooling}), retain only subgroups that satisfy the corresponding admission criterion and rank the admitted subgroups using the associated uncertainty-adjusted gain. Freeze the resulting subgroup order before evaluation.

\textbf{Step 8: Construct the budgeted subgroup policy on the held-out evaluation cohort.} \\
Assign individuals in $\mathcal{D}_{\mathrm{eval}}$ to subgroups using the clustering model and subgroup ranking frozen from the discovery cohort. Allocate the budget only among eligible evaluation individuals with $T_i=1$. Let
\[
B_{\mathrm{eval}} = \operatorname{round} \left( q N_{\mathrm{elig}, \mathrm{eval}} \right),
\]
where $N_{\mathrm{elig}, \mathrm{eval}}$ is the number of eligible individuals in the evaluation cohort. Select eligible evaluation individuals from full clusters in the discovery-frozen ranked order until $B_{\mathrm{eval}}$ individuals have been selected. If the next cluster exceeds the remaining budget, randomly sample the required number of eligible individuals from the boundary cluster $\mathcal{B}_q$ using seed $s$, without using individual-level benefit rankings. To preserve the subgroup-level structure, convert this targeting decision into a cluster-level shift probability:
\[
s_c = \frac{\text{number of selected eligible evaluation individuals in cluster } c}{\text{total eligible evaluation individuals in cluster } c}.
\]
Fully selected clusters have $s_c=1$, unselected clusters have $s_c=0$, and the boundary cluster has $0 < s_c < 1$.

Also, unlike the ungated policy, which exhausts the prespecified budget, a safety-gated policy may leave part of the budget unused when the admitted subgroups contain fewer than \(B_{\mathrm{eval}}\) eligible individuals.

\textbf{Step 9: Evaluate the resulting policy on held-out data.} \\
Using the cluster-level shift probabilities $s_c$ constructed in $\mathcal{D}_{\mathrm{eval}}$, estimate the policy risk and policy utility. 
\Return{$\pi_q, \mathcal{S}_q, \widehat{R}(\pi_q), \widehat{U}(\pi_q)$}

\end{algorithm*}

\subsection{Comparison with a supervised subgrouping approach}

A treatment-effect-informed subgrouping comparator is also evaluated. A causal forest is fitted on the discovery cohort, and a shallow CATE-interpreting tree partitions the covariate space into a limited number of leaves. These leaves are treated analogously to clustering-derived subgroups and are evaluated using the same subgroup-level budget and held-out policy-value estimator. The comparator is included to assess whether effect-guided subgroup construction enhances policy performance compared to phenotype-based clustering. Evaluation is conducted only under the ungated allocation rule. For the NHANES smoking experiment, the tree employs the same MCA-derived effect-modifier representation as the unsupervised methods.

Since supervised subgrouping incorporates treatment and outcome information when constructing leaves, it is more susceptible to data-adaptive subgroup discovery \cite{wang2007statistics} than purely phenotype-based clustering. To mitigate this risk, the supervised CATE-tree comparator was implemented with safeguards against subgroup hunting. The CATE model was fit exclusively on the discovery cohort, and leaf-level rankings were based on out-of-fold CATE estimates, ensuring that each discovery observation received a treatment-effect prediction from a model not trained on that observation. The CATE-interpreting tree was constrained to be shallow, with a limited maximum number of leaves and a minimum leaf-size requirement, thereby restricting the formation of highly specific subgroups around noise. Once the supervised leaves and their discovery-side ranking were established, the policy was fixed and applied without re-ranking to the held-out evaluation cohort.

Several structural differences should be considered when interpreting this comparison. While matching the number of supervised leaves to the number of unsupervised clusters controls for subgroup count, it does not ensure that the partitions are equally expressive. A shallow CATE-interpreting tree forms axis-aligned rectangular regions \cite{HTE_P3}, whereas centroid-based methods such as $K$-means generate Voronoi partitions based on Euclidean proximity and may capture distinct geometric structures in the covariate space. Furthermore, the supervised tree is explicitly optimized to separate estimated treatment effects, whereas unsupervised clusters are optimized solely for covariate similarity.

\subsection{Paired comparisons of held-out policy risk}\label{paired_compares}
.

To directly compare methods evaluated on the same held-out individuals, we conducted paired nonparametric bootstrap analyses of all unique pairwise policy-risk differences. Within each bootstrap replicate, all candidate policies were evaluated on the same resampled evaluation cohort, thereby preserving the dependence among their policy-risk estimates. Two-sided centered-bootstrap p-values were calculated for each pairwise contrast \cite{hall1991two}. The centered-bootstrap p-values reported in the pairwise comparison tables in the supplementary files are the unadjusted values and are provided as descriptive inferential summaries. For multiplicity control, these p-values were jointly adjusted using Holm’s procedure \cite{holm1979simple,abdi2010holm,aickin1996adjusting}, and statistical significance was assessed using the corresponding Holm-adjusted p-values. Within each policy experiment, Holm adjustment was applied across the complete set of unique pairwise comparisons, including comparisons among the unsupervised subgroup policies and comparisons between the supervised CATE-tree policy and the unsupervised policies.

\subsection{Sensitivity Analysis}

Sensitivity to the discovery and evaluation partition was assessed by repeating the analysis across multiple random split seeds. For each seed, subgroups and policy rankings were derived exclusively from the discovery cohort and subsequently applied, without refitting, to the corresponding held-out evaluation cohort. Policy risk and utility were re-estimated using the same doubly robust evaluation procedure.

\section*{Experimental Setup}
\subsection{Evaluation Analyses}
\par 
We set the intervention budget at q = 0.70 to serve as a standardized capacity constraint, facilitating controlled comparisons across subgrouping methods rather than reflecting a clinically optimized threshold. Consistent with capacity-constrained policy-learning frameworks, the budget is treated as an externally specified limit on the proportion of eligible individuals who may be prioritized \cite{kitagawa2018treated}. In practice, the budget value would be determined by resource availability, intervention costs, operational capacity, and, when applicable, equity considerations.

Within this standardized budget, we perform six complementary evaluations. First, we compare held-out policy risk and utility across unsupervised subgrouping methods and the supervised CATE-tree comparator. Second, we assess the impact of Empirical Bernstein and hierarchical Bayesian gating on subgroup admission, the number of individuals targeted, and policy utility. Third, we quantify policy-value variation resulting from stochastic FCM hardening. Fourth, we repeat the evaluation framework across multiple discovery and evaluation split seeds. Fifth, we compare the individuals selected by different policies using Jaccard similarity, targeted-set agreement, and Reference-allocation coverage \cite{jaccard1901etude,powers2020evaluation}.
Lastly, we directly quantified uncertainty in policy-risk differences using paired bootstrap comparisons on the same held-out individuals, with multiplicity control using Holm's procedure.

\subsection{Discovery/Evaluation Split}

A discovery--evaluation split is implemented to reduce optimistic assessment
of data-adaptively learned subgroup policies. All preprocessing parameters,
subgroup definitions, benefit estimates, subgroup rankings, and safety-gate
decisions are learned exclusively from the discovery cohort and then held
fixed. The evaluation cohort is used only for out-of-sample subgroup
assignment and doubly robust policy evaluation. This design separates policy
construction from performance assessment and provides a common held-out
evaluation protocol for the unsupervised subgrouping methods and the supervised
CATE-tree comparator. Importantly, the split does not alter the unsupervised
nature of subgroup discovery, because the clusters are constructed without
using the exposure state, outcome, estimated benefit scores, or subsequent
state-shift decisions.

\begin{figure}
    \centering
    \includegraphics[width=14cm,height=8cm]{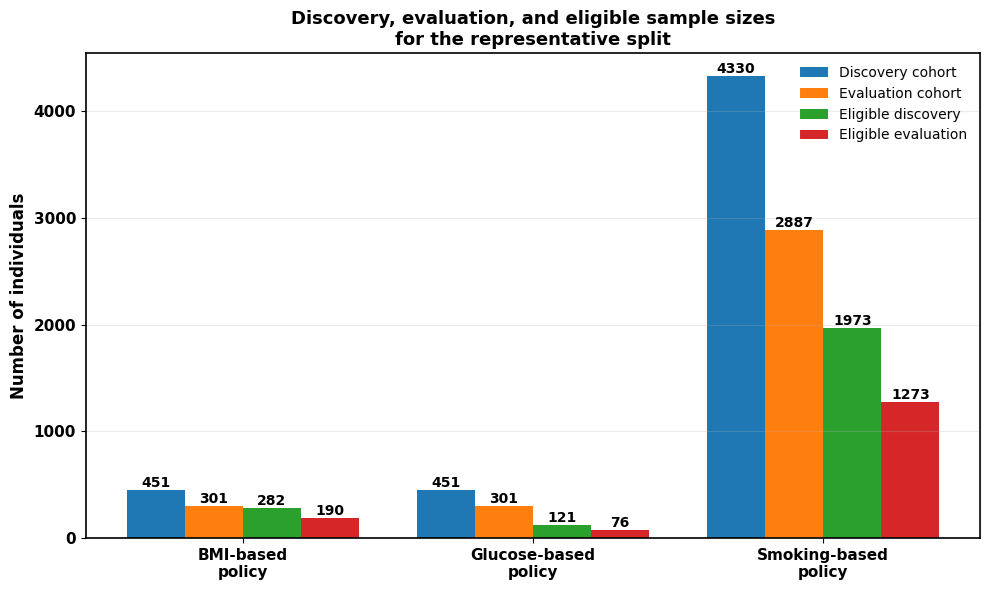}
\caption{Discovery and evaluation sample sizes for the representative
discovery--evaluation split used in the main analysis. Eligible discovery and
eligible evaluation counts denote individuals in the adverse or modifiable
state for the corresponding hypothetical shift. Eligible counts may vary
slightly across alternative split seeds.}

\label{tab:policy_split_eligibility}

\end{figure}

This design is consistent with the cluster-validation literature, which
supports assessing learned cluster structures using either a held-out portion
of the original sample or an independently collected validation dataset
\cite{ullmann2022validation}. Prior work on data-driven subgroup identification
has similarly emphasized the need for validation, resampling, or bias-correction
procedures \cite{foster2011subgroup}. The resulting split sizes and
policy-specific eligibility counts are reported in
Figure ~\ref{tab:policy_split_eligibility}.

\subsection{Computational Setup, Hyperparameters and Evaluation Metrics}

All analyses were conducted using Python. K-means and Bayesian Gaussian
mixture clustering, silhouette score calculations, logistic regression
nuisance models, data splitting, and cross-fitting utilities were implemented
using scikit-learn \cite{scikit}. Fuzzy C-means fitting and out-of-sample
fuzzy-membership prediction were implemented using scikit-fuzzy
\cite{scikitfuzzy}. Multiple correspondence analysis for the NHANES
categorical covariates was performed using Prince \cite{prince}. The supervised
CATE-tree comparator was implemented using EconML \cite{econml}. Hierarchical
Bayesian pooling was implemented using PyMC \cite{pymc}. For causal discovery, we used the CausalLearn \cite{causallearnpackage} package to implement constraint-based structure learning, including the PC and FCI algorithms with the kernel conditional independence (KCI) test, as well as nonlinear NOTEARS. For settings with mixed continuous covariates and a binary outcome, we computed Pillai’s trace–based dependence tests using the pgmpy library \cite{pgmpy}.  The different hyperparameters used in the study are tabulated in Tables \ref{tab:clustering_hyperparameters} and \ref{tab:additional_hyperparameters}.

\begin{table}[!h]
\centering
\scriptsize
\caption{Subgrouping algorithms and hyperparameters used for each policy
experiment. A common reference number of subgroups, $K$, was selected by
maximizing the K-means silhouette score on the discovery cohort only and was
then held fixed across the primary unsupervised comparisons. The selected
silhouette scores were $0.3980$ for BMI, $0.3214$ for glucose, and $0.2405$
for smoking. The supervised CATE-tree comparator was restricted to at most the corresponding
policy-specific $K$ leaves to provide subgroup granularity comparable with the
unsupervised methods}
\label{tab:clustering_hyperparameters}
\begin{tabular}{lll}
\toprule
Policy & Algorithm & Hyperparameters \\
\midrule
BMI & K-means & $K=3$; k-means++ initialization; $n_{\mathrm{init}}=50$ \\ 
& FCM & $K=3$; fuzziness $m=1.7$; 50 random starts; best objective retained \\ 
& Bayesian GMM & $K_{\max}=3$; full covariance; $n_{\mathrm{init}}=50$; finite Dirichlet prior on mixture weights \\ 

\midrule
Glucose 
& K-means 
& $K=4$ clusters; k-means++ initialization; $n_{\mathrm{init}}=50$ \\ 
& FCM & $K=4$; fuzziness $m=1.7$; 50 random starts; best objective retained \\ 
& Bayesian GMM & $K_{\max}=4$; full covariance; $n_{\mathrm{init}}=50$;finite Dirichlet prior on mixture weights

\\
\midrule
Smoking 
& K-means 
& $K=6$ clusters; k-means++ initialization; $n_{\mathrm{init}}=50$ \\ 
& FCM & $K=6$; fuzziness $m=1.7$; 50 random starts; best objective retained \\ 
& Bayesian GMM & $K_{\max}=6$; full covariance; $n_{\mathrm{init}}=50$; finite Dirichlet prior on mixture weights \\
\midrule
All policies
& Supervised CATE-tree
& Causal forest: 400 trees, minimum leaf size 10,minimum split size 20, 3-fold internal cross-fitting\\

\bottomrule
\end{tabular}

\noindent \textit{Note.} 
To establish a standardized comparison across methods, the optimal $K$ was determined via K-means on the discovery set and subsequently applied to FCM. To mitigate convergence to local minima, all algorithms were executed with 50 random initializations, retaining the model with the optimal objective function value. For Bayesian GMM, the maximum number of components ($K_{\max}$) was constrained to match this $K$. While this standardizes the subgroup-policy evaluation, it limits the model's inherent ability to dynamically select an alternative number of mixture components that might better fit the data. For the Smoking-history policy,  seven MCA dimensions were retained
using the Greenacre-adjusted component-selection criterion.

\end{table}

\begin{table*}[!h]
\centering
\scriptsize
\caption{Additional hyperparameters used in the policy-learning and evaluation framework.}
\label{tab:additional_hyperparameters}
\begin{tabular}{lc}
\toprule
Additional hyperparameter & Value \\
\midrule
Total EB error level, $\delta$ & $0.05$ \\
Number of policy-evaluation cross-fitting folds & $5$ \\
Number of bootstrap replicates & $500$ \\
Number of bootstrap replicates for paired bootstrap test & $500$\\
Number of stochastic-hardening draws & $500$
\\
\bottomrule
\end{tabular}
\end{table*}


Because ground-truth individual treatment effects and optimal treatment assignments are unavailable, we evaluate policies using the held-out doubly robust policy risk and utility defined in Equations~\ref{eq:dr_estimator} and~\ref{eq:policy_utility}, rather than oracle metrics such as PEHE or accuracy relative to a true optimal policy. These quantities are averaged over the full overlap-restricted evaluation population, although intervention eligibility and budget allocation are restricted to individuals in the adverse state. We additionally report pairwise allocation-concordance metrics to determine whether different policies select the same eligible individuals. These agreement metrics are descriptive and are not interpreted as accuracy relative to an unknown optimal allocation.

\begin{itemize}

\item Jaccard index \cite{jaccard}: Let $(S_{\pi})$ denote the set of evaluation-set individuals selected by policy $(\pi)$, and let $(S_{\mathrm{ref}})$ denote the set selected by a reference policy, such as a deterministic FCM policy. Jaccard is computed using the equation \ref{eq:jaccard_policy_overlap}. 
  
\begin{equation}
J(\pi,\pi_{\mathrm{ref}})
=
\frac{
\left|S_{\pi}\cap S_{\mathrm{ref}}\right|
}{
\left|S_{\pi}\cup S_{\mathrm{ref}}\right|
}.
\label{eq:jaccard_policy_overlap}
\end{equation}

\item Targeted-set agreement is computed using the equation \ref{eq:targeted_set_agreement}. 
\begin{equation}
A_{\mathrm{target}}(\pi,\pi_{\mathrm{ref}})
=
\frac{
\left|S_{\pi}\cap S_{\mathrm{ref}}\right|
}{
\left|S_{\pi}\right|
}.
\label{eq:targeted_set_agreement}
\end{equation}

\item Reference-allocation coverage is computed using the equation \ref{eq:reference_policy_coverage}. 
\begin{equation}
C_{\mathrm{ref}}(\pi,\pi_{\mathrm{ref}})
=
\frac{
\left|S_{\pi}\cap S_{\mathrm{ref}}\right|
}{
\left|S_{\mathrm{ref}}\right|
}.
\label{eq:reference_policy_coverage}
\end{equation}

\end{itemize}

In addition to these, for stochastic FCM policies we compute MC summaries across repeated membership hardenings. We report MC variability separately from bootstrap 95\% confidence intervals, which quantify held-out sampling uncertainty in the doubly robust policy-value estimates.

\section*{Results}
This section presents the empirical results of the proposed framework. First, the potential causal links identified through the data-driven causal structure discovery procedure and the resulting domain-informed graph used for covariate selection and adjustment are summarized. Next, the policy risk and utility of the unsupervised subgrouping methods are compared under the proposed hypothetical state intervention. These unsupervised subgroup policies are also evaluated against the supervised subgrouping baseline, which is based on treatment-effect-informed leaves. Finally, a sensitivity analysis is conducted to report how the estimated policy risk and utility vary across different discovery and evaluation split seeds. Detailed subgroup-level allocation tables, ranking diagnostics, gating summaries, overlap diagnostics, and additional effect-homogeneity results are provided in the Supplementary Files.

\subsection{Potential Causal Discovery}
\begin{figure}
    \centering
    \includegraphics[width=14cm,height=8cm]{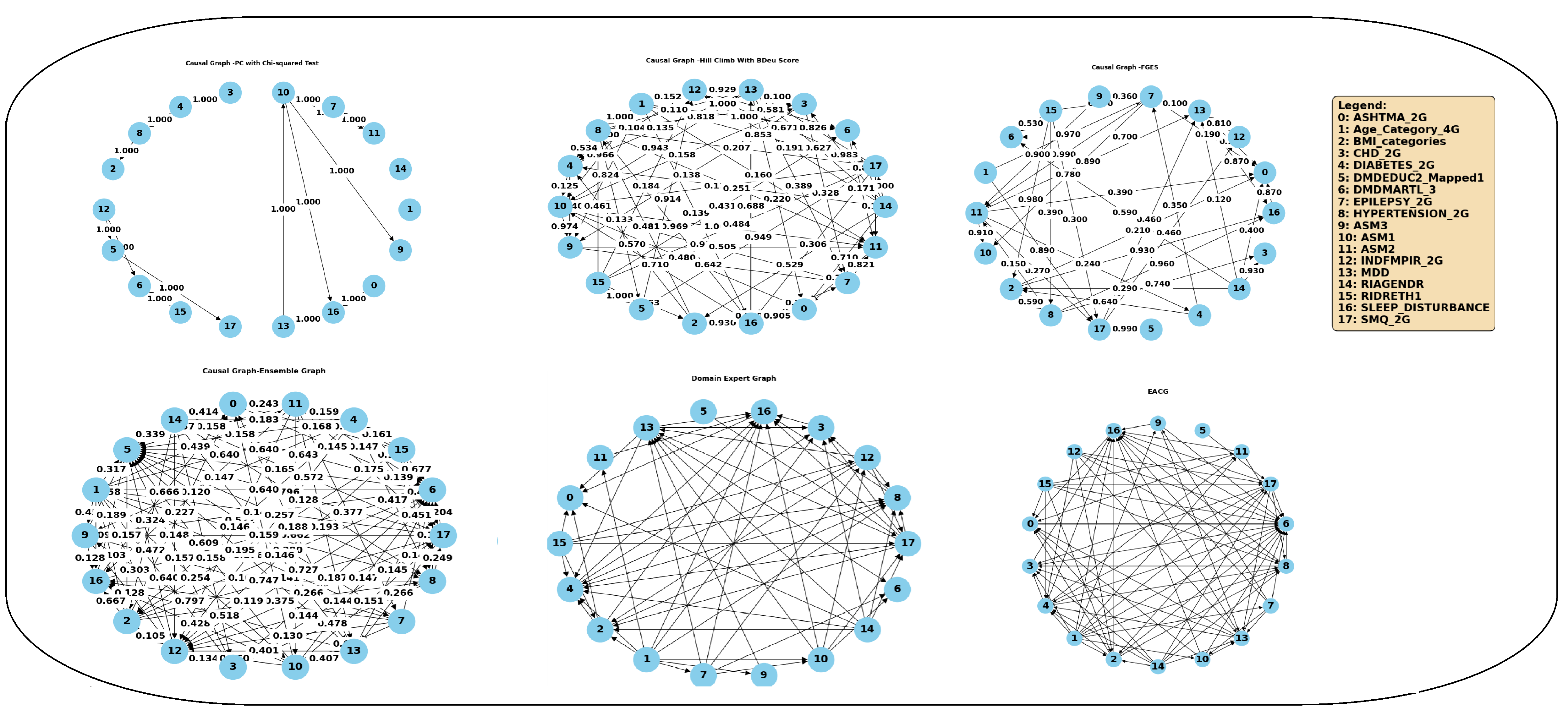}
 \caption{NHANES: Representative outputs of the ensemble causal-structure discovery procedure used for pretreatment covariate selection. The figure is included to illustrate the causal-structure discovery workflow used for pretreatment covariate selection in the present policy-learning pipeline. 
 Representative PC, hill-climbing (BDeu), and FGES graphs compared to our preliminary weighted ensemble (bottom-left). Final Expert-Augmented Causal Graph (EACG) is shown in the bottom-right.}

\label{CD_epilepsy}

\end{figure}

\begin{figure*}[!h]

    \centering
   \includegraphics[width=14cm, height=6cm]{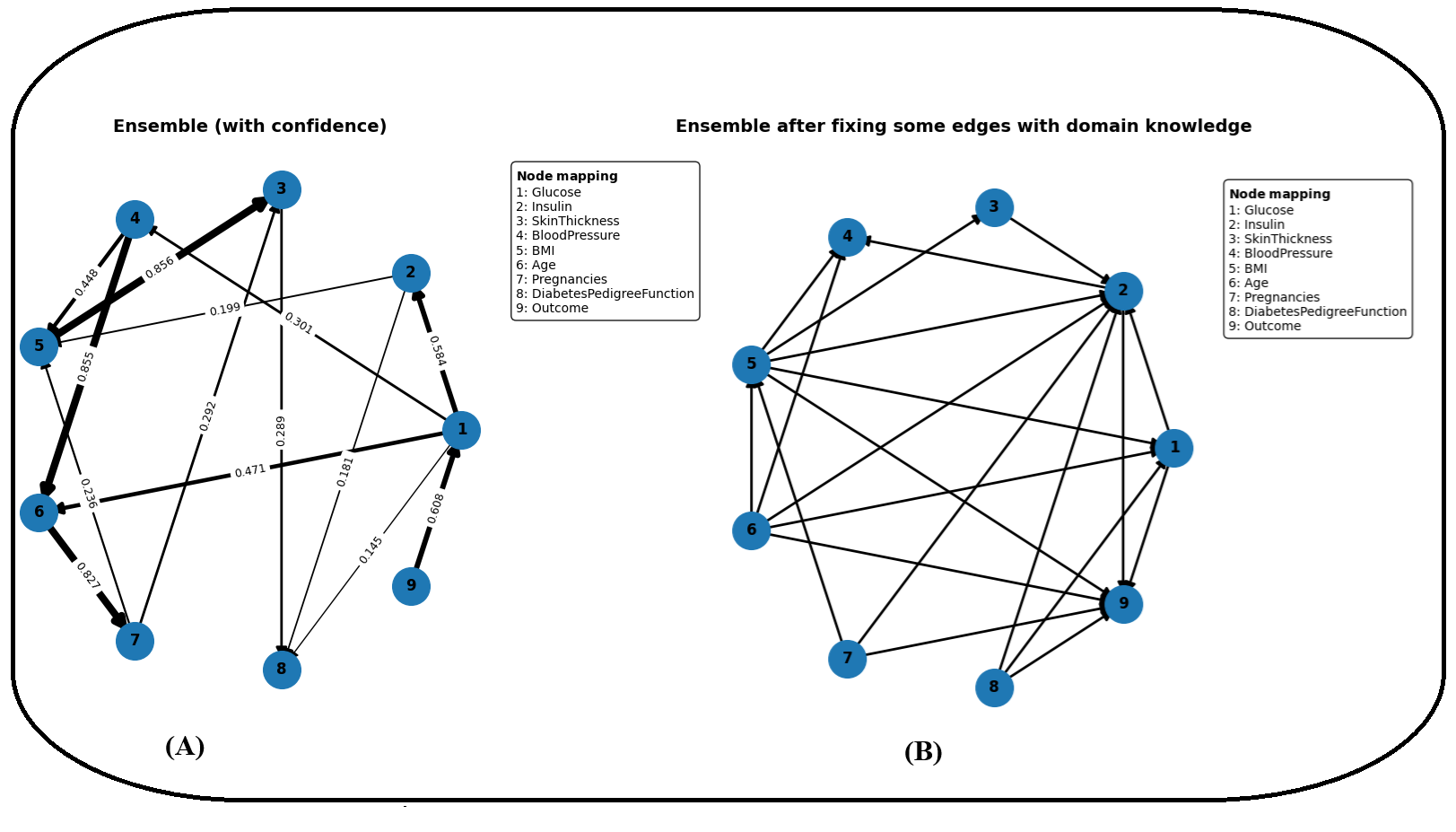}
    \caption{PIMA:
(A) Weighted ensemble causal graph learned from observational data , with edge labels denoting composite confidence. (B) Constraint-corrected ensemble graph obtained by post-processing (A) to enforce basic clinical ordering constraints (such as Age has no parents and Outcome has no children). The ensemble recovers several clinically plausible relations (e.g., $\mathrm{BMI}\!\rightarrow\!\mathrm{SkinThickness}$ and $\mathrm{Glucose}\!\rightarrow\!\mathrm{Insulin}$), but some directions disagree with the reference graph, which lowers the orientation- and path-based components of the score. These discrepancies are expected because the graphs in (A) are learned without injecting domain constraints during discovery.}
\label{CD_PIMA}
\end{figure*}

The ensemble causal-structure discovery procedure used for covariate selection was previously described and empirically evaluated in our previous work \cite{acharya2025understanding}. We reuse this procedure, together with domain knowledge, only to identify plausible pretreatment variables for subgroup construction and policy evaluation. Causal-structure discovery is not treated as a primary methodological contribution of the present study. The figure \ref{CD_epilepsy} shows the graph learnt from the NHANES data. The figure \ref{CD_PIMA} shows the graph learnt from the PIMA data and the graph after post processing it with domain knowledge. We may notice that some of the plausible biomedical relationships are missed or reversed in direction by the graph discovered from data. We attribute this to the small size of the dataset.

\newpage

\subsection{Policy Utility under the Glucose-Based Hypothetical State-Intervention Setting}

\begin{figure*}[!h]
    \centering
    \includegraphics[
        width=0.94\textwidth
    ]{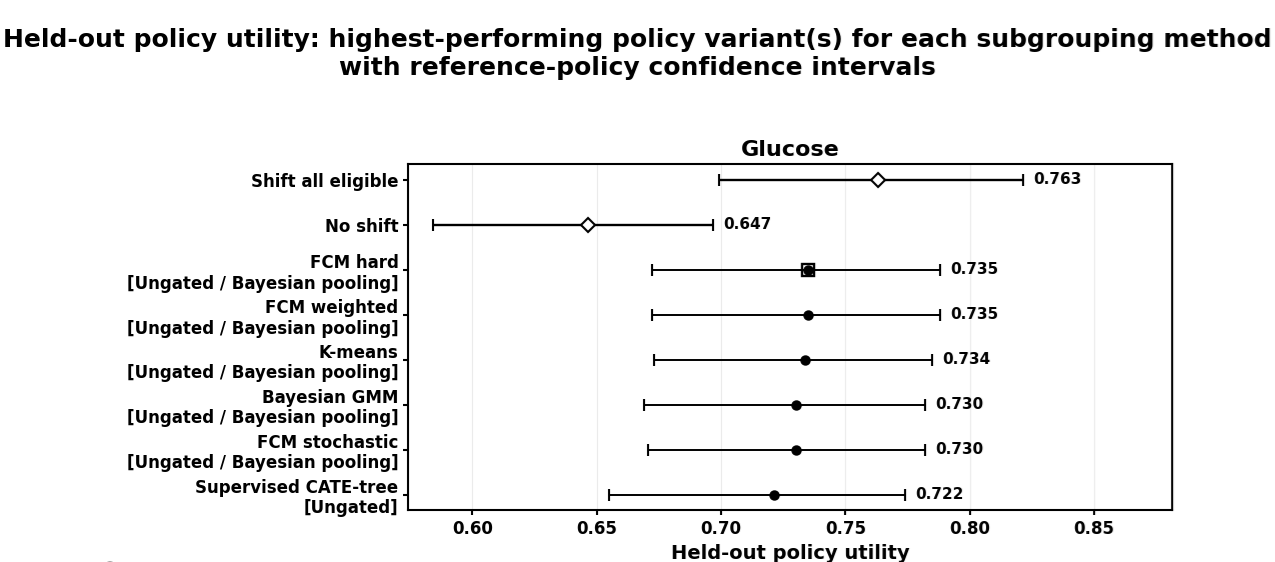}
    \caption{\textbf{Glucose Policy: Summary of representative-split held-out
    policy utility.}
    For each subgrouping method, the highest observed
    held-out utility in the representative discovery--evaluation split is
    displayed. Points denote estimated policy utility, and horizontal bars
    denote bootstrap 95\% confidence intervals. Open diamonds denote the
    no-shift and shift-all-eligible reference policies. An open square
    identifies the highest point estimate among the budget-constrained
    subgroup policies within each experiment. The policy with the highest utility point estimate was achieved by FCM Hard and FCM Weighted.}
    \label{fig:representative_policy_utility_summary_glucose_image}
\end{figure*}

Table~\ref{tab:glucose_policy_results_seed70} reports the representative-split results for the glucose policy. Among the ungated unsupervised policies, held-out utility ranged from (0.7302) to (0.7350), compared with (0.7215) for the supervised CATE-tree comparator. The bootstrap confidence intervals overlapped substantially. Empirical Bernstein gating admitted no clusters and produced a common no-shift utility of (0.6467), whereas Bayesian pooling reproduced the corresponding ungated allocations. Figure~\ref{fig:representative_policy_utility_summary_glucose_image} provides a
visual summary of the representative-split held-out policy
utilities. For stochastic FCM, Table~\ref{tab:stochastic_fcm_glucose_uncertainty_summary} shows that the MC standard deviation of utility was small relative to the bootstrap interval. The EB-gated policy had no MC variation because no cluster passed the gate. The cluster profiles in Table~\ref{tab:glucose_cluster_profiles} were qualitatively consistent between the discovery and evaluation cohorts. Under the ungated allocations, the methods nevertheless prioritized different combinations of younger, older, higher-BMI, and lower-\textit{DiabetesPedigreeFunction} phenotypes.

\begin{figure*}[!h]
\centering
\includegraphics[
    width=\linewidth
]{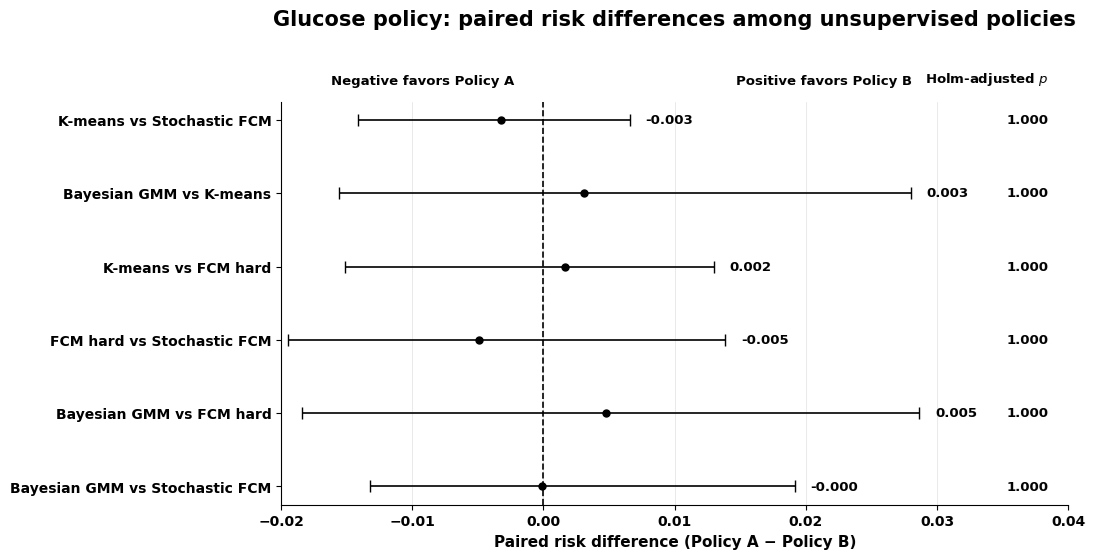}
\caption{
\textbf{Glucose policy:} Paired bootstrap comparisons of held-out
full-population policy risk. The dashed line
denotes zero difference. Holm-adjusted \(p\)-values are shown in
the right-hand column. All confidence intervals included zero,
and no comparison remained statistically detectable after Holm
adjustment.
}
\label{pairwise_bootstrap_glucose_figure}
\end{figure*}

The allocation comparisons tabulated in Table~\ref{tab:seed70_hardened_policy_overlap} provide a complementary result. Hard and membership-weighted FCM selected identical eligible individuals, and their Jaccard overlap with K-means was (0.860). Comparisons involving Bayesian GMM had a lower Jaccard overlap of (0.582). Stochastic FCM followed the same general pattern across its 500 hardening runs as shown in Table~\ref{tab:seed70_stochastic_fcm_overlap}. Thus, the small differences in estimated utility were accompanied by more substantial differences in allocation composition. The overlap metrics summarize realized targeting among eligible individuals, whereas the doubly robust estimator evaluates the induced subgroup-level shift probabilities. The two summaries therefore measure related but distinct aspects of the policies.
The unsupervised glucose-policy comparisons are visualized in
Figure~\ref{pairwise_bootstrap_glucose_figure}. Statistical significance was assessed using Holm-adjusted p-values.

\begin{table}[!h]
\centering
\scriptsize
\caption{\textbf{Glucose policy:} Policy risk and utility. The selected cluster order denotes the discovery-cohort ranking used to allocate the held-out intervention budget. The no-shift and shift-all-eligible rows provide common reference policies evaluated using the same held-out doubly robust estimator. The ungated and Bayesian-pooling allocations achieved similar held-out utilities across the unsupervised subgrouping methods. The EB safety gate was substantially more conservative and reduced all methods to the no-shift reference allocation. The FCM hard and FCM weighted achieved the highest utility point estimate of $0.7350$, compared with $0.6467$ under no shift and $0.7632$ under shifting all eligible individuals. The supervised CATE-tree comparator did not show a clear utility advantage in this representative split.}
\label{tab:glucose_policy_results_seed70}
\begin{tabular}{llcccc}
\toprule
Base method
& Policy variant
& Policy risk
& Policy utility
& 95\% CI for utility
& Selected cluster order \\
\midrule

Reference policy
& No shift &

0.3533 
& 0.6467 
& [0.5842, 0.6969] & --\\

Reference policy
& Shift all eligible
& 0.2368
& 0.7632
& [0.6992, 0.8215]
& -- \\

\midrule

K-means 
& Ungated 
& 0.26660 
& 0.7334 
& [0.6730, 0.7850] 
& 2, 0, 3 \\

K-means 
& EB safety gate 
& 0.3533 
& 0.6467 
& [0.5842, 0.6969] 
& None \\

K-means 
& Bayesian pooling 
& 0.26660
& 0.7334
& [0.6730, 0.7850] 
& 2, 0, 3 \\

\midrule

FCM hard 
& Ungated 
& \textbf{0.2649} 
& \textbf{0.7350} 
& \textbf{[0.6722, 0.7881]} 
& 2, 1, 0 \\

FCM hard 
& EB safety gate 
& 0.3533 
& 0.6467 
& [0.5842, 0.6969] 
& None \\

FCM hard 
& Bayesian pooling 
& 0.2649 
& 0.7350 
& [0.6722, 0.7881] 
& 2, 1, 0 \\

\midrule

FCM weighted 
& Ungated 
& \textbf{0.2649} 
& \textbf{0.7350} 
& \textbf{[0.6722, 0.7881]} 
& 2, 1, 0 \\

FCM weighted 
& EB safety gate 
& 0.3533 
& 0.6467 
& [0.5842, 0.6969] 
& None \\

FCM weighted 
& Bayesian pooling
& 0.2649 
& 0.7350 
& [0.6722, 0.7881] 
& 2, 1, 0 \\

\midrule

FCM stochastic 
& Ungated 
& 0.2698 
& 0.7302
& [0.6708, 0.7821] 
& 2, 1, 0 \\

FCM stochastic 
& EB safety gate 
& 0.3533 
& 0.6467 
& [0.5842, 0.6969] 
& None \\

FCM stochastic 
& Bayesian pooling
& 0.2698 
& 0.7302 
& [0.6708, 0.7821] 
& 2, 1, 0 \\

\midrule

Bayesian GMM 
& Ungated 
& 0.2697 
& 0.7303
& [0.6690, 0.7821] 
& 0, 1, 2 \\

Bayesian GMM 
& EB safety gate
& 0.3533 
& 0.6467 
& [0.5842, 0.6969] 
& None \\

Bayesian GMM 
& Bayesian pooling
& 0.2697 
& 0.7303
& [0.6690, 0.7821] 
& 0, 1, 2 \\

\midrule

Supervised CATE-tree comparator
& Ungated 
& 0.2785 
& 0.7215 
& [0.6549, 0.7741] 
& -\\

\bottomrule
\end{tabular}

\begin{flushleft}
\scriptsize
\textit{Note:} For the stochastic FCM rows, policy risk, policy utility, and confidence intervals summarize the expected policy value over MC stochastic hardening runs.
\end{flushleft}
\end{table}

\begin{table}[!h]
\centering
\scriptsize
\caption{\textbf{Glucose Policy:} Policy-value uncertainty summary for stochastic FCM. Stochastic hardening introduced only modest MC variability for the ungated and Bayesian-gated FCM policies, while the EB-gated policy showed no MC variation because the gate selected no clusters and therefore reduced the allocation to the same no-shift rule in every run.}
\label{tab:stochastic_fcm_glucose_uncertainty_summary}
\begin{tabular}{llcccccc}
\toprule
Policy variant
& Metric
& MC mean
& MC SD
& MC 2.5\%
& MC 97.5\%
& Bootstrap 2.5\%
& Bootstrap 97.5\% \\
\midrule

Ungated stochastic FCM
& Risk
& \textbf{0.269849}
& 0.006148
& 0.257970
& 0.281798
& 0.217948
& 0.329210 \\

Ungated stochastic FCM
& Utility
& \textbf{0.730151}
& 0.006148
& 0.718202
& 0.742030
& 0.670790
& 0.782052 \\

EB-gated stochastic FCM
& Risk
& 0.3533
& 0.000000
& 0.3533
& 0.3533
& 0.3031
& 0.4158 \\

EB-gated stochastic FCM
& Utility
& 0.6467
& 0.000000
& 0.6467
& 0.6467
& 0.5842
& 0.6969 \\

Bayesian-gated stochastic FCM
& Risk
& \textbf{0.269849}
& 0.006148
& 0.257970
& 0.281798
& 0.217948
& 0.329210 \\

Bayesian-gated stochastic FCM
& Utility
& \textbf{0.730151}
& 0.006148
& 0.718202
& 0.742030
& 0.670790
& 0.782052 \\
\bottomrule

\end{tabular}
\end{table}

\begin{table}[!h]
\centering
\scriptsize
\caption{\textbf{Glucose policy:} Coarse cluster-profile summaries based on the pre-treatment features selected through ensemble causal discovery and domain-informed causal reasoning. The discovery- and evaluation-cohort summaries were qualitatively consistent within each clustering family.}

\label{tab:glucose_cluster_profiles}
\begin{tabular}{|p{2.2cm}|p{6cm}|p{6cm}|}
\hline
\textbf{Method} & \textbf{Discovery cluster summary} & \textbf{Evaluation cluster summary} \\
\hline

K-means 
& 
Cluster 0: older. 
Cluster 1: high DiabetesPedigreeFunction, high BMI. 
Cluster 2: high BMI
Cluster 3:younger, lower BMI

& 
Cluster 0: older. 
Cluster 1: high DiabetesPedigreeFunction, high BMI. 
Cluster 2: high BMI
Cluster 3:younger, lower BMI
\\
\hline

FCM 
& 
Cluster 0: younger; lower BMI; lower DiabetesPedigreeFunction .
Cluster 1: older.
Cluster 2: younger; higher BMI.
Cluster 3: higher DiabetesPedigreeFunction.

& 
Cluster 0: younger; lower BMI; lower DiabetesPedigreeFunction. 
Cluster 1: older.
Cluster 2: younger; higher BMI.
Cluster 3: higher DiabetesPedigreeFunction.
\\
\hline

Bayesian GMM 
& 
Cluster 0:  younger
Cluster 1: older; lower DiabetesPedigreeFunction.
Cluster 2: older; higher DiabetesPedigreeFunction.
Cluster 3:higher BMI; higher DiabetesPedigreeFunction.
& 
Cluster 0:  younger
Cluster 1: older; lower DiabetesPedigreeFunction.
Cluster 2: older; higher DiabetesPedigreeFunction.
Cluster 3:higher BMI; higher DiabetesPedigreeFunction.
\\
\hline

\end{tabular}

\end{table}

\begin{table}[!h]
\centering
\scriptsize
\caption{\textbf{Glucose policy:} Pairwise targeted-person overlap between hardened subgroup-based allocations. Each allocation selected 53 eligible individuals in the held-out evaluation cohort.  Targeted-set agreement and Reference-allocation coverage are identical because all compared allocations selected the same number of eligible held-out individuals. K-means and FCM-based allocations selected highly similar eligible individuals, whereas Bayesian GMM selected a more distinct subset despite choosing the same number of eligible individuals.}
\label{tab:seed70_hardened_policy_overlap}
\begin{tabular}{llccccccc}
\toprule
Allocation A & Reference allocation B
& Selected A
& Selected B
& Shared selected
& Selected by either
& Jaccard
& Targeted-set agreement
& Reference-allocation coverage \\
\midrule
K-means & FCM hard
& 53 & 53 & 49 & 57
& 0.860 & 0.925 & 0.925 \\

K-means & FCM weighted
& 53 & 53 & 49 & 57
& 0.860 & 0.925 & 0.925 \\

K-means & Bayesian GMM
& 53 & 53 & 39 & 67
& 0.582 & 0.736 & 0.736 \\

FCM hard & FCM weighted
& 53 & 53 & 53 & 53
& \textbf{1.000} & \textbf{1.000} & \textbf{1.000} \\

FCM hard & Bayesian GMM
& 53 & 53 & 39 & 67
& 0.582 & 0.736 & 0.736 \\

FCM weighted & Bayesian GMM
& 53 & 53 & 39 & 67
& 0.582 & 0.736 & 0.736 \\
\bottomrule
\end{tabular}
\end{table}

\begin{table}[!h]
\centering
\scriptsize
\caption{\textbf{Glucose policy:} Run-wise targeted-person overlap between stochastic FCM and hardened cluster policies. Stochastic FCM was repeated over 500 stochastic hardening runs. Each stochastic run selected 53 eligible individuals, matching the selected count of each fixed hardened policy. Values are reported as mean $\pm$ standard deviation across stochastic runs. Stochastic FCM showed the highest overlap with the deterministic and weighted FCM allocations, slightly lower overlap with K-means, and the lowest overlap with Bayesian GMM.
}
\label{tab:seed70_stochastic_fcm_overlap}
\begin{tabular}{lccccc}
\toprule
Reference allocation

& Jaccard
& Targeted-set agreement
& Reference-allocation coverage \\
\midrule
Bayesian GMM

& $0.567 \pm 0.039$
& $0.723 \pm 0.0315$
& $0.723 \pm 0.0315$ \\

FCM hard

&  $0.705 \pm 0.047$
& $0.826 \pm 0.0326$
& $0.826 \pm 0.0326$ \\

FCM weighted

& $0.705 \pm 0.047$
& $0.826 \pm 0.0326$
& $0.826 \pm 0.0326$ \\

K-means

& $0.696 \pm 0.047$
& $0.820 \pm 0.0324$
& $0.820 \pm 0.0324$ \\
\bottomrule
\end{tabular}
\end{table}


\subsection{Policy Utility under the BMI-Based Hypothetical Intervention}

\begin{figure*}[!h]
    \centering
    \includegraphics[
        width=0.94\textwidth
    ]{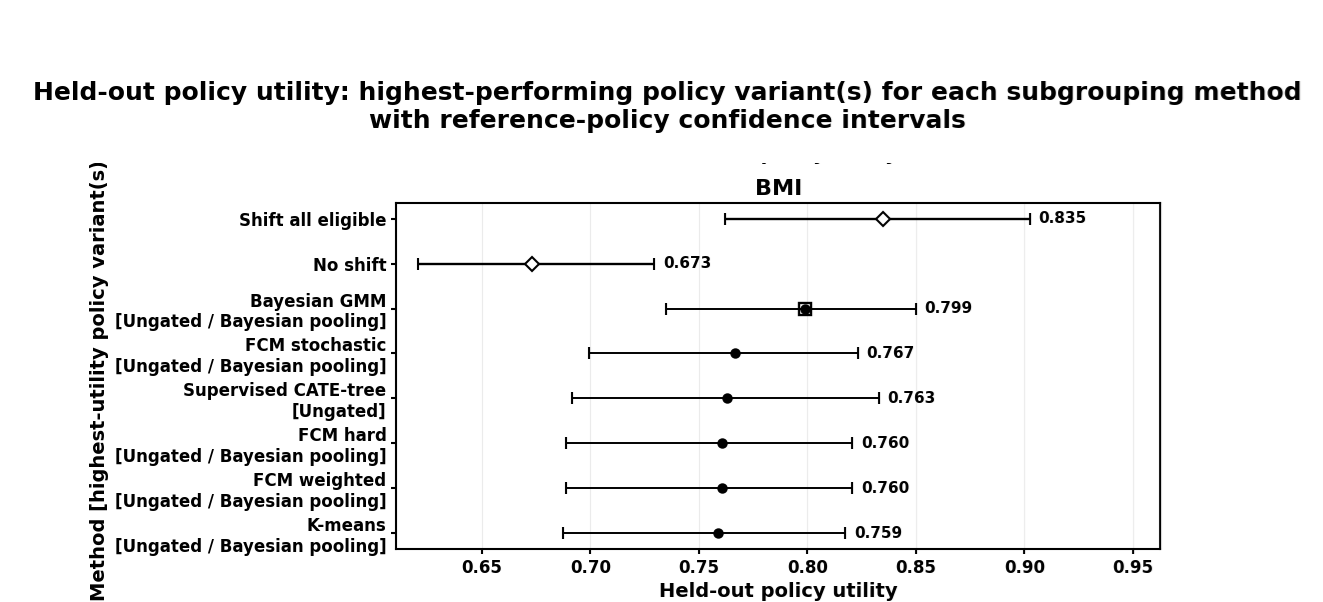}
    \caption{\textbf{BMI Policy}: Summary of representative-split held-out
    policy utility.
    For each subgrouping method, the highest observed
    held-out utility in the representative discovery--evaluation split is
    displayed. Points denote estimated policy utility, and horizontal bars
    denote bootstrap 95\% confidence intervals. Open diamonds denote the
    no-shift and shift-all-eligible reference policies. An open square
    identifies the highest point estimate among the budget-constrained
    subgroup policies within each experiment. The policy with the highest utility point estimate was achieved by Bayesian GMM. }
    \label{fig:representative_policy_utility_summary_bmi}
\end{figure*}

Table~\ref{tab:bmi_policy_results} presents the representative-split BMI-policy results. The ungated utility values were 0.7588 for K-means, 0.7605 for both deterministic and membership-weighted FCM, 0.7667 for stochastic FCM, and 0.7991 for Bayesian GMM. The supervised CATE-tree comparator achieved a utility of 0.7628, with its confidence interval overlapping those of the unsupervised methods. Empirical Bernstein gating excluded one or more lower-ranked clusters, while Bayesian pooling replicated the corresponding ungated allocations. 
Figure~\ref{fig:representative_policy_utility_summary_bmi} provides a
visual summary of the representative-split held-out policy
utilities.

The cluster profiles in Table~\ref{tab:BMI_cluster_profiles} were qualitatively consistent between the discovery and evaluation cohorts. K-means primarily identified younger individuals with lower pregnancy counts. FCM also prioritized younger individuals, but with variation in pregnancy count and \textit{DiabetesPedigreeFunction}, while Bayesian GMM additionally identified an older, higher-pregnancy phenotype. Table~\ref{tab:stochastic_fcm_bmi_uncertainty_summary} indicates modest variation in policy value under stochastic hardening. Tables~\ref{tab:bmi_hardened_policy_overlap} and~\ref{tab: BMI_stochastic_fcm_overlap} demonstrate greater allocation agreement among K-means and FCM policies compared to those involving Bayesian GMM.

\begin{figure*}[!h]
\centering
\includegraphics[
    width=\linewidth
]{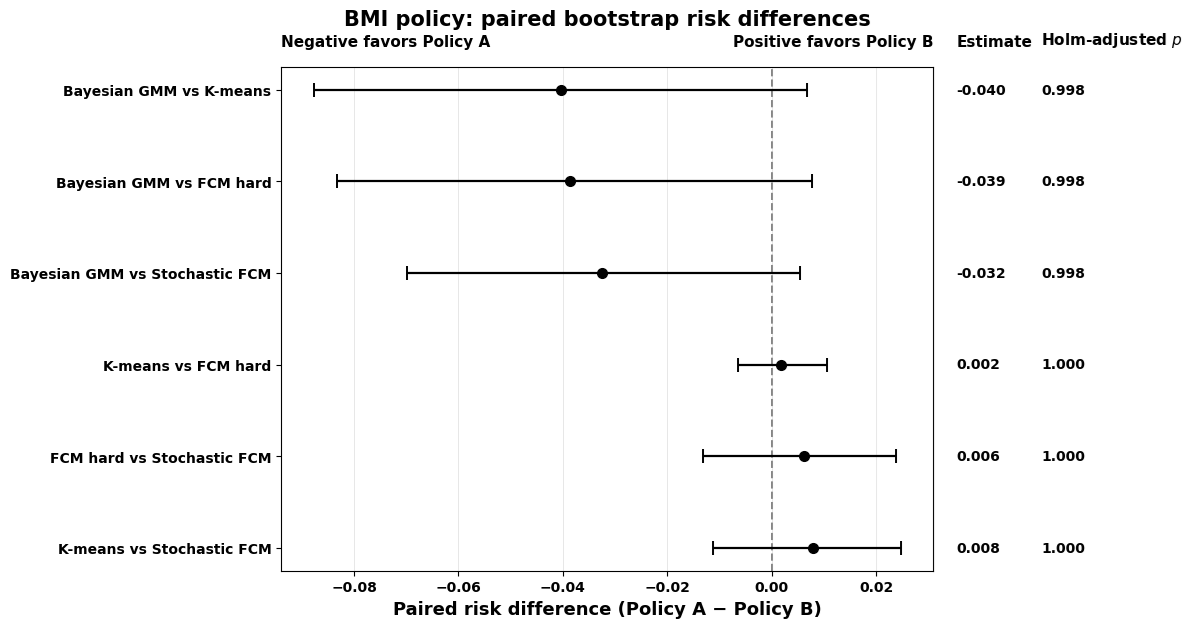}
\caption{
\textbf{BMI policy:} Paired bootstrap comparisons of held-out
full-population policy risk. Points show paired risk differences,
defined as Policy A risk minus Policy B risk, and horizontal bars
show paired 95\% bootstrap confidence intervals. The dashed line
denotes zero difference. Holm-adjusted \(p\)-values are shown in
the right-hand column. All confidence intervals included zero,
and no comparison remained statistically detectable after Holm
adjustment.
}
\label{pairwise_bootstrap_BMI_figure}
\end{figure*}

Paired bootstrap comparison results are shown in the figure ~\ref{pairwise_bootstrap_BMI_figure}. Bayesian GMM attained the lowest full-population risk point estimate compared with the other policies. However, the paired confidence intervals for all risk differences included zero, and no pairwise comparison remained statistically significant after Holm adjustment. Thus, although Bayesian GMM had the lowest estimated risk in this split, the analysis did not provide statistically detectable evidence that it achieved lower held-out risk than the other policies.

\begin{table}[!h]
\centering
\scriptsize
\caption{\textbf{BMI policy:} Policy risk and utility. The selected cluster order denotes the discovery-cohort ranking used to allocate the held-out intervention budget. The no-shift and shift-all-eligible rows provide common reference policies evaluated using the same held-out doubly robust estimator. The ungated and Bayesian-pooling allocations achieved similar held-out utilities across the unsupervised subgrouping methods, while the EB safety gate was generally more conservative. The highest estimated utility among the 70\% budgeted policies was $0.7991$ for the Bayesian GMM, compared with $0.6733$ under no shift and $0.8350$ under shifting all eligible individuals. The supervised CATE-tree comparator did not show a clear utility advantage.}

\label{tab:bmi_policy_results}


\begin{tabular}{llcccl}
\toprule
Subgrouping method
& Policy variant
& Policy risk
& Policy utility
& 95\% CI for utility
& Selected cluster order \\
\midrule

Reference policy
& No shift
& 0.3267
& 0.6733
& [0.6205, 0.7294]
& -- \\

Reference policy
& Shift all eligible
& 0.1650
& 0.8350
& [0.7620, 0.9026]
& -- \\
\midrule
K-means
& Ungated
& 0.2412
& 0.7588
& [0.6875, 0.8174]
& 2, 1 \\

K-means
& EB safety gate
& 0.2752
& 0.7248
& [0.6588, 0.7867]
& 2 \\

K-means
& Bayesian pooling
& 0.2412
& 0.7588
& [0.6875, 0.8174]
& 2, 1 \\

\midrule

FCM hard
& Ungated
& 0.2395
& 0.7605
& [0.6886, 0.8206]
& 1, 2, 0 \\

FCM hard
& EB safety gate
& 0.2468
& 0.7531
& [0.6856, 0.8233]
& 1, 2 \\

FCM hard
& Bayesian pooling
& 0.2395
& 0.7605
& [0.6886, 0.8206]
& 1, 2, 0 \\

\midrule

FCM weighted
& Ungated
& 0.2395
& 0.7605
& [0.6886, 0.8206]
& 1, 2, 0 \\

FCM weighted
& EB safety gate
& 0.2468
& 0.7531
& [0.6856, 0.8233]
& 1, 2 \\

FCM weighted
& Bayesian pooling
& 0.2395
& 0.7605
& [0.6886, 0.8206]
& 1, 2, 0 \\

\midrule

FCM stochastic
& Ungated
& 0.2333
& 0.7667
& [0.6993, 0.8231]
& 1, 2, 0 \\

FCM stochastic
& EB safety gate
& 0.2342
& 0.7658
& [0.6986, 0.8222]
& 1, 2 \\

FCM stochastic
& Bayesian pooling
& 0.2333
& 0.7667
& [0.6993, 0.8231]
& 1, 2, 0 \\

\midrule

Bayesian GMM
& Ungated
& \textbf{0.2009}
& \textbf{0.7991}
& \textbf{[0.7350, 0.8500]}
& 2, 1, 0 \\

Bayesian GMM
& EB safety gate
& 0.2153
& 0.7847
& [0.7333, 0.8561]
& 2, 1 \\

Bayesian GMM
& Bayesian pooling
& 0.2009
& 0.7991
& [0.7350, 0.8500]
& 2, 1, 0 \\

\midrule

Supervised CATE-tree
& Ungated
& 0.2372
& 0.7628
& [0.6916, 0.8330]
& -- \\

\bottomrule
\end{tabular}

\begin{flushleft}
\scriptsize
\textit{Note:} For the stochastic FCM rows, policy risk, policy utility, and confidence intervals summarize the expected policy value over MC stochastic hardening runs.
\end{flushleft}
\end{table}

\begin{table*}[!h]
\centering
\scriptsize

\caption{\textbf{BMI policy:} Coarse cluster-profile summaries based on the pre-treatment features selected through ensemble causal discovery and domain-informed causal reasoning. The discovery- and evaluation-cohort profiles were qualitatively consistent within each clustering family.}

\label{tab:BMI_cluster_profiles}
\begin{tabular}{|p{2.2cm}|p{6cm}|p{6cm}|}
\hline
\textbf{Method} & \textbf{Discovery cluster summary} & \textbf{Evaluation cluster summary} \\
\hline

K-means 
& 
Cluster 0: older, higher Pregnancies. 
Cluster 1: younger,lower Pregnancies,higher DiabetesPedigreeFunction. 
Cluster 2:  younger,lower Pregnancies,lower DiabetesPedigreeFunction

& 
Cluster 0: older, higher Pregnancies. 
Cluster 1: younger,lower Pregnancies,higher DiabetesPedigreeFunction. 
Cluster 2:  younger,lower Pregnancies,lower DiabetesPedigreeFunction

\\
\hline

FCM 
& 
Cluster 0: older,higher Pregnancies.
Cluster 1: younger,lower Pregnancies, lower DiabetesPedigreeFunction
Cluster 2: lower Pregnancies, higher DiabetesPedigreeFunction

& 

Cluster 0: older,higher Pregnancies.
Cluster 1: younger,lower Pregnancies, lower DiabetesPedigreeFunction
Cluster 2: lower Pregnancies, higher DiabetesPedigreeFunction

\\
\hline

Bayesian GMM 
& 
Cluster 0:  older, higher Pregnancies, lower DiabetesPedigreeFunction
Cluster 1:older, higher Pregnancies, higher DiabetesPedigreeFunction
Cluster 2:  younger, lower Pregnancies

& 
Cluster 0:  older, higher Pregnancies, lower DiabetesPedigreeFunction
Cluster 1:older, higher Pregnancies, higher DiabetesPedigreeFunction
Cluster 2:  younger, lower Pregnancies
\\
\hline

\end{tabular}

\end{table*}

\begin{table}[!h]
\centering
\scriptsize
\caption{\textbf{BMI Policy :} Policy-value uncertainty summary for stochastic FCM. Stochastic hardening introduced only modest MC variability for the ungated, Empirical Bernstein (EB)-gated, and Bayesian-gated FCM policies. The ungated and Bayesian-gated policies produced nearly identical summaries, while the EB-gated policy showed different MC variability due to variation in the expected number of targeted eligible individuals across stochastic hardenings.}
\label{tab:stochastic_fcm_bmi_uncertainty_summary}
\begin{tabular}{llcccccc}
\toprule
Policy variant
& Metric
& MC mean
& MC SD
& MC 2.5\%
& MC 97.5\%
& Bootstrap 2.5\%
& Bootstrap 97.5\% \\
\midrule

Ungated stochastic FCM
& Risk
& \textbf{0.233339}
& 0.010465
& 0.211809
& 0.253332
& 0.176923
& 0.300743 \\

Ungated stochastic FCM
& Utility
& \textbf{0.766661}
& 0.010465
& 0.746668
& 0.788191
& 0.699257
& 0.823077 \\

EB-gated stochastic FCM
& Risk
& 0.234182
& 0.010841
& 0.211809
& 0.256545
& 0.177771
& 0.301447 \\

EB-gated stochastic FCM
& Utility
& 0.765818
& 0.010841
& 0.743455
& 0.788191
& 0.698553
& 0.822229 \\

Bayesian-gated stochastic FCM
& Risk
& \textbf{0.233339}
& 0.010465
& 0.211809
& 0.253332
& 0.176923
& 0.300743 \\

Bayesian-gated stochastic FCM
& Utility
& \textbf{0.766661}
& 0.010465
& 0.746668
& 0.788191
& 0.699257
& 0.823077 \\

\bottomrule
\end{tabular}
\end{table}

\begin{table}[!h]
\centering
\scriptsize
\caption{\textbf{BMI Policy:} Pairwise targeted-person overlap between hardened subgroup-based allocations. Each allocation selected 133 eligible individuals in the held-out evaluation cohort. Targeted-set agreement and Reference-allocation coverage are identical because all compared allocations selected the same number of eligible held-out individuals. K-means and FCM-based allocations selected highly similar eligible individuals, whereas Bayesian GMM selected a more distinct subset despite choosing the same number of eligible individuals.}
\label{tab:bmi_hardened_policy_overlap}
\begin{tabular}{llccccccc}
\toprule
Allocation A & Reference allocation B
& Selected A
& Selected B
& Shared selected
& Selected by either
& Jaccard
& Targeted-set agreement
& Reference-allocation coverage \\
\midrule
K-means & FCM hard
& 133 & 133 & 131 & 135
& 0.970 & 0.985 & 0.985 \\

K-means & FCM weighted
& 133 & 133 & 131 & 135
& 0.970 & 0.985 & 0.985 \\

K-means & Bayesian GMM
& 133 & 133 & 109 & 157
& 0.694 & 0.820 & 0.820 \\

FCM hard & FCM weighted
& 133 & 133 & 133 & 133
& \textbf{1.000} & \textbf{1.000} & \textbf{1.000} \\

FCM hard & Bayesian GMM
& 133 & 133 & 110 & 156
& 0.705 & 0.827 & 0.827 \\

FCM weighted & Bayesian GMM
& 133 & 133 & 110 & 156
& 0.705 & 0.827 & 0.827 \\
\bottomrule
\end{tabular}
\end{table}

\begin{table}[!h]
\centering
\scriptsize
\caption{\textbf{BMI Policy:} Run-wise targeted-person overlap between stochastic FCM and hardened cluster policies. Stochastic FCM was repeated over 500 stochastic hardening runs. Each stochastic run selected 133 eligible individuals. Values are reported as mean $\pm$ standard deviation across stochastic runs. Stochastic FCM showed the highest overlap with K-means and the deterministic/weighted FCM allocations, and the lowest overlap with Bayesian GMM.}
\label{tab: BMI_stochastic_fcm_overlap}
\begin{tabular}{lccc}
\toprule
Reference allocation
& Jaccard
& Targeted-set agreement
& Reference-allocation coverage \\
\midrule

Bayesian GMM
& $0.683 \pm 0.024$
& $0.811 \pm 0.017$
& $0.811 \pm 0.017$ \\

FCM hard
& $0.829 \pm 0.029$
& $0.906 \pm 0.017$
& $0.906 \pm 0.017$ \\

FCM weighted
& $0.829 \pm 0.029$
& $0.906 \pm 0.017$
& $0.906 \pm 0.017$ \\

K-means
& $0.830 \pm 0.029$
& $0.907 \pm 0.017$
& $0.907 \pm 0.017$ \\

\bottomrule
\end{tabular}
\end{table}

\subsection{Policy Utility Under Smoking-Based Hypothetical State-Intervention Setting }

In the smoking-policy experiment, although categorical-native alternatives
such as latent class analysis (LCA) \cite{LCA} and $k$-modes \cite{kmodes} were also considered,
all primary subgrouping methods were applied to the same MCA
representation. This approach standardized the feature representation, ensuring that observed
differences more directly reflected the subgrouping algorithms rather than
simultaneous differences in both the representation and the clustering method.

\begin{figure*}[!h]
    \centering
    \includegraphics[
        width=0.94\textwidth
    ]{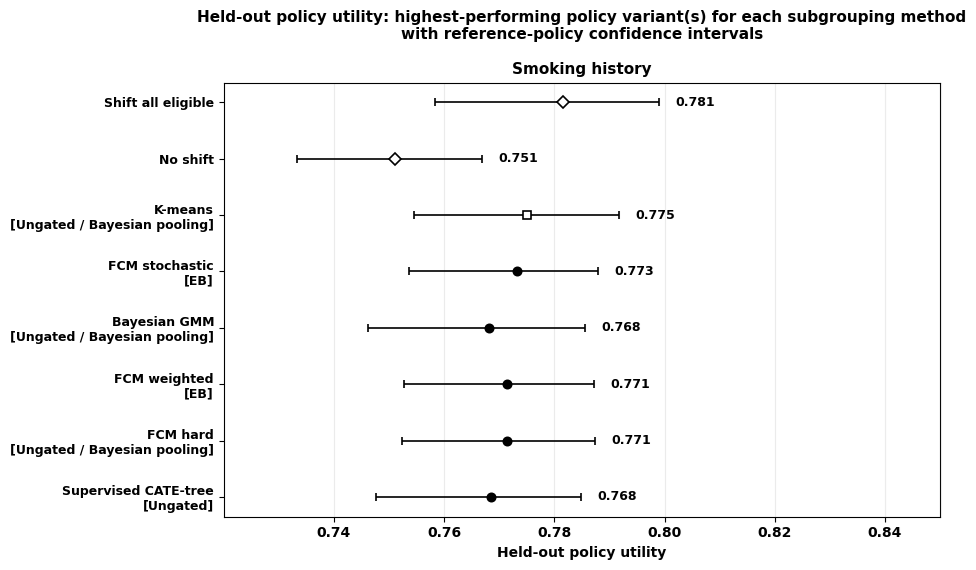}
    \caption{\textbf{Smoking Policy: Summary of representative-split held-out
    policy utility.}
    For each subgrouping method, the highest observed
    held-out utility in the representative discovery--evaluation split is
    displayed. Points denote estimated policy utility, and horizontal bars
    denote bootstrap 95\% confidence intervals. Open diamonds denote the
    no-shift and shift-all-eligible reference policies. An open square
    identifies the highest point estimate among the budget-constrained
    subgroup policies within each experiment. Here, KMeans achieved the policy with the highest utility point estimate utility.}
    \label{fig:representative_policy_utility_summary_smoking}
\end{figure*}

Table~\ref{tab:smoking_policy_results} presents the held-out smoking-policy results. Ungated utility ranged from 0.7681 for Bayesian Gaussian mixture model (GMM) to 0.7751 for K-means, with substantially overlapping bootstrap intervals. The supervised CATE-tree utility was 0.7685. 
Figure~\ref{fig:representative_policy_utility_summary_smoking} provides a
visual summary of the representative-split held-out policy utilities.
The coarse cluster profile is provided in Table~\ref{tab:smoking_cluster_profiles}.
For stochastic fuzzy c-means (FCM), as shown in Table~\ref{tab:smoking_stochastic_fcm_uncertainty}, the Monte Carlo (MC) standard deviation of utility ranged from approximately 0.0032 to 0.0038, whereas the bootstrap intervals were wider. In this split, sampling uncertainty contributed more to policy-value uncertainty than stochastic membership hardening.

\begin{figure*}[!h]
\centering
\includegraphics[
    width=\linewidth
]{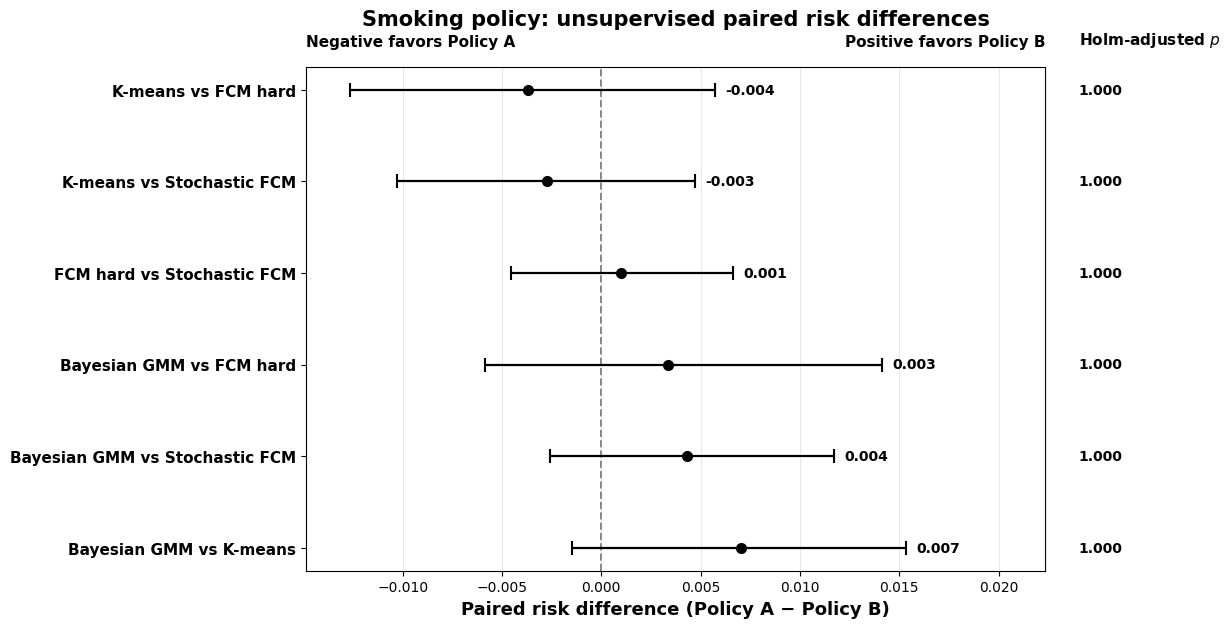}
\caption{
\textbf{Smoking policy:} Paired bootstrap comparisons of held-out policy risk. The dashed line
denotes zero difference. Holm-adjusted \(p\)-values are shown in
the right-hand column. All confidence intervals included zero,
and no comparison remained statistically detectable after Holm
adjustment.
}
\label{pairwise_bootstrap_smoking_figure}
\end{figure*}

Pairwise agreement between the fixed ungated allocations is presented in
Table~\ref{tab:smoking_mca_policy_overlap}.
The fixed ungated policies selected the same number of eligible individuals, but not the same individuals. Deterministic and membership-weighted FCM produced identical allocations, while their Jaccard overlap with Bayesian GMM was 0.506. Across the 500 stochastic hardening runs, as shown in Table~\ref{tab:smoking_stochastic_fcm_overlap}, stochastic FCM remained most similar to the deterministic FCM allocations. These results distinguish the stability of aggregate policy utility from the stability of individual targeting. Paired bootstrap comparison results are shown in the figure ~\ref{pairwise_bootstrap_smoking_figure}.
Although the EB-gated
weighted and stochastic fuzzy c-means (FCM) variants produced marginally more favorable
held-out point estimates, the primary paired comparison employed the ungated
variants to ensure a consistent policy construction rule across subgrouping
methods and to prevent data-dependent selection of variants.

K-means attained the lowest full-population risk point estimate compared with the other policies. However, the paired confidence intervals for all risk differences included zero, and no pairwise comparison remained statistically significant after Holm adjustment. Thus, although K-means had the lowest estimated risk in this split, the analysis did not provide statistically detectable evidence that it achieved lower held-out risk than the other policies.

\begin{table}[!h]
\centering
\scriptsize
\caption{\textbf{Smoking-history policy:} Policy risk and utility. The selected cluster order denotes the discovery-cohort ranking used to allocate the held-out intervention budget. The no-shift and shift-all-eligible rows provide common reference policies evaluated using the same doubly robust estimator. The ungated and Bayesian-pooling allocations achieved similar held-out utilities across the MCA-based subgrouping methods, while the EB safety gate was generally more conservative. The K-means achieved the highest utility point estimate of $0.7751$, compared with $0.7510$ under no shift and $0.7815$ under shifting all eligible individuals}.

\label{tab:smoking_policy_results}
\begin{tabular}{llcccl}
\toprule
Subgrouping method
& Policy variant
& Policy risk
& Policy utility
& 95\% CI for utility
& Selected cluster order \\
\midrule

Reference policy
& No shift
& 0.2490
& 0.7510
& [0.7333, 0.7669]
& -- \\

Reference policy
& Shift all eligible
& 0.2185
& 0.7815
& [0.7583, 0.7990]
& -- \\

\midrule

K-means
& Ungated
& \textbf{0.2249}
& \textbf{0.7751}
& \textbf{[0.7546, 0.7918]}
& 1, 5, 0, 2 \\

K-means
& EB safety gate
& 0.2366
& 0.7634
& [0.7437, 0.7798]
& 1 \\

K-means
& Bayesian pooling
& 0.2249
& 0.7751
& [0.7546, 0.7918]
& 1, 5, 0, 2 \\

\midrule

FCM hard
& Ungated
& 0.2286
& 0.7714
& [0.7524, 0.7873]
& 2, 0, 3, 1 \\

FCM hard
& EB safety gate
& 0.2396
& 0.7604
& [0.7427, 0.7748]
& 2 \\

FCM hard
& Bayesian pooling
& 0.2286
& 0.7714
& [0.7524, 0.7873]
& 2, 0, 3, 1 \\

\midrule

FCM weighted
& Ungated
& 0.2286
& 0.7714
& [0.7524, 0.7873]
& 0, 2, 3, 1 \\

FCM weighted
& EB safety gate
& 0.2285
& 0.7715
& [0.7528, 0.7872]
& 2, 0, 3, 5 \\

FCM weighted
& Bayesian pooling
& 0.2286
& 0.7714
& [0.7524, 0.7873]
& 0, 2, 3, 1 \\

\midrule

FCM stochastic
& Ungated
& 0.2276
& 0.7724
& [0.7528, 0.7873]
& 0, 2, 3, 1 \\

FCM stochastic
& EB safety gate
& 0.2267
& 0.7733
& [0.7536, 0.7879]
& 2, 0, 3, 5 \\

FCM stochastic
& Bayesian pooling
& 0.2276
& 0.7724
& [0.7528, 0.7873]
& 0, 2, 3, 1 \\

\midrule

Bayesian GMM
& Ungated
& 0.2319
& 0.7681
& [0.7462, 0.7856]
& 0, 1, 4 \\

Bayesian GMM
& EB safety gate
& 0.2425
& 0.7575
& [0.7390, 0.7736]
& 0 \\

Bayesian GMM
& Bayesian pooling
& 0.2319
& 0.7681
& [0.7462, 0.7856]
& 0, 1, 4 \\

\midrule

Supervised CATE-tree
& Ungated
& 0.2314
& 0.7685
& [0.7477, 0.7849]
& -- \\

\bottomrule
\end{tabular}

\begin{flushleft}
\scriptsize
\textit{Note:} For the stochastic FCM rows, policy risk, policy utility, and confidence intervals summarize the expected policy value over MC stochastic hardening runs.
\end{flushleft}
\end{table}

The resulting policy allocations were not determined by any single
demographic or clinical characteristic. For instance, the high-epilepsy clusters identified by K-means and Bayesian GMM
were not selected, despite having positive estimated subgroup effects. It was
because their small eligible sample sizes produced comparatively low discovery-cohort aggregate gains.

\begin{table}[!h]
\centering
\scriptsize
\caption{\textbf{Smoking-history policy:} Clusters were learned in MCA factor space and subsequently interpreted using
the original categorical effect modifiers by comparing within-cluster
category prevalence with their discovery-cohort reference prevalence. Coarse cluster profile summaries. The discovery and evaluation summaries were qualitatively consistent within each clustering family.}
\label{tab:smoking_cluster_profiles}
\begin{tabular}{|p{2.2cm}|p{6cm}|p{6cm}|}
\hline
\textbf{Method} & \textbf{Discovery cluster summary} & \textbf{Evaluation cluster summary} \\
\hline
K-means
&
Cluster 0: more Other Race / Multiracial participants; higher educational attainment; more men; lower epilepsy prevalence.\newline Cluster 1: more Non-Hispanic White participants; higher educational attainment; moderate-to-higher income; lower epilepsy prevalence.\newline Cluster 2: more Mexican American participants; lower educational attainment; lower epilepsy prevalence; more men.\newline Cluster 3: higher epilepsy prevalence; lower income.\newline Cluster 4: more Other Hispanic participants; lower epilepsy prevalence.\newline Cluster 5: more Non-Hispanic Black participants; more high-school graduates; more women; lower epilepsy prevalence.
&
Cluster 0: more Other Race / Multiracial participants; higher educational attainment; more men; lower epilepsy prevalence.\newline Cluster 1: more Non-Hispanic White participants; higher educational attainment; moderate-to-higher income; lower epilepsy prevalence.\newline Cluster 2: more Mexican American participants; lower educational attainment; lower epilepsy prevalence; more men.\newline Cluster 3: higher epilepsy prevalence; more women.\newline Cluster 4: more Other Hispanic participants; lower epilepsy prevalence; more women.\newline Cluster 5: more Non-Hispanic Black participants; lower income; more women; lower epilepsy prevalence.
\\
\hline
FCM
&
Cluster 0: more adults aged 40--60 years; higher educational attainment; moderate-to-higher income; more women.\newline Cluster 1: lower educational attainment; more Mexican American participants; lower income.\newline Cluster 2: younger adults; higher educational attainment; more Other Race / Multiracial participants; moderate-to-higher income.\newline Cluster 3: more high-school graduates; more Non-Hispanic Black participants; lower income.\newline Cluster 4: older adults; more Non-Hispanic White participants; moderate-to-higher income; more men.\newline Cluster 5: more Other Hispanic participants; older adults.
&
Cluster 0: more adults aged 40--60 years; higher educational attainment; moderate-to-higher income; more women.\newline Cluster 1: lower educational attainment; more Mexican American participants; lower income.\newline Cluster 2: younger adults; higher educational attainment; more Other Race / Multiracial participants; moderate-to-higher income.\newline Cluster 3: more Non-Hispanic Black participants; more high-school graduates; lower income.\newline Cluster 4: older adults; more Non-Hispanic White participants; moderate-to-higher income; more men.\newline Cluster 5: more Other Hispanic participants; older adults; more men; lower epilepsy prevalence.
\\
\hline
Bayesian GMM
&
Cluster 0: more adults aged 40--60 years; lower epilepsy prevalence.\newline Cluster 1: more Non-Hispanic White participants; older adults; lower epilepsy prevalence.\newline Cluster 2: more Non-Hispanic Black participants; younger adults; lower epilepsy prevalence.\newline Cluster 3: higher epilepsy prevalence; lower income.\newline Cluster 4: more Mexican American participants; lower educational attainment; lower income; lower epilepsy prevalence.\newline Cluster 5: more Other Hispanic participants; lower educational attainment; lower epilepsy prevalence.
&
Cluster 0: more adults aged 40--60 years; higher educational attainment; lower epilepsy prevalence.\newline Cluster 1: more Non-Hispanic White participants; older adults; lower epilepsy prevalence.\newline Cluster 2: more Non-Hispanic Black participants; younger adults; lower epilepsy prevalence.\newline Cluster 3: higher epilepsy prevalence; more women.\newline Cluster 4: more Mexican American participants; lower educational attainment; lower income; lower epilepsy prevalence.\newline Cluster 5: more Other Hispanic participants; lower educational attainment; lower epilepsy prevalence.
\\
\hline
\end{tabular}
\end{table}

\begin{table}[!h]
\centering
\scriptsize
\caption{\textbf{Smoking-history policy}: Policy-value uncertainty summary for stochastic FCM.
Stochastic hardening introduced only modest MC variability for the ungated and Bayesian-gated stochastic FCM policies. The ungated and Bayesian-pooled policies produced identical uncertainty summaries in this split, while the EB-gated stochastic policy produced a slightly higher mean utility and selected a smaller expected number of eligible individuals.}
\label{tab:smoking_stochastic_fcm_uncertainty}
\begin{tabular}{llcccccc}
\toprule
Policy variant
& Metric
& MC mean
& MC SD
& MC 2.5\%
& MC 97.5\%
& Bootstrap 2.5\%
& Bootstrap 97.5\% \\
\midrule

Ungated stochastic FCM
& Risk
& 0.227589
& 0.003157
& 0.221581
& 0.233557
& 0.212726
& 0.247242 \\

Ungated stochastic FCM
& Utility
& 0.772411
& 0.003157
& 0.766443
& 0.778419
& 0.752758
& 0.787274 \\

EB-gated stochastic FCM
& Risk
& \textbf{0.226747}
& 0.003796
& 0.219481
& 0.234687 
& 0.212062
& 0.246400 \\

EB-gated stochastic FCM
& Utility
& \textbf{0.773253}
& 0.003796
&  0.765313
& 0.780519
& 0.753600
& 0.787938 \\

Bayesian-gated stochastic FCM
& Risk
& 0.227589
& 0.003157
& 0.221581
& 0.233557
& 0.212726
& 0.247242 \\

Bayesian-gated stochastic FCM
& Utility
& 0.772411
& 0.003157
& 0.766443
& 0.778419
& 0.752758
& 0.787274 \\

\bottomrule
\end{tabular}
\end{table}

\begin{table}[!h]
\centering
\scriptsize
\caption{\textbf{Smoking-history policy:}Pairwise targeted-person overlap between ungated policies.
The table compares final held-out targeted-person allocations across MCA-based subgrouping methods. All methods selected the same total number of eligible individuals, but overlap varied across methods, indicating that different clustering algorithms prioritized partly different individuals.}
\label{tab:smoking_mca_policy_overlap}
\begin{tabular}{llcccccccc}
\toprule
Allocation A
& Reference allocation B
& Selected A
& Selected B
& Shared selected
& Selected by either
& Jaccard
& Targeted-set agreement
& Reference-allocation coverage
 \\
\midrule

K-means
& FCM hard
& 891
& 891
& 647
& 1135
& 0.570
& 0.726
& 0.726
 \\

K-means
& FCM weighted
& 891
& 891
& 647
& 1135
& 0.570
& 0.726
& 0.726
 \\

K-means
& Bayesian GMM
& 891
& 891
& 674
& 1108
& 0.608
& 0.757
& 0.757
 \\

FCM hard
& FCM weighted
& 891
& 891
& 891
& 891
& \textbf{1.000}
& \textbf{1.000}
& \textbf{1.000}
 \\

FCM hard
& Bayesian GMM
& 891
& 891
& 599
& 1183
& 0.506
& 0.672
& 0.672
 \\

FCM weighted
& Bayesian GMM
& 891
& 891
& 599
& 1183
& 0.506
& 0.672
& 0.672
 \\

\bottomrule
\end{tabular}
\end{table}

\begin{table}[!h]
\centering
\scriptsize
\caption{\textbf{Smoking-history policy:}Run-wise targeted-person overlap between stochastic FCM and ungated MCA-based policies.
Each row summarizes 500 stochastic hardening runs compared with a fixed ungated reference allocation. Values are reported as MC mean $\pm$ standard deviation.}
\label{tab:smoking_stochastic_fcm_overlap}
\begin{tabular}{lccc}
\toprule
Reference allocation
& Jaccard
& Targeted-set agreement
& Reference-allocation coverage \\
\midrule

Bayesian GMM
& $0.5357 \pm 0.0104$
& $0.6976 \pm 0.0088$
& $0.6976 \pm 0.0088$ \\

FCM hard
& $0.5766 \pm 0.0117$
& $0.7314 \pm 0.0094$
& $0.7314 \pm 0.0094$ \\

FCM weighted
& $0.5766 \pm 0.0117$
& $0.7314 \pm 0.0094$
& $0.7314 \pm 0.0094$ \\

K-means
& $0.5184 \pm 0.0094$
& $0.6827 \pm 0.0082$
& $0.6827 \pm 0.0082$ \\

\bottomrule
\end{tabular}
\end{table}



\subsection{Paired Held-Out Risk Comparisons: Supervised versus Unsupervised Policies} 
\begin{figure*}[!h]
    \centering
    \caption{\textbf{Glucose policy:} Paired bootstrap comparisons of the supervised CATE-tree policy with the unsupervised subgroup policies. The dashed vertical line denotes no risk difference. All confidence intervals included zero, and none of the supervised-versus-unsupervised comparisons was statistically significant after multiplicity adjustment.}
    \includegraphics[
        width=0.94\textwidth
    ]{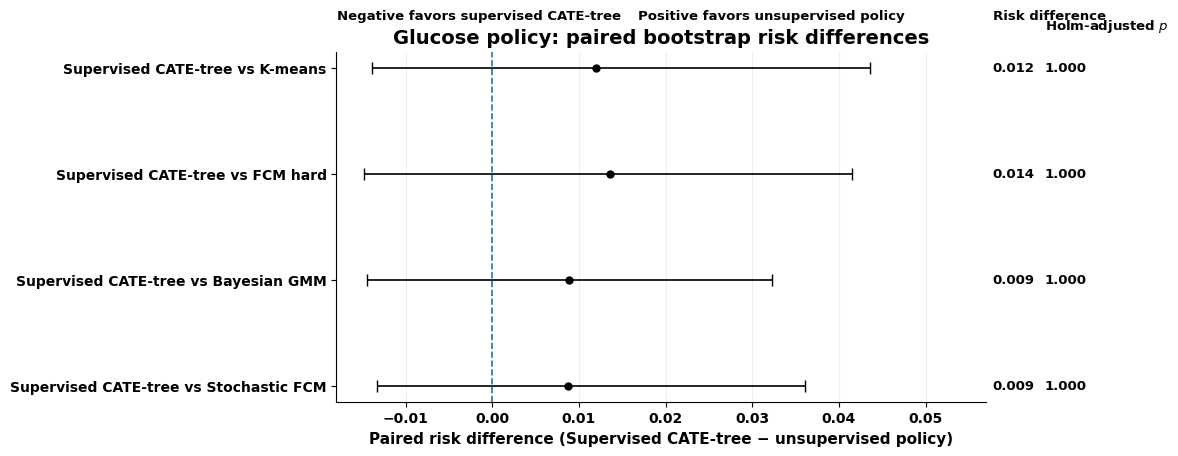}
    \label{supervised_p_values_glucose}
\end{figure*}

\begin{figure*}[!h]
    \centering
    \caption{\textbf{BMI policy:} Paired bootstrap comparisons of the supervised CATE-tree policy with the unsupervised subgroup policies. The dashed vertical line denotes no risk difference. All confidence intervals included zero, and none of the supervised-versus-unsupervised comparisons was statistically significant after multiplicity adjustment.}
    \includegraphics[
        width=0.94\textwidth
    ]{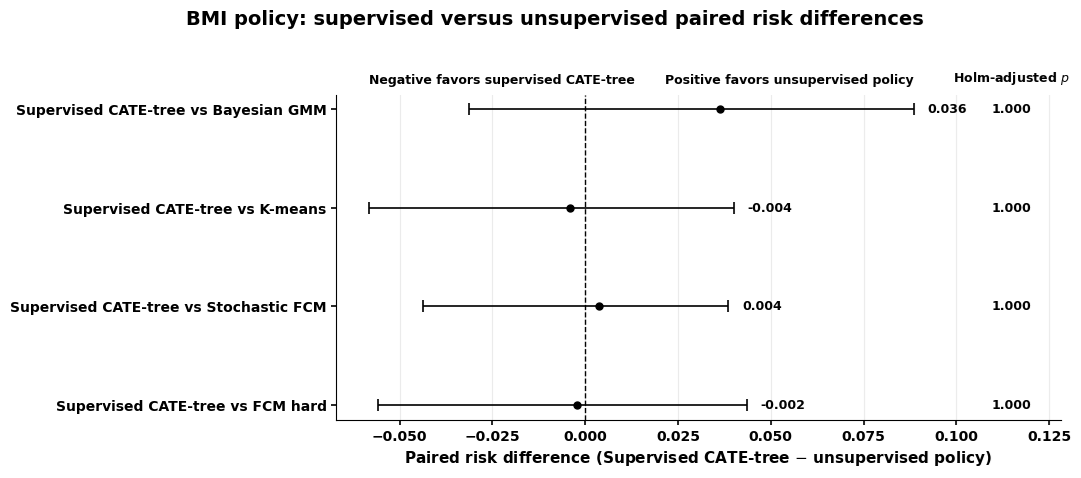}
    \label{supervised_p_values_bmi}
\end{figure*}

\begin{figure*}[!h]
    \centering
    \caption{\textbf{Smoking policy:} Paired bootstrap comparisons of the supervised CATE-tree policy with the unsupervised subgroup policies. The dashed vertical line denotes no risk difference. All confidence intervals included zero, and none of the supervised-versus-unsupervised comparisons was statistically significant after multiplicity adjustment.}
    \includegraphics[
        width=0.94\textwidth
    ]{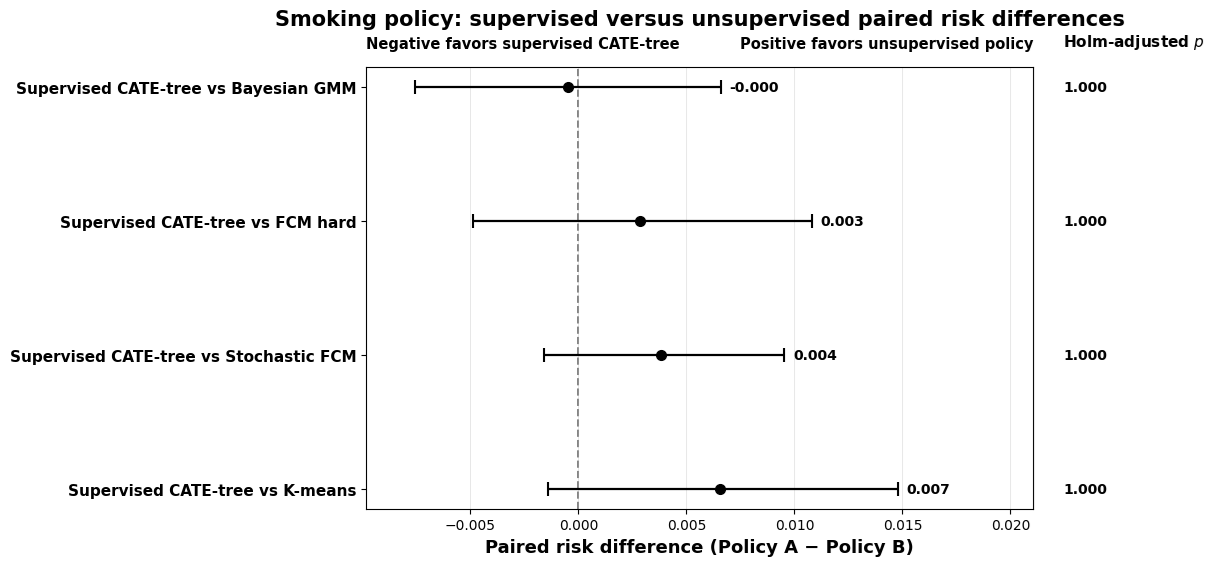}
    \label{supervised_p_values_smoking}
\end{figure*}

This section highlights only the supervised versus unsupervised contrasts for readability. However, the multiplicity adjustment was applied globally across the full family of all pairwise policy comparisons as explained under the section \ref{paired_compares}.
The supervised-versus-unsupervised glucose-policy comparisons are
visualized in Figure~\ref{supervised_p_values_glucose}, the BMI-policy comparisons in Figure~\ref{supervised_p_values_bmi}, and the
smoking-policy comparisons in Figure~\ref{supervised_p_values_smoking}. Exact numerical results,
including the paired confidence intervals and unadjusted
centered-bootstrap \(p\)-values, are reported in Supplementary Files. No comparison remained
statistically significant after Holm adjustment.


\subsection{Sensitivity Analysis}
Across the three policy experiments, differences among the ungated subgrouping methods were generally small relative to discovery--evaluation split variability. The supervised CATE-tree comparator showed no consistent utility advantage, Bayesian pooling usually remained close to ungated allocation, and Empirical Bernstein gating was more restrictive. The policy-specific analyses below therefore focus primarily on differences in allocation overlap and on departures from this common pattern.

\subsubsection{Sensitivity Analysis: Glucose-Based Hypothetical State Intervention}

\begin{figure*}[!h]

    \centering
   \includegraphics[width=11cm, height=9cm]{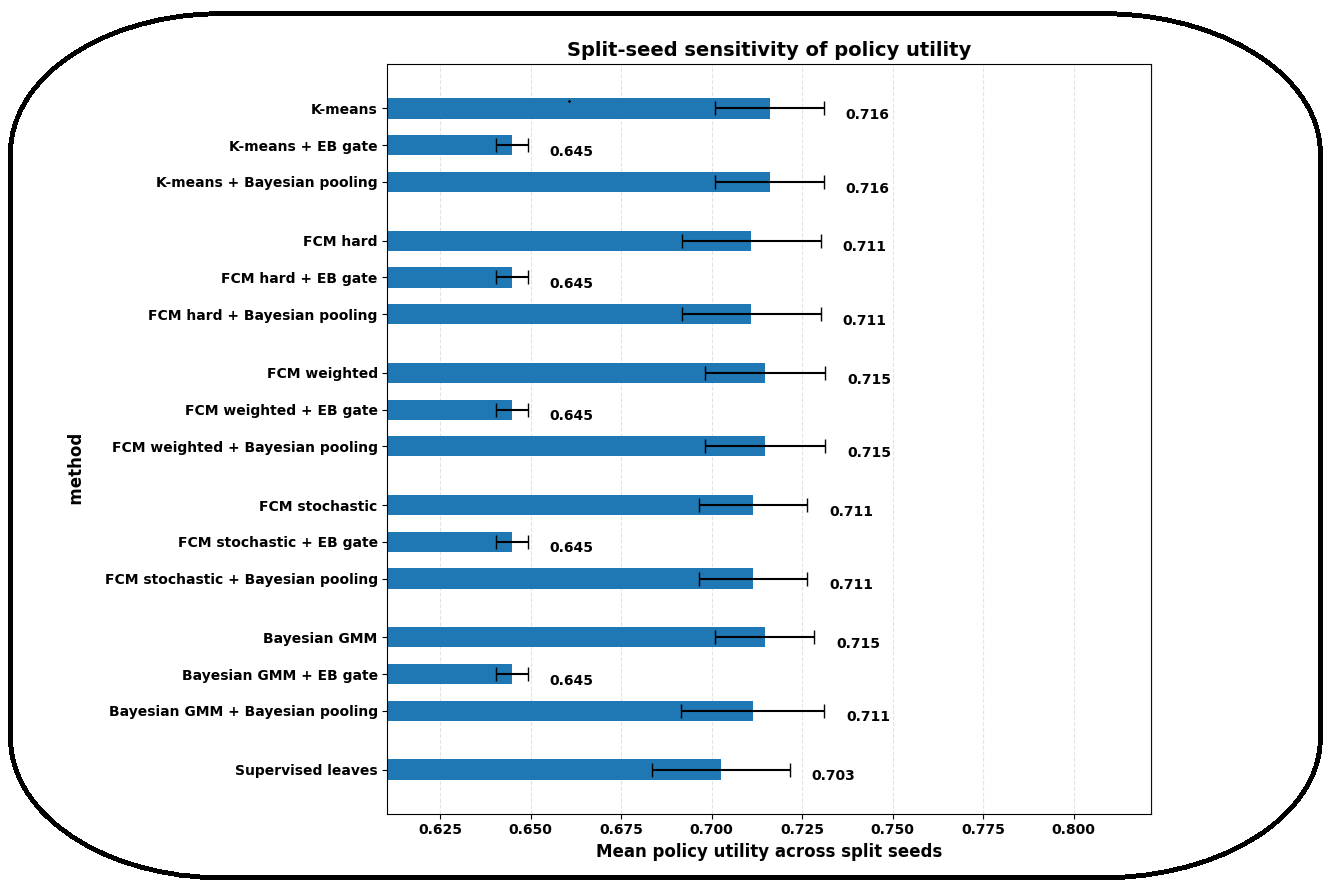}
    \caption{\textbf{Glucose Policy:}Split-seed sensitivity of policy utility across four discovery/evaluation splits for the glucose policy experiment. Bars show the mean held-out policy utility across split seeds, and horizontal error bars show standard deviation across seeds. 
    The ungated unsupervised subgrouping methods produced similar mean utilities, with K-means, weighted FCM, and Bayesian GMM yielding the highest mean values across the examined splits. Empirical Bernstein safety gating was consistently more conservative, resulting in lower mean utility across methods. Bayesian pooling generally preserved utilities close to the corresponding ungated policies, while the supervised CATE-tree comparator did not show a consistent utility advantage over the unsupervised subgroup policies.}

\label{gluc_sens}
\end{figure*}

\begin{figure*}[!ht]
\centering

\begin{minipage}[t]{0.40\textwidth}
\vspace{0pt}
\centering

\includegraphics[
    width=\linewidth
]{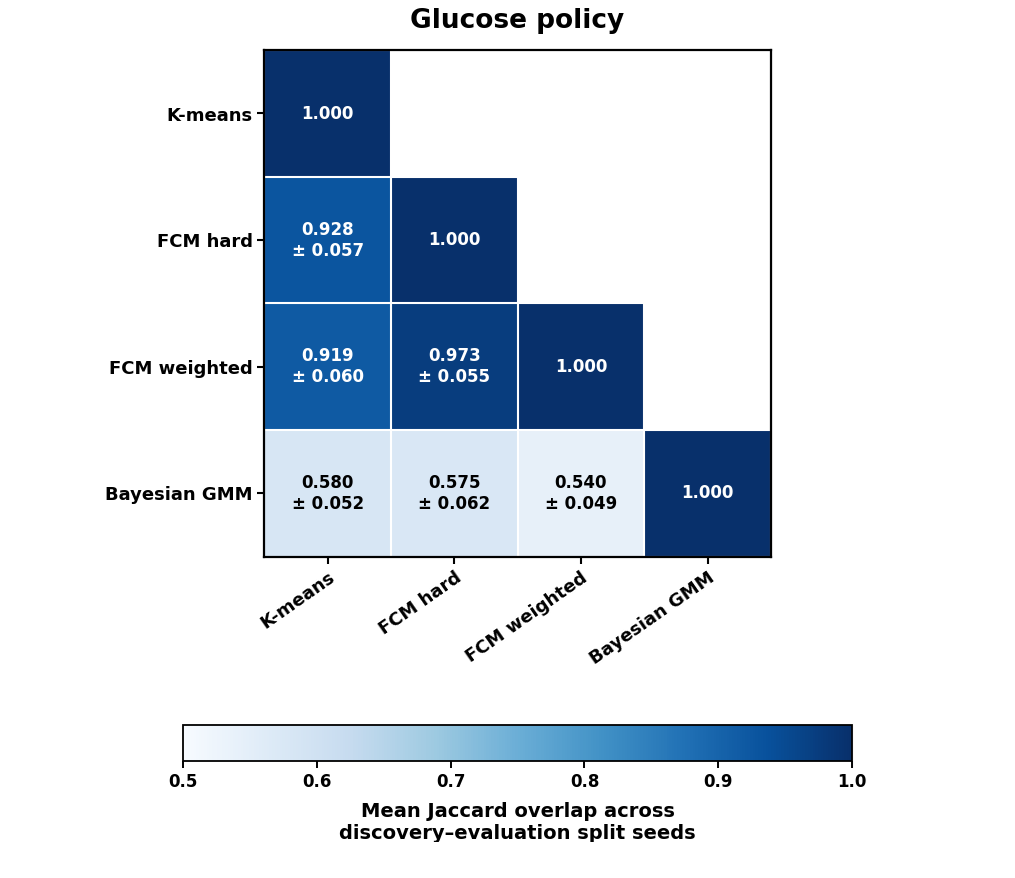}

\captionof{figure}{
\textbf{Glucose policy:} Visual summary of the mean pairwise Jaccard
overlap among the fixed ungated subgroup policies across the examined
discovery--evaluation split seeds. Cells report the mean Jaccard overlap,
with the standard deviation shown below the mean. The diagonal equals
one by definition.
}
\label{fig:glucose_jaccard_heatmap}

\end{minipage}
\hfill
\begin{minipage}[t]{0.57\textwidth}
\vspace{0pt}
\centering
\scriptsize

\captionof{table}{
\textbf{Glucose policy:} Mean targeted-person overlap metrics across
split seeds. Values are reported as mean $\pm$ standard deviation.
FCM hard and membership-weighted allocations showed the highest
agreement with each other and with K-means, whereas Bayesian GMM
consistently selected a more distinct subset of eligible individuals
despite achieving comparable policy utility.
}
\label{tab:overlap_metrics_mean_sd_across_seeds}

\resizebox{\linewidth}{!}{
\begin{tabular}{lccc}
\toprule
Allocation pair
& Jaccard
& \makecell{Targeted-set\\agreement}
& \makecell{Reference-allocation\\coverage} \\
\midrule

K-means vs FCM hard
& $0.928 \pm 0.057$
& $0.962 \pm 0.031$
& $0.962 \pm 0.031$ \\

K-means vs FCM weighted
& $0.919 \pm 0.060$
& $0.957 \pm 0.032$
& $0.957 \pm 0.032$ \\

K-means vs Bayesian GMM
& $0.580 \pm 0.052$
& $0.733 \pm 0.042$
& $0.733 \pm 0.042$ \\

FCM hard vs FCM weighted
& $0.973 \pm 0.055$
& $0.986 \pm 0.029$
& $0.986 \pm 0.029$ \\

FCM hard vs Bayesian GMM
& $0.575 \pm 0.062$
& $0.728 \pm 0.052$
& $0.728 \pm 0.052$ \\

FCM weighted vs Bayesian GMM
& $0.540 \pm 0.049$
& $0.700 \pm 0.042$
& $0.700 \pm 0.042$ \\

\bottomrule
\end{tabular}
}

\end{minipage}

\end{figure*}

Figure~\ref{gluc_sens} summarizes split-seed sensitivity of held-out policy utility for the glucose-based hypothetical state intervention. Table~\ref{tab:overlap_metrics_mean_sd_across_seeds}, figure~\ref {fig:glucose_jaccard_heatmap}, and table~\ref{tab:stochastic_fcm_overlap_across_seeds} further show that comparable policy utility does not necessarily imply identical targeted individuals. For the glucose policy, K-means and the FCM variants selected largely overlapping eligible populations across splits, whereas Bayesian GMM produced a more distinct allocation. The stochastic FCM overlap results in Table~\ref{tab:stochastic_fcm_overlap_across_seeds} provide a complementary stability check. Across split seeds, stochastic FCM remained closest to deterministic and weighted FCM, followed by K-means, and showed lower overlap with Bayesian GMM.


\begin{table}[!h]
\centering
\scriptsize
\caption{\textbf{Glucose policy:} Across-seed summary of run-wise targeted-person overlap between stochastic FCM and hardened allocations. For each split seed, stochastic FCM was repeated over 500 stochastic hardening runs; values are reported as the mean $\pm$ standard deviation of the seed-level means across split seeds. Stochastic FCM remained most similar to the deterministic and weighted FCM allocations, followed by K-means, and showed the lowest overlap with Bayesian GMM.}

\label{tab:stochastic_fcm_overlap_across_seeds}
\begin{tabular}{lcccc}
\toprule
Reference Allocation
 
& Jaccard 
& Targeted-set agreement 
& Reference-allocation coverage \\
\midrule
Bayesian GMM 

& $0.554 \pm 0.043$ 
& $0.712 \pm 0.036$ 
& $0.712 \pm 0.036$ \\

FCM hard 

& $0.718 \pm 0.016$ 
& $0.835 \pm 0.011$ 
& $0.835 \pm 0.011$ \\

FCM weighted 

& \textbf{$0.723 \pm 0.015$} 
& $0.839 \pm 0.010$ 
& $0.839 \pm 0.010$ \\

K-means 

& $0.713 \pm 0.020$ 
& $0.831 \pm 0.013$ 
& $0.831 \pm 0.013$ \\
\bottomrule
\end{tabular}
\end{table}

\subsubsection{Sensitivity Analysis: BMI-Based Hypothetical State Intervention}

\begin{figure*}[!h]

    \centering
   \includegraphics[width=11cm, height=9cm]{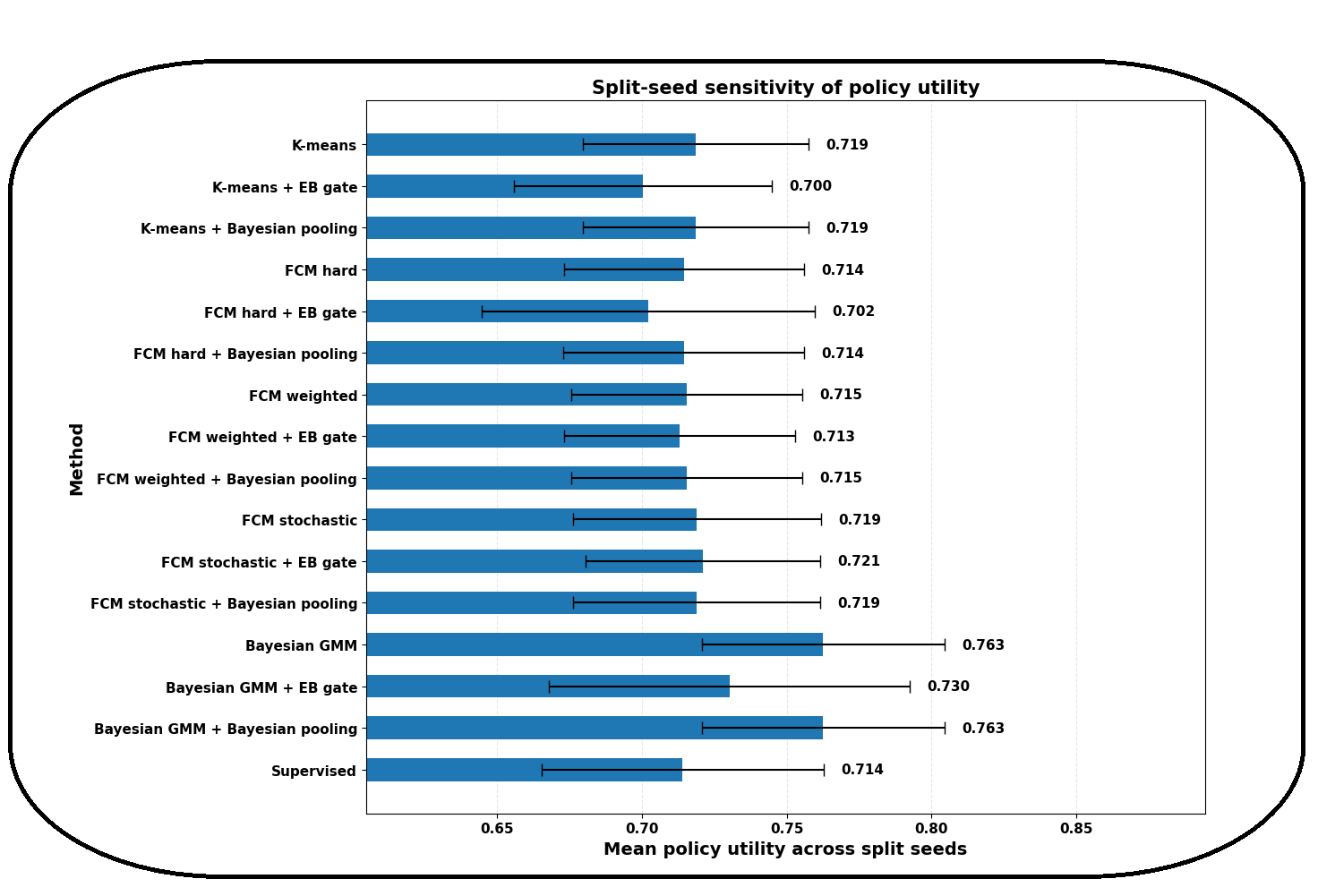}

\caption{\textbf{BMI policy :} Split-seed sensitivity of held-out policy utility across four discovery/evaluation splits. Bars show the mean policy utility across split seeds, and horizontal error bars show the standard deviation across seeds. The ungated unsupervised subgrouping methods produced broadly similar utilities, while Bayesian GMM achieved the highest mean utility among the evaluated methods. Bayesian pooling closely tracked the corresponding ungated policies, indicating that the posterior gate was non-restrictive or only weakly restrictive in these runs. Empirical Bernstein gating was generally more conservative, reducing mean utility for most methods, although the stochastic FCM variant showed a small point-estimated improvement under EB gating. The supervised CATE-tree comparator performed similarly to several unsupervised methods but did not exceed the best unsupervised policy.}
\label{fig:BMI_split_seed_utility}
\label{BMI_sens}
\end{figure*}

\begin{figure*}[!ht]
\centering

\begin{minipage}[t]{0.40\textwidth}
\vspace{0pt}
\centering

\includegraphics[
    width=\linewidth
]{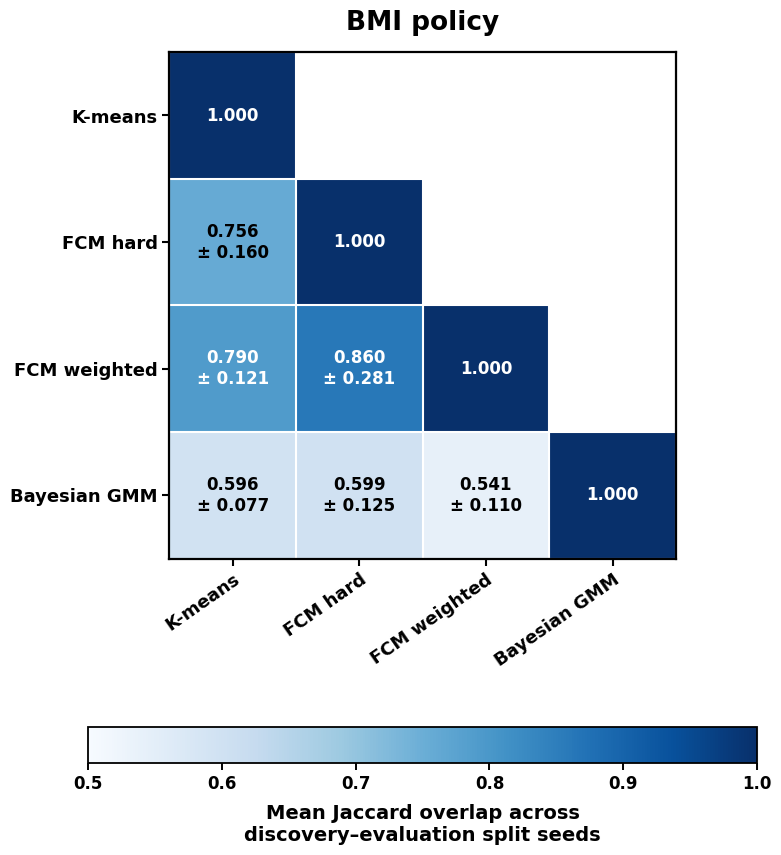}

\captionof{figure}{
\textbf{BMI policy:} Visual summary of the mean pairwise Jaccard
overlap among the fixed ungated subgroup policies across four
discovery--evaluation split seeds. Cells report the mean overlap,
with the standard deviation shown below the mean. The diagonal
equals one by definition.
}
\label{fig:bmi_jaccard_heatmap}

\end{minipage}
\hfill
\begin{minipage}[t]{0.57\textwidth}
\vspace{0pt}
\centering
\scriptsize

\captionof{table}{
\textbf{BMI policy:} Mean pairwise targeted-person overlap across four
random seeds, reported as mean $\pm$ standard deviation. The greatest
overlap was observed between hard and membership-weighted FCM, whereas
Bayesian GMM generally selected more distinct individuals. Targeted-set
agreement and reference-allocation coverage are identical because all
compared allocations selected the same number of eligible held-out
individuals.
}
\label{tab:bmi_overlap_mean_sd_across_seeds}

\resizebox{\linewidth}{!}{
\begin{tabular}{lccc}
\toprule
Allocation pair
& Jaccard
& \makecell{Targeted-set\\agreement}
& \makecell{Reference-allocation\\coverage} \\
\midrule

K-means vs FCM hard
& $0.756 \pm 0.160$
& $0.854 \pm 0.102$
& $0.854 \pm 0.102$ \\

K-means vs FCM weighted
& $0.790 \pm 0.121$
& $0.879 \pm 0.071$
& $0.879 \pm 0.071$ \\

K-means vs Bayesian GMM
& $0.596 \pm 0.077$
& $0.745 \pm 0.060$
& $0.745 \pm 0.060$ \\

FCM hard vs FCM weighted
& $0.860 \pm 0.281$
& $0.902 \pm 0.195$
& $0.902 \pm 0.195$ \\

FCM hard vs Bayesian GMM
& $0.599 \pm 0.125$
& $0.744 \pm 0.098$
& $0.744 \pm 0.098$ \\

FCM weighted vs Bayesian GMM
& $0.541 \pm 0.110$
& $0.698 \pm 0.087$
& $0.698 \pm 0.087$ \\

\bottomrule
\end{tabular}
}

\end{minipage}

\end{figure*}

The split-seed sensitivity of held-out BMI-policy utility is shown in Figure~\ref{fig:BMI_split_seed_utility}. For the BMI policy, Bayesian GMM attained the highest mean utility, although its across-seed variability overlapped that of the other methods.
Pairwise overlap between the fixed ungated allocations is summarized in
Table~\ref{tab:bmi_overlap_mean_sd_across_seeds} and figure ~\ref{fig:bmi_jaccard_heatmap}. The strongest average
agreement was observed between deterministic hard and membership-weighted
FCM, although this comparison also showed substantial variation
across split seeds. K-means likewise showed relatively high overlap with both
FCM variants. In contrast, comparisons involving Bayesian GMM
generally produced lower Jaccard indices, indicating that its policy tended to
prioritize a more distinct subset of eligible individuals.

The run-wise overlap of stochastic FCM with the fixed allocations is
reported in Table~\ref{tab:bmi_stochastic_fcm_overlap_across_seeds}. 
Stochastic FCM was closest to membership-weighted FCM, although the amount of agreement varied across splits.

\begin{table}[!h]
\centering
\scriptsize
\caption{\textbf{BMI policy:} Across-seed summary of run-wise targeted-person overlap between stochastic FCM and hardened allocations. For each split seed, stochastic FCM was repeated over 500 stochastic hardening runs. Values are reported as the mean $\pm$ standard deviation of the seed-level means across split seeds. Stochastic FCM was most similar on average to the membership-weighted FCM allocation, followed by K-means and deterministic FCM hardening, and showed the lowest overlap with Bayesian GMM.}
\label{tab:bmi_stochastic_fcm_overlap_across_seeds}
\begin{tabular}{lccc}
\toprule
Reference allocation
& Jaccard
& Targeted-set agreement
& Reference-allocation coverage \\
\midrule
Bayesian GMM
& $0.551 \pm 0.089$
& $0.707 \pm 0.071$
& $0.707 \pm 0.071$ \\

FCM hard
& $0.690 \pm 0.149$
& $0.809 \pm 0.112$
& $0.809 \pm 0.112$ \\

FCM weighted
& $0.758 \pm 0.053$
& $0.861 \pm 0.034$
& $0.861 \pm 0.034$ \\

K-means
& $0.710 \pm 0.080$
& $0.828 \pm 0.052$
& $0.828 \pm 0.052$ \\
\bottomrule
\end{tabular}
\end{table}



\subsubsection{Sensitivity Analysis: Smoking-Based Hypothetical State Intervention}

\begin{figure*}[!h]

    \centering
   \includegraphics[width=11cm, height=9cm]{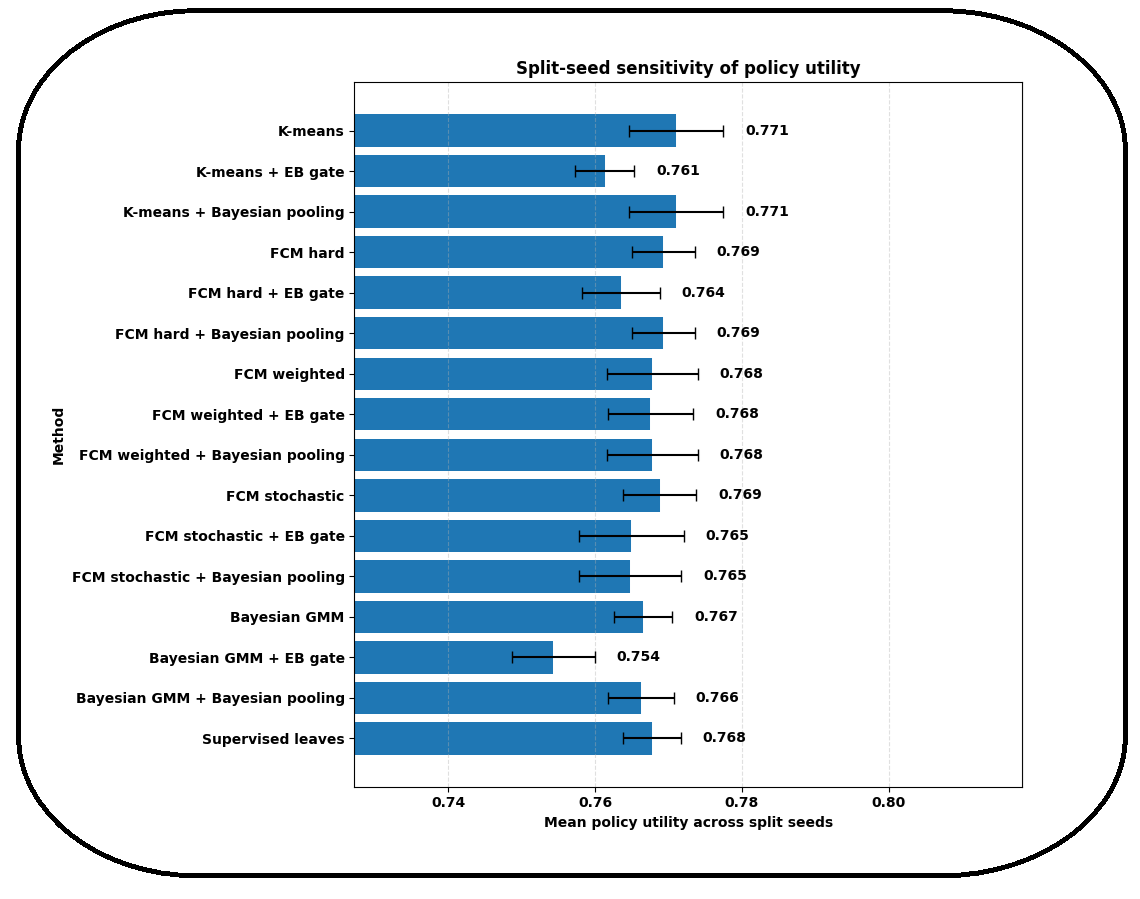}

\caption{\textbf{Smoking-history policy :} Split-seed sensitivity of held-out policy utility across four discovery/evaluation splits. Bars show the mean policy utility across split seeds, and horizontal error bars show the standard deviation across seeds.The ungated unsupervised methods
produced broadly similar mean utilities, with K-means attaining the highest
mean utility, closely followed by hard and stochastic FCM
. Bayesian pooling generally tracked the corresponding ungated
policies, indicating that the posterior gate was non-restrictive or only
weakly restrictive in most runs. Empirical Bernstein gating was more
conservative and generally reduced mean utility, with the largest reduction
observed for Bayesian GMM. The supervised CATE-tree comparator achieved a
comparable utility with several unsupervised methods but
below the best-performing K-means policy. }
\label{fig:smoking_split_seed_utility}

\end{figure*}

\begin{figure*}[!h]
\centering

\begin{minipage}[t]{0.40\textwidth}
\vspace{0pt}
\centering

\includegraphics[
    width=\linewidth
]{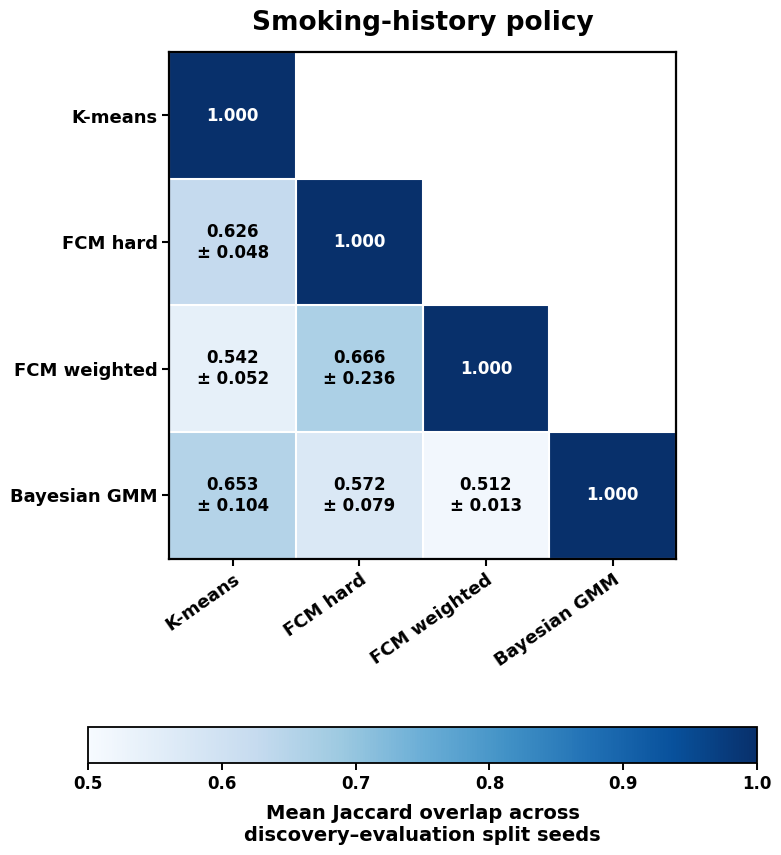}

\captionof{figure}{
\textbf{Smoking-history policy:} Visual summary of the mean pairwise
Jaccard overlap among the fixed ungated subgroup policies across the
examined discovery--evaluation split seeds. Cells report the mean
Jaccard overlap, with the standard deviation shown below the mean.
The diagonal equals one by definition.
}
\label{fig:smoking_jaccard_heatmap}

\end{minipage}
\hfill
\begin{minipage}[t]{0.57\textwidth}
\vspace{0pt}
\centering
\scriptsize

\captionof{table}{
\textbf{Smoking-history policy:} Mean targeted-person overlap metrics
across split seeds. Values are reported as mean $\pm$ standard
deviation. FCM hard and membership-weighted FCM showed the highest
average overlap, whereas membership-weighted FCM and Bayesian GMM
selected the most distinct targeted populations.
}
\label{tab:smoking_mean_targeted_overlap_across_seeds}

\resizebox{\linewidth}{!}{
\begin{tabular}{lccc}
\toprule
Allocation pair
& Jaccard
& \makecell{Targeted-set\\agreement}
& \makecell{Reference-allocation\\coverage} \\
\midrule

K-means vs FCM hard
& $0.6260 \pm 0.0481$
& $0.7692 \pm 0.0365$
& $0.7692 \pm 0.0365$ \\

K-means vs FCM weighted
& $0.5423 \pm 0.0516$
& $0.7022 \pm 0.0446$
& $0.7022 \pm 0.0446$ \\

K-means vs Bayesian GMM
& $0.6529 \pm 0.1039$
& $0.7866 \pm 0.0730$
& $0.7866 \pm 0.0730$ \\

FCM hard vs FCM weighted
& $0.6660 \pm 0.2358$
& $0.7828 \pm 0.1575$
& $0.7828 \pm 0.1575$ \\

FCM hard vs Bayesian GMM
& $0.5724 \pm 0.0787$
& $0.7258 \pm 0.0614$
& $0.7258 \pm 0.0614$ \\

FCM weighted vs Bayesian GMM
& $0.5125 \pm 0.0132$
& $0.6777 \pm 0.0114$
& $0.6777 \pm 0.0114$ \\

\bottomrule
\end{tabular}
}

\end{minipage}

\end{figure*}

The split-seed sensitivity of held-out smoking-policy utility is shown in
Figure~\ref{fig:smoking_split_seed_utility}. K-means attained the highest mean utility, but the difference from the FCM variants was small. Allocation agreement was less uniform than in the glucose experiment and varied appreciably across some split seeds as shown in the Table~\ref{tab:smoking_mean_targeted_overlap_across_seeds} and figure ~\ref{fig:smoking_jaccard_heatmap}. The run-wise overlap between stochastic FCM and the fixed ungated
allocations is summarized in
Table~\ref{tab:smoking_stochastic_fcm_overlap_across_seeds}. Stochastic FCM remained closest to the deterministic FCM policies, while the lowest average fixed-policy overlap occurred between membership-weighted FCM and Bayesian GMM.

\begin{table}[!h]
\centering
\scriptsize

\caption{\textbf{Smoking-history contrast:} Across-seed targeted-person overlap between stochastic FCM and fixed hardened policies. Within each split seed, overlap was computed from the realized binary selections across 500 stochastic-hardening runs and averaged within seed. Values are the mean $\pm$ standard deviation of these seed-level means. Stochastic FCM overlapped most with FCM hard and weighted policies, while overlap with Bayesian GMM and K-means was lower and similar.}

\label{tab:smoking_stochastic_fcm_overlap_across_seeds}

\begin{tabular}{lccc}
\toprule
Reference allocation
& Jaccard
& Targeted-set agreement
& Reference-allocation coverage \\
\midrule

Bayesian GMM
& $0.5423 \pm 0.0051$
& $0.7031 \pm 0.0043$
& $0.7031 \pm 0.0043$ \\

FCM hard
& $0.5566 \pm 0.0134$
& $0.7150 \pm 0.0110$
& $0.7150 \pm 0.0110$ \\

FCM weighted
& $0.5525 \pm 0.0175$
& $0.7116 \pm 0.0145$
& $0.7116 \pm 0.0145$ \\

K-means
& $0.5419 \pm 0.0157$
& $0.7027 \pm 0.0134$
& $0.7027 \pm 0.0134$ \\

\bottomrule
\end{tabular}

\end{table}



\FloatBarrier

\section*{Discussion}

In all three policy experiments, pre-treatment phenotype-based subgrouping yielded similar held-out policy-utility point estimates across K-means, FCM variants, Bayesian GMM, and the supervised CATE-tree comparator. However, the relative ranking of these methods varied by policy and by discovery and evaluation splits. Multiplicity was addressed at two stages. For subgroup admission, the Empirical Bernstein safety gate applied a Bonferroni-style union-bound adjustment across the K subgroup-specific bounds within each clustering solution, with total error probability set at ($\delta$=0.05). For held-out policy comparison, all unique pairwise policy-risk contrasts were evaluated using paired bootstrap resampling of the same individuals, and the resulting (p)-values were jointly adjusted using Holm's procedure. All paired confidence intervals for risk differences included zero, and no comparison remained statistically significant after adjustment.

A lack of statistically significant differences should not be interpreted as evidence of policy equivalence. Formal equivalence assessment requires a prespecified clinically or practically acceptable margin and evidence that the confidence interval for each policy-risk difference lies entirely within the corresponding equivalence bounds \cite{piaggio2006reporting}. Comparisons across the BMI, glucose, and smoking-history experiments were not combined into a single global testing family, as these experiments involved distinct state contrasts, outcomes, eligible populations, and policy estimands. Analyses across discovery and evaluation split seeds, stochastic-hardening draws, safety-gate variants, and allocation-overlap measures were considered descriptive sensitivity analyses rather than independent confirmatory tests. Therefore, the results do not establish statistically detectable differences in held-out risk favoring any single algorithm. Method selection should also account for subgroup interpretability, allocation stability, data representation, and deployment requirements.

Although a wide range of clustering algorithms exists, this comparison was intentionally focused rather than exhaustive. Inductive methods were chosen for their capacity to provide direct and reusable rules for assigning held-out individuals, as well as to represent distinct subgrouping assumptions. K-means served as a hard centroid-based baseline, FCM facilitated the examination of graded memberships and alternative approaches to membership uncertainty, and Bayesian GMM functioned as a probabilistic model-based comparator.

The supervised comparison highlights the distinction between treatment-effect separation and subgroup-level policy performance. As anticipated, the causal-forest-derived interpreting tree produced more effect-homogeneous leaves (see Supplementary Tables S14, S29, and S44) because its partition was explicitly guided by estimated treatment-effect variation. However, increased within-group effect homogeneity did not consistently result in higher held-out utility under the same budget and evaluation procedure. This observation pertains only to the supervised comparator evaluated in this study. Alternative CATE estimators, tree-depth constraints, or distillation procedures could yield different partitions, rankings, and held-out policy utilities. Therefore, these results should not be interpreted as establishing a general advantage of unsupervised subgrouping over supervised HTE methods.

The supervised comparator functioned as a subgroup-level policy rather than a fully individualized treatment rule. Although estimated CATE information was utilized to construct and rank the supervised leaves, eligible individuals were not ranked directly by their individual estimated CATEs. Instead, leaves were ranked according to their aggregate discovery-cohort gain, and a common policy decision was applied to eligible members of each fully selected leaf, with a shared fractional shift probability assigned to the boundary leaf. Consequently, the proposed framework was not compared with an individualized top-(q) policy that ranks all eligible individuals by estimated CATE and allocates the intervention to those with the highest predicted benefit. Such individualized policies may yield different utilities and should be included as separate comparators in future research.

Across the subgroup policies, clusters or leaves were ranked using aggregate gain, defined as the estimated mean benefit multiplied by the eligible subgroup size. This criterion incorporated both the magnitude of the estimated benefit and the number of individuals who could potentially benefit into the prioritization process. It was chosen as a prespecified policy-ranking objective rather than as a uniquely optimal welfare criterion, and alternative ranking measures could result in different subgroup orders and allocations.

The comparable policy utilities observed, despite incomplete targeted-person overlap, are plausible because the competing policies often differed more in the composition of the selected eligible population than in the total estimated benefit allocated. In most experiments, the estimated mean benefit was positive for the majority of eligible subgroups. As a result, the policy problem frequently involved prioritizing among several potentially beneficial phenotypes rather than clearly distinguishing beneficial from harmful groups. Under the common budget, all ungated policies selected the same number of eligible individuals and differed primarily in which individuals were excluded or assigned near the allocation boundary. Consequently, one policy could substitute individuals from one positively ranked subgroup with individuals from another subgroup with a similar estimated average benefit, resulting in notable disagreement in individual targeting but only a minor change in aggregate policy utility.

The fixed 70\% budget served as a common person-level capacity constraint rather than as an all-or-none subgroup rule. Due to unequal subgroup sizes, the budget could necessitate partial allocation within a boundary subgroup. Its relatively high level may have also reduced differences between policies by aligning their allocations more closely with the shift-all-eligible reference. The shift-all-eligible reference represents an unconstrained shift-all-eligible reference policy rather than a clinically viable strategy, as it assumes unlimited operational capacity and disregards intervention-related toxicities or contraindications that would restrict real-world implementation.

Another explanation for the similar utilities is that policy risk was averaged over the entire overlap-restricted evaluation population, rather than solely over the eligible individuals whose allocations differed. Although the hypothetical intervention was applied only to eligible individuals observed in the adverse state, individuals already in the lower-risk state were included in the analysis. During discovery, these individuals contributed to the estimation of the lower-state outcome model used to construct model-based benefit scores. During held-out evaluation, they provided information about the target-state outcome distribution and contributed to the intervention-specific doubly robust score through their observed outcomes and propensity-corrected residuals. Eligible individuals who were not selected for shifting also remained part of the population-level policy-risk estimand.

Because allocation differences were limited to a subset of eligible individuals, their impact on population-level policy utility was relatively modest. Policy value was evaluated using the boundary subgroup's fractional shift probability, while targeted-person overlap was calculated from a realized binary boundary allocation. Random selection within a boundary subgroup could alter which specific individuals appeared in the overlap calculation without affecting the expected subgroup-level policy value. Collectively, these results suggest that multiple alternative allocations may exist on a relatively flat estimated policy-value surface, although this interpretation remains subject to uncertainty in subgroup-benefit estimates and held-out policy values.

The allocation-overlap analyses demonstrate that policy value does not uniquely determine which individuals a policy selects. Methods with similar estimated utilities sometimes prioritized substantially different eligible individuals, particularly in comparisons involving Bayesian GMM. Therefore, policy utility should be interpreted in conjunction with subgroup profiles, allocation overlap, and the practical implications of the selected population. This consideration is especially important when subgroup policies are intended to serve as transparent operational rules rather than solely as numerical value-maximizing procedures.

The directional overlap measures should also be interpreted in relation to the number of individuals selected by each policy. In the ungated comparisons, all policies utilized the same 70\% intervention budget and therefore selected the same number of eligible individuals. As a result, targeted-set agreement and reference-allocation coverage were numerically identical due to equal denominators. However, this equality does not indicate that the policies selected highly similar individuals. The distinction between the two directional measures becomes more informative for gated policies, which may leave part of the available budget unused and thus select different numbers of individuals.

The safety gates embodied different evidentiary standards. Empirical Bernstein gating imposed a finite-sample, multiplicity-aware lower-bound requirement and was frequently restrictive, resulting in outcomes ranging from partial budget use to a no-shift policy. Hierarchical Bayesian pooling borrowed information across subgroups and generally maintained allocations closer to their corresponding ungated policies. Neither approach is universally preferable. A conservative lower-bound rule may be appropriate for costly or potentially harmful interventions, whereas partial pooling may be more suitable for lower-risk preventive, screening, or behavioral actions. However, Bayesian pooling remains conditional on the subgroup estimates and standard errors provided to the hierarchical model and does not propagate all upstream uncertainty arising from causal-graph selection, nuisance estimation, clustering, or sample splitting.

Hard and membership-weighted FCM could yield different allocations even when based on the same fuzzy clustering solution. Membership weighting modified subgroup benefit estimates and aggregate gains, potentially reversing the ordering of clusters with similar scores. These changes could affect the boundary subgroup and targeted population while leaving overall policy utility largely unchanged. Stochastic FCM addressed a distinct source of uncertainty by propagating fuzzy membership ambiguity into held-out allocation. Across the experiments, Monte Carlo variability in policy utility was small relative to bootstrap sampling uncertainty. This finding suggests that aggregate utility was relatively insensitive to plausible changes in fuzzy assignment within the examined splits, but it does not imply that individual memberships or targeting decisions were invariant.

Nevertheless, all causal and policy interpretations remain dependent on underlying assumptions. These interpretations rely on the adequacy of the graph-informed adjustment set, conditional exchangeability, positivity, consistency, and accurate estimation of the nuisance functions. The reported uncertainty intervals also condition on the subgroup policy learned in the discovery cohort and do not account for all sources of uncertainty arising during subgroup discovery and policy construction. These findings support phenotype-first subgrouping as a transparent approach to observational policy prioritization, while emphasizing that method selection should account for estimated performance, subgroup interpretability, allocation composition, uncertainty, and deployment stability.

The main takeaways are:

\begin{itemize}

\item No single subgrouping algorithm consistently dominated.

\item Greater treatment-effect homogeneity did not necessarily
translate into greater policy utility.

\item Safety gates represent different evidentiary standards.

\item Fuzzy membership uncertainty affected allocation more than
aggregate policy value.

\item Similar policy utility did not imply similar targeting.

\item Policy utility alone is insufficient for choosing a policy.

\item Because the analyses are observational and the discovered graphs represent only potential causal links, all causal and policy interpretations remain assumption-dependent and require external, longitudinal, or prospective validation.

\end{itemize}






\section*{Limitations and Future Work}
The datasets analyzed are observational, with each individual observed under only one treatment or exposure state, rendering the corresponding counterfactual outcome unavailable. As a result, all policy-utility estimates rely on causal identification assumptions, including consistency, conditional exchangeability, and positivity. Conditional exchangeability is contingent upon the assumed potential causal structure, the adequacy of the selected adjustment set, and the absence of significant unmeasured confounding. Violations of these assumptions may introduce bias into the estimated individual benefit scores, subgroup rankings, and policy utilities. Additionally, restricting our analysis to complete cases imposes the Missing Completely at Random (MCAR) assumption, thereby compounding the foundational requirements for causal inference. If the missingness mechanism is related to unobserved confounders, this restriction may introduce selection bias and further compromise unconfoundedness. Although held-out policy value is evaluated using a doubly robust estimator, this approach does not mitigate bias arising from unmeasured confounding or other failures of causal identification. Because ground-truth individual treatment effects and optimal treatment assignments are unavailable, individual-effect estimation cannot be assessed using metrics such as ITE mean squared error or PEHE. Instead, marginal bootstrap confidence intervals are reported for each held-out policy-value estimate.

The policies represent hypothetical state contrasts, such as obese versus non-obese, elevated versus lower glucose, and lifetime-smoking-history-positive versus non-smoker states. These contrasts may inform prioritization but do not specify the intervention by which the state would be altered. For example, the BMI policy should not be interpreted as the expected value of a deployed weight-loss intervention, as BMI is not a well-defined intervention and different mechanisms for changing BMI may have distinct causal effects \cite{hernan2008does}. Similarly, the lifetime-smoking-history contrast does not correspond to an acute smoking-cessation intervention, since past smoking exposure cannot be retrospectively changed. The reported utilities should therefore be interpreted as state-contrast-based prioritization estimands rather than as the effects of specific clinical programs. The subgroup policy assigns a common decision to eligible individuals within the same selected cluster, enhancing transparency and operational simplicity but potentially sacrificing individual-level precision, particularly for individuals near cluster boundaries or whose estimated benefit diverges substantially from the subgroup average. The framework first estimates individual model-based benefit scores and then averages them within subgroups, thereby avoiding potentially unstable direct treated-versus-untreated comparisons within small or imbalanced clusters. However, this approach makes subgroup ranking dependent on the accuracy of the benefit-estimation models. Since benefit scores were estimated using separate logistic generalized linear models (GLMs), unmodeled nonlinear response patterns may distort the estimated state contrasts, potentially affecting subgroup rankings and policy allocations \cite{GLM_limitation}. Several design choices were implemented to support controlled comparisons across subgrouping methods. The upper component bound of the Bayesian Gaussian Mixture Model (GMM) was restricted to the common reference number of clusters used by K-means and FCM, improving comparability across clustering families but limiting the Bayesian GMM's ability to select a substantially different mixture complexity. In the NHANES smoking-history analysis, the supervised CATE tree was fitted using the same MCA representation as the unsupervised methods to ensure a representation-aligned comparison. However, splits on latent MCA dimensions are less directly interpretable than rules based on the original categorical variables and require interpretation through the corresponding category contributions \cite{MCA_limit}. Some demographic attributes, including sex and race/ethnicity, were used for
both subgroup construction and confounding adjustment. Consequently, the
resulting NHANES policies may depend on these attributes and should be viewed
as methodological rather than deployment-ready.

Future research should compare the proposed subgroup policy directly with individualized CATE-based policies, including causal forests, Bayesian causal forests, and distilled student policies. Such comparisons would quantify the tradeoff between subgroup-level interpretability and individualized policy utility. Causal-response clustering approaches, such as causal K-means \cite{causalkmeans}, should also be included as effect-guided subgrouping comparators. Semi-synthetic experiments that preserve realistic covariate distributions while generating known counterfactual outcomes would further enable evaluation using both policy utility and ground-truth ITE metrics such as PEHE or mean squared error. Several methodological extensions warrant examination. For categorical data, simultaneous dimension-reduction and clustering approaches, such as Multiple Correspondence K-means, may address limitations associated with sequential tandem analysis \cite{MCA_PAPER6}. Future experiments could also permit Bayesian GMMs to use larger truncation levels, allowing the effective number of active components to be inferred more flexibly. Additional inductive or approximately inductive clustering methods, including Balanced Iterative Reducing and Clustering using Hierarchies (BIRCH), should be considered for larger observational cohorts \cite{BIRCH}. Another direction is to evaluate the framework in clinical settings where subgroup-level decisions more closely resemble actual care pathways. Oncology applications involving standard versus escalated therapy, radiation versus observation, or risk-adapted treatment protocols may be particularly relevant. Future work should also examine
fairness-constrained policies that retain sensitive attributes for adjustment
but exclude them from the deployed rule. Finally, the current framework considers only single-time state contrasts. Extending it to longitudinal settings with time-varying treatments, covariates, and outcomes would require dynamic treatment-regime or sequential policy-learning methods.


\section*{Conclusion}
This study examined whether pretreatment phenotype-based subgroups can provide interpretable units for budget-constrained policy prioritization in observational data. Across the BMI, glucose, and smoking-history experiments, the unsupervised subgrouping methods generally produced held-out policy utilities comparable in magnitude to those of the supervised CATE-tree comparator. Although the supervised tree generated greater within-group effect homogeneity, this did not consistently translate into higher held-out policy utility. Paired bootstrap comparisons did not identify statistically detectable policy-risk differences after Holm adjustment. Accordingly, the policies with the most favorable point estimates should be interpreted as descriptive leaders within the representative splits rather than as demonstrably superior policies. Consistent with this interpretation, no single clustering algorithm uniformly achieved the most favorable point estimate across the three policy experiments. The results also demonstrate that similar estimated policy utility does not necessarily imply similar allocation. Different subgrouping methods sometimes prioritized meaningfully different eligible individuals, while stochastic FCM indicated that fuzzy membership uncertainty introduced only modest variation in aggregate policy utility in the examined experiments. Empirical Bernstein gating was substantially more conservative than hierarchical Bayesian pooling, illustrating that the choice of safety gate represents a context-dependent decision regarding evidentiary stringency and intervention risk. These findings were obtained under a fixed 70\% intervention budget, and both relative policy performance and allocation composition may differ under alternative budget constraints. Moreover, because the analyses were observational and the policies represented hypothetical state contrasts, all estimates remain conditional on the stated causal-identification and modeling assumptions. External and prospective validation using clinically defined and implementable interventions is therefore required before these policies can support real-world decision-making.


\section*{Author contributions statement}

V.A. and B.Y conceived the experiments. V.A. conducted the experiments, V.A. and B.Y. analyzed the results.  All authors reviewed the manuscript

\section*{Competing interests}
The authors declare no competing interests.

\section*{Additional information} Supplementary Information is provided with this manuscript. 

\section*{Acknowledgements}
The authors express their gratitude to Madeline C. Fields and Lara V. Marcuse of the Icahn School of Medicine at Mount Sinai and the Mount Sinai Hospital, New York, NY, USA, for their assistance in elucidating potential causal relationships relevant to this study. Appreciation is also extended to Dr. Vishwas D. Pai of Pai Onco Care Center and Hubli Scan Center, India, for his valuable guidance and support.

\clearpage

\bibliography{sample}

@article{hernan2008does,
  title={Does obesity shorten life? The importance of well-defined interventions to answer causal questions},
  author={Hern{\'a}n, Miguel A and Taubman, Sarah L},
  journal={International journal of obesity},
  volume={32},
  number={3},
  pages={S8--S14},
  year={2008},
  publisher={Nature Publishing Group}
}

@inproceedings{pima_diabetes_dataset,
  title={Using the ADAP learning algorithm to forecast the onset of diabetes mellitus},
  author={Smith, Jack W and Everhart, James E and Dickson, William C and Knowler, William C and Johannes, Robert Scott},
  booktitle={Proceedings of the annual symposium on computer application in medical care},
  pages={261},
  year={1988}
}

@article{gelman2006prior,
  title={Prior distributions for variance parameters in hierarchical models},
  author={Gelman, Andrew},
  journal={Bayesian Analysis},
  volume={1},
  number={3},
  pages={515--533},
  year={2006}
}

@article{polson2012half,
  title={On the half-Cauchy prior for a global scale parameter},
  author={Polson, Nicholas G. and Scott, James G.},
  journal={Bayesian Analysis},
  volume={7},
  number={4},
  pages={887--902},
  year={2012}
}

@article{morris1983parametric,
  title={Parametric Empirical Bayes Inference: Theory and Applications},
  author={Morris, Carl N.},
  journal={Journal of the American Statistical Association},
  volume={78},
  number={381},
  pages={47--55},
  year={1983}
}

@article{bernstein,
  title={Empirical bernstein bounds and sample variance penalization},
  author={Maurer, Andreas and Pontil, Massimiliano},
  journal={arXiv preprint arXiv:0907.3740},
  year={2009}
}

@article{ullmann2022validation,
  title={Validation of cluster analysis results on validation data: A systematic framework},
  author={Ullmann, Theresa and Hennig, Christian and Boulesteix, Anne-Laure},
  journal={Wiley Interdisciplinary Reviews: Data Mining and Knowledge Discovery},
  volume={12},
  number={3},
  pages={e1444},
  year={2022},
  publisher={Wiley Online Library}
}

@article{foster2011subgroup,
  title={Subgroup identification from randomized clinical trial data},
  author={Foster, Jared C and Taylor, Jeremy MG and Ruberg, Stephen J},
  journal={Statistics in medicine},
  volume={30},
  number={24},
  pages={2867--2880},
  year={2011},
  publisher={Wiley Online Library}
}

@article{jaccard1901etude,
  title={{\'E}tude comparative de la distribution florale dans une portion des Alpes et des Jura},
  author={Jaccard, Paul},
  journal={Bull Soc Vaudoise Sci Nat},
  volume={37},
  pages={547--579},
  year={1901}
}

@inproceedings{macqueen1967some,
  title={Some methods for classification and analysis of multivariate observations},
  author={MacQueen, James},
  booktitle={Proceedings of the Fifth Berkeley Symposium on Mathematical Statistics and Probability},
  volume={1},
  pages={281--297},
  year={1967},
  publisher={University of California Press}
}

@article{dunn1973fuzzy,
  title={A fuzzy relative of the ISODATA process and its use in detecting compact well-separated clusters},
  author={Dunn, Joseph C.},
  journal={Journal of Cybernetics},
  volume={3},
  number={3},
  pages={32--57},
  year={1973}
}

@article{bezdek1984fcm,
  title={FCM: The fuzzy c-means clustering algorithm},
  author={Bezdek, James C. and Ehrlich, Robert and Full, William},
  journal={Computers \& Geosciences},
  volume={10},
  number={2--3},
  pages={191--203},
  year={1984},
  doi={10.1016/0098-3004(84)90020-7}
}

@book{bezdek1981pattern,
  title={Pattern Recognition with Fuzzy Objective Function Algorithms},
  author={Bezdek, James C.},
  publisher={Plenum Press},
  address={New York},
  year={1981}
}

@article{wu2012analysis,
  title={Analysis of parameter selections for fuzzy c-means},
  author={Wu, Kuo-Lung},
  journal={Pattern Recognition},
  volume={45},
  number={1},
  pages={407--415},
  year={2012}
}

@misc{wu2026statistical,
  title={Statistical Inference for Fuzzy Clustering},
  author={Wu, Qiuyi and Zhu, Zihan and Zhang, Anru R.},
  year={2026},
  eprint={2601.02656},
  archivePrefix={arXiv},
  primaryClass={stat.ME}
}

@inproceedings{rasmussen2000infinite,
  title={The Infinite Gaussian Mixture Model},
  author={Rasmussen, Carl Edward},
  booktitle={Advances in Neural Information Processing Systems},
  volume={12},
  pages={554--560},
  year={2000},
  publisher={MIT Press}
}

@article{blei2006variational,
  title={Variational inference for Dirichlet process mixtures},
  author={Blei, David M. and Jordan, Michael I.},
  journal={Bayesian Analysis},
  volume={1},
  number={1},
  pages={121--144},
  year={2006}
}

@article{fuzzycmeans2,
  title={A simple and fast method to determine the parameters for fuzzy c--means cluster analysis},
  author={Schw{\"a}mmle, Veit and Jensen, Ole N{\o}rregaard},
  journal={Bioinformatics},
  volume={26},
  number={22},
  pages={2841--2848},
  year={2010},
  publisher={Oxford University Press}
}

@article{hill2011bayesian,
  title={Bayesian nonparametric modeling for causal inference},
  author={Hill, Jennifer L.},
  journal={Journal of Computational and Graphical Statistics},
  volume={20},
  number={1},
  pages={217--240},
  year={2011},
  publisher={Taylor \& Francis}
}

@article{athey2021policy,
  title={Policy Learning With Observational Data},
  author={Athey, Susan and Wager, Stefan},
  journal={Econometrica},
  volume={89},
  number={1},
  pages={133--161},
  year={2021},
  publisher={Wiley}
}

@article{lindeman2018updated,
  title={Updated molecular testing guideline for the selection of lung cancer patients for treatment with targeted tyrosine kinase inhibitors: guideline from the College of American Pathologists, the International Association for the Study of Lung Cancer, and the Association for Molecular Pathology},
  author={Lindeman, Neal I and Cagle, Philip T and Aisner, Dara L and Arcila, Maria E and Beasley, Mary Beth and Bernicker, Eric H and Colasacco, Carol and Dacic, Sanja and Hirsch, Fred R and Kerr, Keith and others},
  journal={Archives of pathology \& laboratory medicine},
  volume={142},
  number={3},
  pages={321--346},
  year={2018},
  publisher={the College of American Pathologists}
}

@article{sepulveda2017molecular,
  title={Molecular biomarkers for the evaluation of colorectal cancer: guideline from the American Society for Clinical Pathology, College of American Pathologists, Association for Molecular Pathology, and American Society of Clinical Oncology},
  author={Sepulveda, Antonia R and Hamilton, Stanley R and Allegra, Carmen J and Grody, Wayne and Cushman-Vokoun, Allison M and Funkhouser, William K and Kopetz, Scott E and Lieu, Christopher and Lindor, Noralane M and Minsky, Bruce D and others},
  journal={American journal of clinical pathology},
  volume={147},
  number={3},
  pages={221--260},
  year={2017},
  publisher={Oxford University Press}
}

@article{BIRCH,
  title={BIRCH: an efficient data clustering method for very large databases},
  author={Zhang, Tian and Ramakrishnan, Raghu and Livny, Miron},
  journal={ACM sigmod record},
  volume={25},
  number={2},
  pages={103--114},
  year={1996},
  publisher={ACM New York, NY, USA}
}

@article{HTE_P1,
  title={Assessing and reporting heterogeneity in treatment effects in clinical trials: a proposal},
  author={Kent, David M and Rothwell, Peter M and Ioannidis, John PA and Altman, Doug G and Hayward, Rodney A},
  journal={Trials},
  volume={11},
  number={1},
  pages={85},
  year={2010},
  publisher={Springer}
}

@article{HTE_P2,
  title={Using group data to treat individuals: understanding heterogeneous treatment effects in the age of precision medicine and patient-centred evidence},
  author={Dahabreh, Issa J and Hayward, Rodney and Kent, David M},
  journal={International journal of epidemiology},
  volume={45},
  number={6},
  pages={2184--2193},
  year={2016},
  publisher={Oxford University Press}
}

@article{HTE_P3,
  title={Recursive partitioning for heterogeneous causal effects},
  author={Athey, Susan and Imbens, Guido},
  journal={Proceedings of the National Academy of Sciences},
  volume={113},
  number={27},
  pages={7353--7360},
  year={2016},
  publisher={National Academy of Sciences}
}

@article{zhao2026mapping,
  title={Mapping phenotypic heterogeneity and cardiometabolic risk in obesity using a tree-based dimensionality reduction framework},
  author={Zhao, Yan and Wang, Weihao and Wang, Zihao and Ma, Jing and Zhang, Jia and Shao, Jian and Zhou, Kaixin and Pan, Qi and Nie, Zedong and Xu, Guogang and others},
  journal={Journal of Translational Medicine},
  volume={24},
  number={1},
  pages={401},
  year={2026},
  publisher={Springer}
}

@article{HTE_P4,
  title={Bayesian analysis of heterogeneous treatment effects for patient-centered outcomes research},
  author={Henderson, Nicholas C and Louis, Thomas A and Wang, Chenguang and Varadhan, Ravi},
  journal={Health Services and Outcomes Research Methodology},
  volume={16},
  number={4},
  pages={213--233},
  year={2016},
  publisher={Springer}
}

@article{virtualtwins,
  title={Subgroup identification from randomized clinical trial data},
  author={Foster, Jared C and Taylor, Jeremy MG and Ruberg, Stephen J},
  journal={Statistics in medicine},
  volume={30},
  number={24},
  pages={2867--2880},
  year={2011},
  publisher={Wiley Online Library}
}

@article{causalforest,
  title={Estimation and inference of heterogeneous treatment effects using random forests},
  author={Wager, Stefan and Athey, Susan},
  journal={Journal of the American Statistical Association},
  volume={113},
  number={523},
  pages={1228--1242},
  year={2018},
  publisher={Taylor \& Francis}
}

@techreport{GATES,
  title={Generic machine learning inference on heterogeneous treatment effects in randomized experiments, with an application to immunization in India},
  author={Chernozhukov, Victor and Demirer, Mert and Duflo, Esther and Fernandez-Val, Ivan},
  year={2018},
  institution={National Bureau of Economic Research}
}

@article{causalkmeans,
  title={Causal K-means clustering},
  author={Kim, Kwangho and Kim, Jisu and Kennedy, Edward H},
  journal={Journal of the Royal Statistical Society Series B: Statistical Methodology},
  pages={qkag068},
  year={2026},
  publisher={Oxford University Press UK}
}

@inproceedings{wang2025causal,
  title={Causal Clustering for Conditional Average Treatment Effects Estimation and Subgroup Discovery},
  author={Wang, Zilong and Ayer, Turgay and Yang, Shihao},
  booktitle={2025 IEEE EMBS International Conference on Biomedical and Health Informatics (BHI)},
  pages={1-11},
  year={2025},
  organization={IEEE}
}

@inproceedings{thomas2015high,
  title={High-confidence off-policy evaluation},
  author={Thomas, Philip and Theocharous, Georgios and Ghavamzadeh, Mohammad},
  booktitle={Proceedings of the AAAI Conference on Artificial Intelligence},
  volume={29},
  number={1},
  year={2015},
}

@article{jin2025policy,
  title={Policy learning “without” overlap: Pessimism and generalized empirical Bernstein’s inequality},
  author={Jin, Ying and Ren, Zhimei and Yang, Zhuoran and Wang, Zhaoran},
  journal={The Annals of Statistics},
  volume={53},
  number={4},
  pages={1483--1512},
  year={2025},
  publisher={Institute of Mathematical Statistics}
}

@article{gelman2012we,
  title={Why we (usually) don't have to worry about multiple comparisons},
  author={Gelman, Andrew and Hill, Jennifer and Yajima, Masanao},
  journal={Journal of research on educational effectiveness},
  volume={5},
  number={2},
  pages={189--211},
  year={2012},
  publisher={Taylor \& Francis}
}

@inproceedings{shalit2017,
  title={Estimating individual treatment effect: generalization bounds and algorithms},
  author={Shalit, Uri and Johansson, Fredrik D and Sontag, David},
  booktitle={International conference on machine learning},
  pages={3076--3085},
  year={2017},
  organization={PMLR}
}

@article{holland1986statistics,
  title={Statistics and causal inference},
  author={Holland, Paul W},
  journal={Journal of the American statistical Association},
  volume={81},
  number={396},
  pages={945--960},
  year={1986},
  publisher={Taylor \& Francis}
}

@article{schisterman2009overadjustment,
  title={Overadjustment bias and unnecessary adjustment in epidemiologic studies},
  author={Schisterman, Enrique F and Cole, Stephen R and Platt, Robert W},
  journal={Epidemiology},
  volume={20},
  number={4},
  pages={488--495},
  year={2009},
  publisher={LWW}
}

@article{kunzel2019metalearners,
  title={Metalearners for estimating heterogeneous treatment effects using machine learning},
  author={K{\"u}nzel, S{\"o}ren R and Sekhon, Jasjeet S and Bickel, Peter J and Yu, Bin},
  journal={Proceedings of the national academy of sciences},
  volume={116},
  number={10},
  pages={4156--4165},
  year={2019},
  publisher={National Academy of Sciences}
}

@article{vanderweele2019principles,
  title={Principles of confounder selection: TJ VanderWeele},
  author={VanderWeele, Tyler J},
  journal={European journal of epidemiology},
  volume={34},
  number={3},
  pages={211--219},
  year={2019},
  publisher={Springer}
}

@article{su2015hierarchical,
  title={A hierarchical fuzzy cluster ensemble approach and its application to big data clustering},
  author={Su, Pan and Shang, Changjing and Shen, Qiang},
  journal={Journal of Intelligent \& Fuzzy Systems},
  volume={28},
  number={6},
  pages={2409--2421},
  year={2015},
  publisher={SAGE Publications Sage UK: London, England}
}

@article{nie2021quasi,
  title={Quasi-oracle estimation of heterogeneous treatment effects},
  author={Nie, Xinkun and Wager, Stefan},
  journal={Biometrika},
  volume={108},
  number={2},
  pages={299--319},
  year={2021},
  publisher={Oxford University Press}
}

@article{bellavia2025unsupervised,
  title={Unsupervised clustering approach to assess heterogeneity of treatment effects across patient phenotypes in randomized clinical trials},
  author={Bellavia, Andrea and Ran, Xinhui and Zimerman, Andre and Antman, Elliott M and Giugliano, Robert P and Morrow, David A and Murphy, Sabina A},
  journal={Contemporary Clinical Trials},
  volume={148},
  pages={107778},
  year={2025},
  publisher={Elsevier}
}

@article{hahn2020bayesian,
  title={Bayesian regression tree models for causal inference: Regularization, confounding, and heterogeneous effects (with discussion)},
  author={Hahn, P Richard and Murray, Jared S and Carvalho, Carlos M},
  journal={Bayesian Analysis},
  volume={15},
  number={3},
  pages={965--1056},
  year={2020},
  publisher={International Society for Bayesian Analysis}
}

@article{sinha2021comparison,
  title={Comparison of machine learning clustering algorithms for detecting heterogeneity of treatment effect in acute respiratory distress syndrome: a secondary analysis of three randomised controlled trials},
  author={Sinha, Pratik and Spicer, Alexandra and Delucchi, Kevin L and McAuley, Daniel F and Calfee, Carolyn S and Churpek, Matthew M},
  journal={EBioMedicine},
  volume={74},
  year={2021},
  publisher={Elsevier}
}

@article{bhatraju2019identification,
  title={Identification of acute kidney injury subphenotypes with differing molecular signatures and responses to vasopressin therapy},
  author={Bhatraju, Pavan K and Zelnick, Leila R and Herting, Jerald and Katz, Ronit and Mikacenic, Carmen and Kosamo, Susanna and Morrell, Eric D and Robinson-Cohen, Cassianne and Calfee, Carolyn S and Christie, Jason D and others},
  journal={American journal of respiratory and critical care medicine},
  volume={199},
  number={7},
  pages={863--872},
  year={2019},
  publisher={American Thoracic Society}
}

@article{zampieri2019heterogeneous,
  title={Heterogeneous effects of alveolar recruitment in acute respiratory distress syndrome: a machine learning reanalysis of the Alveolar Recruitment for Acute Respiratory Distress Syndrome Trial},
  author={Zampieri, Fernando G and Costa, Eduardo L and Iwashyna, Theodore J and Carvalho, Carlos RR and Damiani, Lucas P and Taniguchi, Leandro U and Amato, Marcelo BP and Cavalcanti, Alexandre B and others},
  journal={British journal of anaesthesia},
  volume={123},
  number={1},
  pages={88--95},
  year={2019},
  publisher={Elsevier}
}

@article{wen2023intervention,
  title={Intervention treatment distributions that depend on the observed treatment process and model double robustness in causal survival analysis},
  author={Wen, Lan and Marcus, Julia L and Young, Jessica G},
  journal={Statistical methods in medical research},
  volume={32},
  number={3},
  pages={509--523},
  year={2023},
  publisher={SAGE Publications Sage UK: London, England}
}

@article{psbart,
  title={K-Fold Causal BART for CATE Estimation},
  author={Souto, Hugo Gobato and Neto, Francisco Louzada},
  journal={arXiv preprint arXiv:2409.05665},
  year={2024}
}

@article{dudik2011doubly,
  title={Doubly robust policy evaluation and learning},
  author={Dud{\'\i}k, Miroslav and Langford, John and Li, Lihong},
  journal={arXiv preprint arXiv:1103.4601},
  year={2011}
}

@inproceedings{policy_p4,
  title={Doubly robust distributionally robust off-policy evaluation and learning},
  author={Kallus, Nathan and Mao, Xiaojie and Wang, Kaiwen and Zhou, Zhengyuan},
  booktitle={International Conference on Machine Learning},
  pages={10598--10632},
  year={2022},
  organization={PMLR}
}

@inproceedings{policy_p5,
  title={More efficient off-policy evaluation through regularized targeted learning},
  author={Bibaut, Aurelien and Malenica, Ivana and Vlassis, Nikos and Van Der Laan, Mark},
  booktitle={International Conference on Machine Learning},
  pages={654--663},
  year={2019},
  organization={PMLR}
}

@article{policy_p6,
  title={Doubly robust estimation in missing data and causal inference models},
  author={Bang, Heejung and Robins, James M},
  journal={Biometrics},
  volume={61},
  number={4},
  pages={962--973},
  year={2005},
  publisher={Oxford University Press}
}

@article{kitagawa2018treated,
  title={Who should be treated? empirical welfare maximization methods for treatment choice},
  author={Kitagawa, Toru and Tetenov, Aleksey},
  journal={Econometrica},
  volume={86},
  number={2},
  pages={591--616},
  year={2018},
  publisher={Wiley Online Library}
}

@article{clustering1,
  title={Explicit learning curves for transduction and application to clustering and compression algorithms},
  author={Derbeko, Philip and El-Yaniv, Ran and Meir, Ron},
  journal={Journal of Artificial Intelligence Research},
  volume={22},
  pages={117--142},
  year={2004}
}

@inproceedings{clustering2,
  title={Inductive vs. transductive clustering using kernel functions and pairwise constraints},
  author={Miyamoto, Sadaaki and Terami, Akihisa},
  booktitle={2011 11th International Conference on Intelligent Systems Design and Applications},
  pages={1258--1264},
  year={2011},
  organization={IEEE}
}

@article{clustering4,
  title={Out-of-sample extensions for lle, isomap, mds, eigenmaps, and spectral clustering},
  author={Bengio, Yoshua and Paiement, Jean-fran{\c{c}}cois and Vincent, Pascal and Delalleau, Olivier and Roux, Nicolas and Ouimet, Marie},
  journal={Advances in neural information processing systems},
  volume={16},
  year={2003}
}

@article{clustering8,
  title={An implementation of the HDBSCAN* clustering algorithm},
  author={Stewart, Geoffrey and Al-Khassaweneh, Mahmood},
  journal={Applied Sciences},
  volume={12},
  number={5},
  pages={2405},
  year={2022},
  publisher={MDPI}
}

@article{hoyer2008nonlinear,
  title={Nonlinear causal discovery with additive noise models},
  author={Hoyer, Patrik and Janzing, Dominik and Mooij, Joris M and Peters, Jonas and Sch{\"o}lkopf, Bernhard},
  journal={Advances in neural information processing systems},
  volume={21},
  year={2008}
}

@inproceedings{acharya2025understanding,
  title={Understanding the Impact of Epilepsy and Depression on Sleep Disorder: Beyond Associations},
  author={Acharya, Vasundhara and Yener, B{\"u}lent and Fields, Madeline C and Marcuse, Lara V},
  booktitle={2025 IEEE EMBS International Conference on Biomedical and Health Informatics (BHI)},
  pages={1--7},
  year={2025},
  organization={IEEE}
}

@inproceedings{spirtes2001anytime,
  title={An anytime algorithm for causal inference},
  author={Spirtes, Peter},
  booktitle={International Workshop on Artificial Intelligence and Statistics},
  pages={278--285},
  year={2001},
  organization={PMLR}
}

@inproceedings{zheng2020learning,
  title={Learning sparse nonparametric dags},
  author={Zheng, Xun and Dan, Chen and Aragam, Bryon and Ravikumar, Pradeep and Xing, Eric},
  booktitle={International conference on artificial intelligence and statistics},
  pages={3414--3425},
  year={2020},
  organization={Pmlr}
}

@article{cheng2022age,
  title={Age-related changes in the risk of high blood pressure},
  author={Cheng, Weibin and Du, Yumeng and Zhang, Qingpeng and Wang, Xin and He, Chaocheng and He, Jingjun and Jing, Fengshi and Ren, Hao and Guo, Mengzhuo and Tian, Junzhang and others},
  journal={Frontiers in cardiovascular medicine},
  volume={9},
  pages={939103},
  year={2022},
  publisher={Frontiers Media SA}
}

@article{li2016mechanisms,
  title={Mechanisms in endocrinology: parity and risk of type 2 diabetes: a systematic review and dose-response meta-analysis},
  author={Li, Peiyun and Shan, Zhilei and Zhou, Li and Xie, Manling and Bao, Wei and Zhang, Yan and Rong, Ying and Yang, Wei and Liu, Liegang},
  journal={European journal of endocrinology},
  volume={175},
  number={5},
  pages={R231--R245},
  year={2016},
  publisher={Oxford University Press}
}

@article{ruiz2020skinfold,
  title={Skinfold thickness and the incidence of type 2 diabetes mellitus and hypertension: an analysis of the PERU MIGRANT study},
  author={Ruiz-Alejos, Andrea and Carrillo-Larco, Rodrigo M and Miranda, J Jaime and Gilman, Robert H and Smeeth, Liam and Bernab{\'e}-Ortiz, Antonio},
  journal={Public health nutrition},
  volume={23},
  number={1},
  pages={63--71},
  year={2020},
  publisher={Cambridge University Press}
}

@article{emdin2015usual,
  title={Usual blood pressure and risk of new-onset diabetes: evidence from 4.1 million adults and a meta-analysis of prospective studies},
  author={Emdin, Connor A and Anderson, Simon G and Woodward, Mark and Rahimi, Kazem},
  journal={Journal of the American College of Cardiology},
  volume={66},
  number={14},
  pages={1552--1562},
  year={2015},
  publisher={American College of Cardiology Foundation Washington, DC}
}

@article{fazeli2020aging,
  title={Aging is a powerful risk factor for type 2 diabetes mellitus independent of body mass index},
  author={Fazeli, Pouneh K and Lee, Hang and Steinhauser, Matthew L},
  journal={Gerontology},
  volume={66},
  number={2},
  pages={209--210},
  year={2020},
  publisher={S. Karger AG}
}

@inproceedings{pillai_test_ankur,
  title={A simple unified approach to testing high-dimensional conditional independences for categorical and ordinal data},
  author={Ankan, Ankur and Textor, Johannes},
  booktitle={Proceedings of the AAAI Conference on Artificial Intelligence},
  volume={37},
  number={10},
  pages={12180--12188},
  year={2023}
}

@article{PC,
  title={An algorithm for fast recovery of sparse causal graphs},
  author={Spirtes, Peter and Glymour, Clark},
  journal={Social science computer review},
  volume={9},
  number={1},
  pages={62--72},
  year={1991},
  publisher={Sage Publications Sage CA: Thousand Oaks, CA}
}

@article{hill_climb,
  title={Learning directed acyclic graphs via bootstrap aggregating},
  author={Wang, Ru and Peng, Jie},
  journal={arXiv preprint arXiv:1406.2098},
  year={2014}
}

@article{fast_ges,
  title={A million variables and more: the fast greedy equivalence search algorithm for learning high-dimensional graphical causal models, with an application to functional magnetic resonance images},
  author={Ramsey, Joseph and Glymour, Madelyn and Sanchez-Romero, Ruben and Glymour, Clark},
  journal={International journal of data science and analytics},
  volume={3},
  pages={121--129},
  year={2017},
  publisher={Springer}
}

@article{powers2020evaluation,
  title={Evaluation: from precision, recall and F-measure to ROC, informedness, markedness and correlation},
  author={Powers, David MW},
  journal={arXiv preprint arXiv:2010.16061},
  year={2020}
}

@article{jaccard,
  title={Graph comparison for causal discovery},
  author={Cottam, Joseph and Glenski, Maria and Shaw, Yi and Rabello, Ryan and Golding, Austin and Volkova, Svitlana and Arendt, Dustin},
  journal={Visualization in data science},
  year={2021}
}

@misc{purnell2015definitions,
  title={Definitions, classification, and epidemiology of obesity},
  author={Purnell, Jonathan Q},
  year={2015},
}

@article{bansal2015prediabetes,
  title={Prediabetes diagnosis and treatment: A review},
  author={Bansal, Nidhi},
  journal={World journal of diabetes},
  volume={6},
  number={2},
  pages={296},
  year={2015}
}

@article{wang2007statistics,
  title={Statistics in medicine—reporting of subgroup analyses in clinical trials},
  author={Wang, Rui and Lagakos, Stephen W and Ware, James H and Hunter, David J and Drazen, Jeffrey M},
  journal={New England Journal of Medicine},
  volume={357},
  number={21},
  pages={2189--2194},
  year={2007},
  publisher={Mass Medical Soc}
}

@book{greenacre1984theory,
  author    = {Greenacre, Michael J.},
  title     = {Theory and Applications of Correspondence Analysis},
  publisher = {Academic Press},
  year      = {1984}
}

@book{greenacre2006multiple,
  editor    = {Greenacre, Michael J. and Blasius, J{\"o}rg},
  title     = {Multiple Correspondence Analysis and Related Methods},
  publisher = {Chapman and Hall/CRC},
  year      = {2006}
}

@article{abdi2007multiple,
  author  = {Abdi, Herv{\'e} and Valentin, Dominique},
  title   = {Multiple Correspondence Analysis},
  journal = {Encyclopedia of Measurement and Statistics},
  year    = {2007}
}

@article{nenadic2007correspondence,
  author  = {Nenadi{\'c}, Oleg and Greenacre, Michael},
  title   = {Correspondence Analysis in R, with Two- and Three-dimensional Graphics: The ca Package},
  journal = {Journal of Statistical Software},
  volume  = {20},
  number  = {3},
  pages   = {1--13},
  year    = {2007}
}

@article{MCA_PAPER1,
  title={Use of multiple correspondence analysis and K-means to explore associations between risk factors and likelihood of colorectal cancer: Cross-sectional study},
  author={Florensa, D{\'\i}dac and Mateo-Forn{\'e}s, Jordi and Solsona, Francesc and Pedrol Aige, Teresa and Mesas Julio, Miquel and Pi{\~n}ol, Ramon and Godoy, Pere},
  journal={Journal of Medical Internet Research},
  volume={24},
  number={7},
  pages={e29056},
  year={2022},
  publisher={JMIR Publications Toronto, Canada}
}

@article{MCA_PAPER2,
  title={The use of multiple correspondence analysis to explore associations between categories of qualitative variables in healthy ageing},
  author={Costa, Patr{\'\i}cio Soares and Santos, Nadine Correia and Cunha, Pedro and Cotter, Jorge and Sousa, Nuno},
  journal={Journal of aging research},
  volume={2013},
  number={1},
  pages={302163},
  year={2013},
  publisher={Wiley Online Library}
}

@article{MCA_PAPER3,
  title={Correspondence analysis is a useful tool to uncover the relationships among categorical variables},
  author={Sourial, Nadia and Wolfson, Christina and Zhu, Bin and Quail, Jacqueline and Fletcher, John and Karunananthan, Sathya and Bandeen-Roche, Karen and B{\'e}land, Fran{\c{c}}ois and Bergman, Howard},
  journal={Journal of clinical epidemiology},
  volume={63},
  number={6},
  pages={638--646},
  year={2010},
  publisher={Elsevier}
}

@article{MCA_PAPER4,
  title={Multimorbidity patterns with K-means nonhierarchical cluster analysis},
  author={Viol{\'a}n, Concepci{\'o}n and Roso-Llorach, Albert and Foguet-Boreu, Quint{\'\i} and Guisado-Clavero, Marina and Pons-Vigu{\'e}s, Mariona and Pujol-Ribera, Enriqueta and Valderas, Jose M},
  journal={BMC family practice},
  volume={19},
  number={1},
  pages={108},
  year={2018},
  publisher={Springer}
}

@article{MCA_PAPER5,
  title={An extension of multiple correspondence analysis for identifying heterogeneous subgroups of respondents},
  author={Hwang, Heungsun and Dillon, William R and Takane, Yoshio},
  journal={Psychometrika},
  volume={71},
  number={1},
  pages={161--171},
  year={2006},
  publisher={Springer-Verlag}
}

@incollection{MCA_PAPER6,
  title={Multiple correspondence k-means: simultaneous versus sequential approach for dimension reduction and clustering},
  author={Fordellone, Mario and Vichi, Maurizio},
  booktitle={Data Science and Social Research: Epistemology, Methods, Technology and Applications},
  pages={81--95},
  year={2017},
  publisher={Springer}
}

@inproceedings{kyono2021miracle,
	title        = {MIRACLE: Causally-Aware Imputation via Learning Missing Data Mechanisms},
	author       = {Kyono, Trent and Zhang, Yao and Bellot, Alexis and van der Schaar, Mihaela},
	year         = 2021,
	booktitle    = {Conference on Neural Information Processing Systems(NeurIPS) 2021}
}

@article{LCA,
  title={Exploratory latent structure analysis using both identifiable and unidentifiable models},
  author={Goodman, Leo A},
  journal={Biometrika},
  volume={61},
  number={2},
  pages={215--231},
  year={1974},
  publisher={Oxford University Press}
}

@article{kmodes,
  title={Extensions to the k-means algorithm for clustering large data sets with categorical values},
  author={Huang, Zhexue},
  journal={Data mining and knowledge discovery},
  volume={2},
  number={3},
  pages={283--304},
  year={1998},
  publisher={Springer}
}

@article{MCA_variant,
  title={Multiple correspondence analysis via polynomial transformations of ordered categorical variables},
  author={Lombardo, Rosaria and Meulman, Jacqueline J},
  journal={Journal of Classification},
  volume={27},
  number={2},
  pages={191--210},
  year={2010},
  publisher={Springer}
}

@article{MCA_limit,
  title={The need for interpretable features: Motivation and taxonomy},
  author={Zytek, Alexandra and Arnaldo, Ignacio and Liu, Dongyu and Berti-Equille, Laure and Veeramachaneni, Kalyan},
  journal={ACM SIGKDD Explorations Newsletter},
  volume={24},
  number={1},
  pages={1--13},
  year={2022},
  publisher={ACM New York, NY, USA}
}

@article{GLM_limitation,
  title={A reluctant additive model framework for interpretable nonlinear individualized treatment rules},
  author={Maronge, Jacob M and D HULING, JARED and Chen, Guanhua},
  journal={The annals of applied statistics},
  volume={17},
  number={4},
  pages={3384},
  year={2023}
}

@article{scikit,
  title={Scikit-learn: Machine learning in Python},
  author={Pedregosa, Fabian and Varoquaux, Ga{\"e}l and Gramfort, Alexandre and Michel, Vincent and Thirion, Bertrand and Grisel, Olivier and Blondel, Mathieu and Prettenhofer, Peter and Weiss, Ron and Dubourg, Vincent and others},
  journal={the Journal of machine Learning research},
  volume={12},
  pages={2825--2830},
  year={2011},
  publisher={JMLR. org}
}

@article{scikitfuzzy,
  title={JDWarner/scikit-fuzzy: Scikit-Fuzzy 0.5. 0},
  author={Warner, Josh and Sexauer, Jason and Van den Broeck, Wouter and Kinoshita, Bruno P and Balinski, Jakub and Clauss, Christian and Unnikrishnan, Aishwarya and Miretti, Marco and Castel{\~a}o, Guilherme and Arruda Pontes, Felipe and others},
  journal={Zenodo},
  year={2024}
}

@misc{econml,
  author={Keith Battocchi and Eleanor Dillon and Maggie Hei and Greg Lewis and Paul Oka and Miruna Oprescu and Vasilis Syrgkanis},
  title={{EconML}: {A Python Package for ML-Based Heterogeneous Treatment Effects Estimation}},
  howpublished={https://github.com/py-why/EconML},
  note={Version 0.x},
  year={2019}
}

@software{prince,
    author = {Halford, Max},
    license = {MIT},
    title = {{Prince}},
    url = {https://github.com/MaxHalford/prince}
}

@article{pymc,
  title = {{PyMC}: A Modern and Comprehensive Probabilistic Programming Framework in {P}ython},
  author = {Oriol Abril-Pla and Virgile Andreani and Colin Carroll and Larry Dong and Christopher J. Fonnesbeck and Maxim Kochurov and Ravin Kumar and Junpeng Lao and Christian C. Luhmann and Osvaldo A. Martin and Michael Osthege and Ricardo Vieira and Thomas Wiecki and Robert Zinkov },
  journal = {{PeerJ} Computer Science},
  volume = {9},
  number = {e1516},
  doi = {10.7717/peerj-cs.1516},
  year = {2023}
}

@article{causallearnpackage,
  title={Causal-learn: Causal discovery in python},
  author={Zheng, Yujia and Huang, Biwei and Chen, Wei and Ramsey, Joseph and Gong, Mingming and Cai, Ruichu and Shimizu, Shohei and Spirtes, Peter and Zhang, Kun},
  journal={Journal of Machine Learning Research},
  volume={25},
  number={60},
  pages={1--8},
  year={2024}
}

@article{pgmpy,
  title={Pgmpy: a python toolkit for Bayesian networks},
  author={Ankan, Ankur and Textor, Johannes},
  journal={Journal of Machine Learning Research},
  volume={25},
  number={265},
  pages={1--8},
  year={2024}
}

@article{abdi2010holm,
  title={Holm’s sequential Bonferroni procedure},
  author={Abdi, Herv{\'e}},
  journal={Encyclopedia of research design},
  volume={1},
  number={8},
  pages={1--8},
  year={2010},
  publisher={Thousand Oaks, California}
}

@article{holm1979simple,
  title={A simple sequentially rejective multiple test procedure},
  author={Holm, Sture},
  journal={Scandinavian journal of statistics},
  pages={65--70},
  year={1979},
  publisher={JSTOR}
}

@article{aickin1996adjusting,
  title={Adjusting for multiple testing when reporting research results: the Bonferroni vs Holm methods.},
  author={Aickin, Mikel and Gensler, Helen},
  journal={American journal of public health},
  volume={86},
  number={5},
  pages={726--728},
  year={1996},
  publisher={American Public Health Association}
}

@article{hall1991two,
  title={Two guidelines for bootstrap hypothesis testing},
  author={Hall, Peter and Wilson, Susan R},
  journal={Biometrics},
  pages={757--762},
  year={1991},
  publisher={JSTOR}
}

@article{piaggio2006reporting,
  title={Reporting of noninferiority and equivalence randomized trials: an extension of the CONSORT statement},
  author={Piaggio, Gilda and Elbourne, Diana R and Altman, Douglas G and Pocock, Stuart J and Evans, Stephen JW and CONSORT Group, for the and others},
  journal={Jama},
  volume={295},
  number={10},
  pages={1152--1160},
  year={2006},
  publisher={American Medical Association}
}
\end{document}